\documentclass[10pt,journal,compsoc]{IEEEtran}

\pdfoutput=1

\ifCLASSOPTIONcompsoc
  \usepackage[nocompress]{cite}
\else
  \usepackage{cite}
\fi

\ifCLASSINFOpdf
\else
\fi

\usepackage{isl-shortcuts}
\usepackage{tabularx}
\usepackage{wrapfig}
\pagecolor{white}

\usepackage{xcolor}
\usepackage[export]{adjustbox}

\newcommand{\myfootnotesize}{\fontsize{8pt}{10pt}\selectfont}

\begin{document}
\title{Enhancing photorealism enhancement}

\author{Stephan R. Richter,~
        Hassan Abu AlHaija,~
        and~Vladlen Koltun}

\markboth{}%
{Richter \MakeLowercase{\textit{et al.}}: Enhancing Photorealism Enhancement}

\IEEEtitleabstractindextext{%
\begin{abstract}
We present an approach to enhancing the realism of synthetic images. The images are enhanced by a convolutional network that leverages intermediate representations produced by conventional rendering pipelines. The network is trained via a novel adversarial objective, which provides strong supervision at multiple perceptual levels. We analyze scene layout distributions in commonly used datasets and find that they differ in important ways. We hypothesize that this is one of the causes of strong artifacts that can be observed in the results of many prior methods. To address this we propose a new strategy for sampling image patches during training. We also introduce multiple architectural improvements in the deep network modules used for photorealism enhancement. We confirm the benefits of our contributions in controlled experiments and report substantial gains in stability and realism in comparison to recent image-to-image translation methods and a variety of other baselines.

\end{abstract}
}

\maketitle

\IEEEdisplaynontitleabstractindextext
\IEEEpeerreviewmaketitle

\IEEEraisesectionheading{\section{Introduction}\label{sec:introduction}}

\IEEEPARstart{P}{hotorealism} has been the defining goal of computer graphics for half a century. In 1977, Newell and Blinn~\cite{NewellBlinn1977} surveyed a decade of work on this problem. In the ensuing four decades, substantial further progress has been made, due in part to physically based simulation of light transport~\cite{Pharr2016}, principled representation of material appearance~\cite{Nicodemus1977,Weyrich2009}, and photogrammetric modeling~\cite{Knapitsch2017}. These techniques and their approximations have been integrated into real-time rendering pipelines, substantially advancing the realism of computer games~\cite{Moller2018}. Nevertheless, a look at even the most sophisticated real-time games will quickly reveal that photorealism has not been achieved. An ineffable difference in the appearance of simulation and reality remains.

In recent years, a complementary set of techniques has been developed in computer vision and machine learning. These techniques, based on deep learning, convolutional networks, and adversarial training, bypass physical modeling of geometric layout, material appearance, and light transport. Instead, images are synthesized by convolutional networks trained on large datasets. These techniques have been used to synthesize representative images from a given domain~\cite{Karras2018,Brock2019,Karras2019}, to convert semantic label maps to photographic images~\cite{Isola2017,ChenKoltun2017,Wang2018:CVPR,Qi2018,Park2019:CVPR,Liu2019,Mallya2020}, and to attempt to bridge the appearance gap between synthetic and real images~\cite{Shrivastava2017,Li2017,Hoffman2018,Huang2018,Lee2018,Wu2018,Cherian2019,Ma2019,Dundar2018}. Images synthesized by these approaches capture aspects of photographic appearance that often elude even state-of-the-art computer games. On the flip side, these approaches are largely disconnected from the rendering pipelines that drive computer games, can be hard to control, and often produce jarring artifacts that would be unacceptable in production-quality media.

In this work, we take a step towards melding these two complementary routes to photorealism.
We seek to build on the infrastructure developed in the production of modern games and enhance their photorealism via techniques developed in the deep learning community. Our starting point is a set of intermediate buffers (G-buffers) produced by game engines during the rendering process~\cite{Saito1990,Moller2018}. These buffers provide detailed information on geometry, materials, and lighting in the scene. We train convolutional networks with these auxiliary inputs to enhance the realism of images produced by the rendering pipeline.
To integrate these buffers into the photorealism enhancement flow, we design new network components that modulate features from a rendered image according to information extracted from the buffers.
\begin{figure*}[t]
	\centering
	\newcolumntype{P}[1]{>{\centering\arraybackslash}p{#1}}
	\renewcommand{\arraystretch}{0.5}
	\begin{tabular}{@{}P{7.5cm}@{}P{3cm}@{}P{7.5cm}@{}}
		\footnotesize Rendered images from GTA V  & & \footnotesize Enhancement by our method (trained to mimic Cityscapes)\\
		\begin{minipage}{7.5cm}
			\begin{tikzpicture}
			\node[anchor=north west,inner sep=0] (image) at (0,0) {\includegraphics[width=7.5cm]{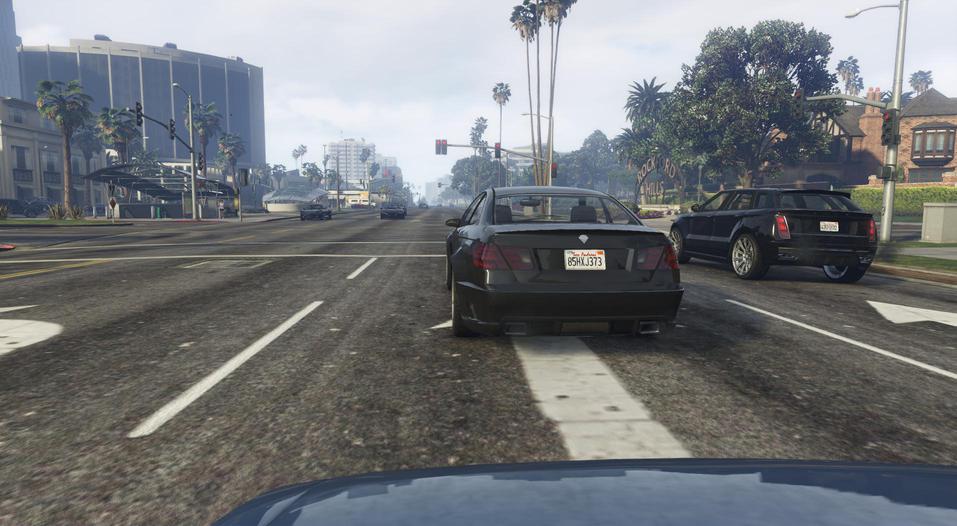}};
			\begin{scope}[x={(image.north east)},y={(image.south west)}]
				\draw[BurntOrange,thick] (0.445, 0.35) rectangle (0.645, 0.59);
		    \end{scope}
			\end{tikzpicture}
		\end{minipage}
		\vspace{0.1mm}
		&
		\begin{minipage}{3cm}
		\includegraphics[width=3cm,trim={15cm 7.5cm 12cm 6.41cm},clip,cfbox=BurntOrange 0.5mm -0.5mm]{figures/comparison/gta/17328.jpg}
		\includegraphics[width=3cm,trim={15cm 7.5cm 12cm 6.41cm},clip,cfbox=Cerulean 0.5mm -0.5mm]{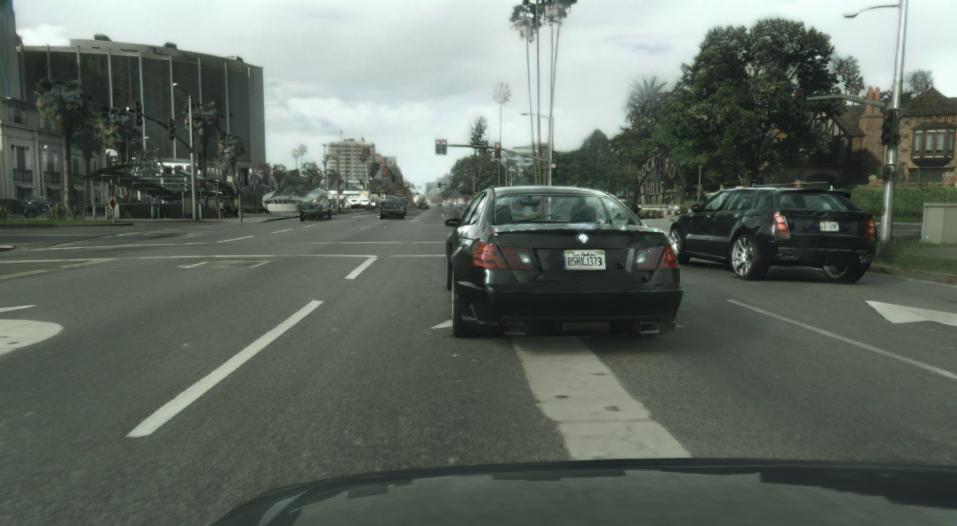}
		\end{minipage}
		\vspace{0.1mm}
		&
		\begin{minipage}{7.5cm}
			\begin{tikzpicture}
				\node[anchor=north west,inner sep=0] (image) at (0,0) {\includegraphics[width=7.5cm]{figures/comparison/ours/17328.jpg}};
				\begin{scope}[x={(image.north east)},y={(image.south west)}]
					\draw[Cerulean,thick] (0.445, 0.35) rectangle (0.645, 0.59);
			    \end{scope}
				\end{tikzpicture}
		\end{minipage}
		\vspace{0.1mm}\\
		\begin{minipage}{7.5cm}
			\begin{tikzpicture}
			\node[anchor=north west,inner sep=0] (image) at (0,0) {\includegraphics[width=7.5cm]{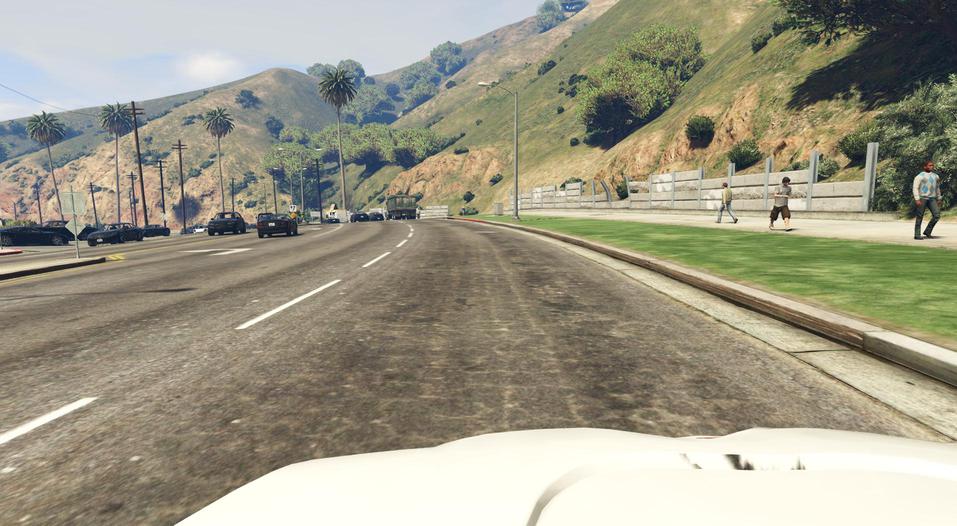}};
			\begin{scope}[x={(image.north east)},y={(image.south west)}]
				\draw[BurntOrange,thick] (0.415, 0.05) rectangle (0.615, 0.29);
		    \end{scope}
			\end{tikzpicture}
		\end{minipage}
		\vspace{0.1mm}
		&
		\begin{minipage}{3cm}
		\includegraphics[width=3cm,trim={14cm 13cm 13cm 0.91cm},clip,cfbox=BurntOrange 0.5mm -0.5mm]{figures/comparison/gta/01321.jpg}
		\includegraphics[width=3cm,trim={14cm 13cm 13cm 0.91cm},clip,cfbox=Cerulean 0.5mm -0.5mm]{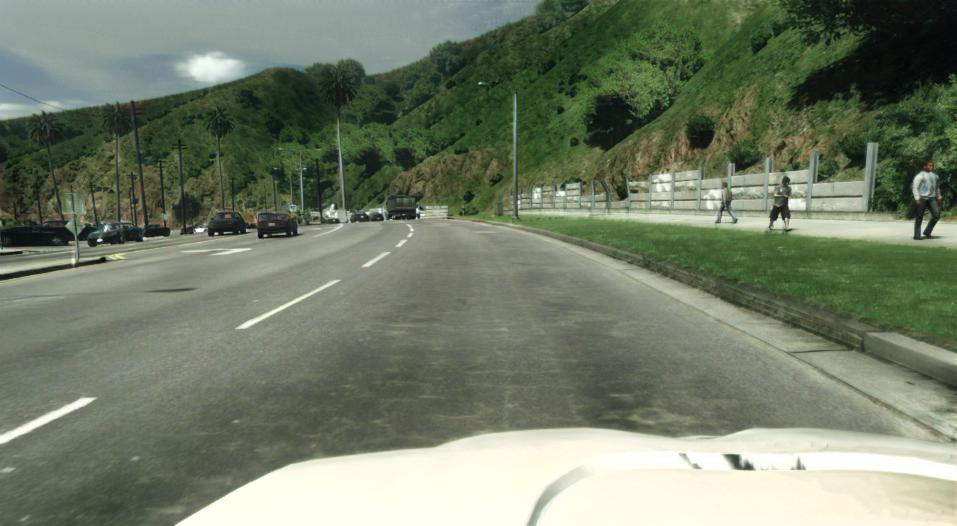}
		\end{minipage}
		\vspace{0.1mm}
		&
		\begin{minipage}{7.5cm}
			\begin{tikzpicture}
				\node[anchor=north west,inner sep=0] (image) at (0,0) {\includegraphics[width=7.5cm]{figures/comparison/ours/01321.jpg}};
				\begin{scope}[x={(image.north east)},y={(image.south west)}]
					\draw[Cerulean,thick] (0.415, 0.05) rectangle (0.615, 0.29);
			    \end{scope}
				\end{tikzpicture}
		\end{minipage}
		\vspace{0.1mm}\\
		\begin{minipage}{7.5cm}
			\begin{tikzpicture}
			\node[anchor=north west,inner sep=0] (image) at (0,0) {\includegraphics[width=7.5cm]{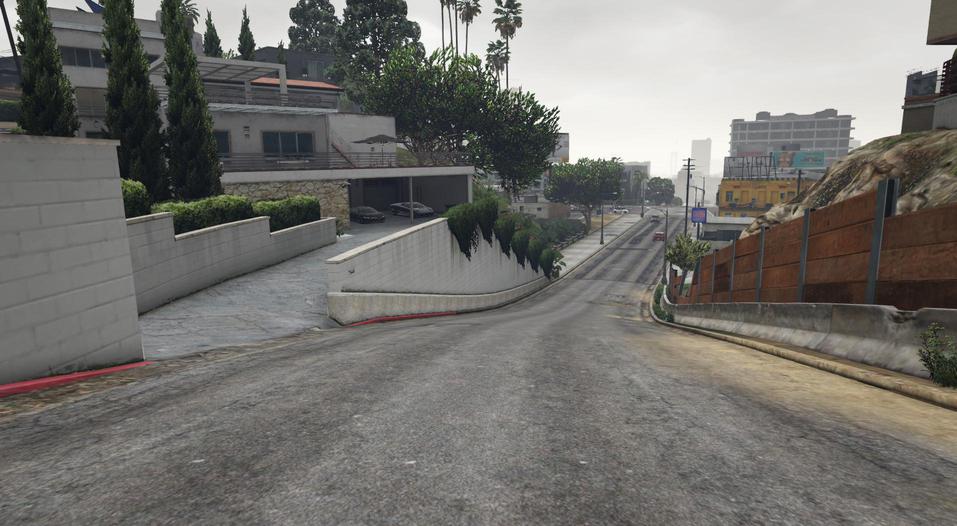}};
			\begin{scope}[x={(image.north east)},y={(image.south west)}]
				\draw[BurntOrange,thick] (0.682, 0.68) rectangle (0.882, 0.92);
		    \end{scope}
			\end{tikzpicture}
		\end{minipage}
		\vspace{0.1mm}
		&
		\begin{minipage}{3cm}
		\includegraphics[width=3cm,trim={23cm 1.5cm 4cm 12.41cm},clip,cfbox=BurntOrange 0.5mm -0.5mm]{figures/comparison/gta/17651.jpg}
		\includegraphics[width=3cm,trim={23cm 1.5cm 4cm 12.41cm},clip,cfbox=Cerulean 0.5mm -0.5mm]{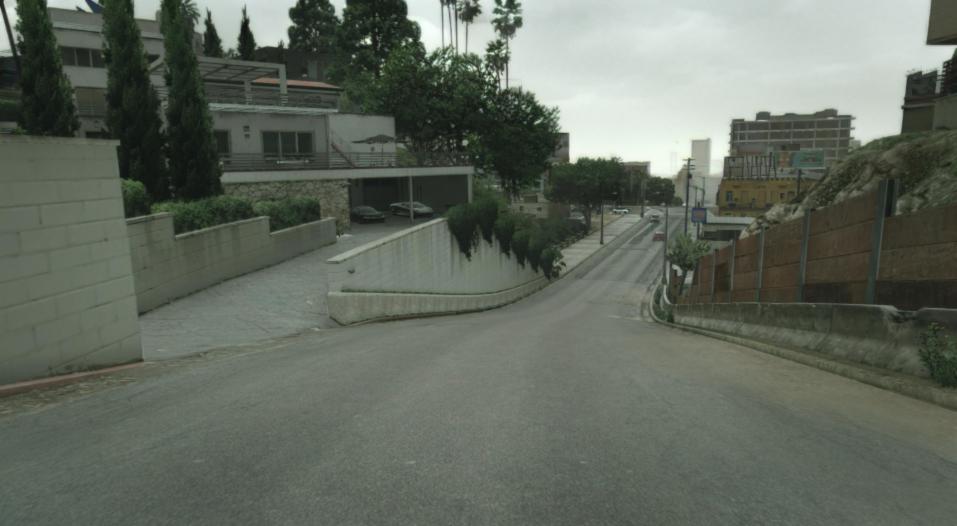}
		\end{minipage}
		\vspace{0.1mm}
		&
		\begin{minipage}{7.5cm}
			\begin{tikzpicture}
				\node[anchor=north west,inner sep=0] (image) at (0,0) {\includegraphics[width=7.5cm]{figures/comparison/ours/17651.jpg}};
				\begin{scope}[x={(image.north east)},y={(image.south west)}]
					\draw[Cerulean,thick] (0.682, 0.68) rectangle (0.882, 0.92);
			    \end{scope}
				\end{tikzpicture}
		\end{minipage}\\
	\end{tabular}
	\caption{We train convolutional networks to enhance the photorealism of rendered images, using intermediate buffers produced by a conventional rendering engine. Left: frames from a modern computer game (GTA V)~\cite{Richter2016}. Right: same frames enhanced by our approach to mimic the style of Cityscapes~\cite{Cordts2016}. Enhancements by our method are semantically consistent and non-trivial. For example, our method adds gloss to cars (\nth{1} row), reforests parched hills to mimic German climate (\nth{2} row), and paves roads with smoother asphalt (\nth{3} row). Insets magnify marked regions.}
	\label{fig:teaser}
\end{figure*}

We also seek to eliminate artifacts that can be seen in the results of prior deep-learning approaches, which often hallucinate objects.
To this end, we analyze the datasets that are commonly used for photorealism enhancement.
Our analysis reveals that their scene layouts differ in ways that can explain artifacts commonly seen in prior work.
To better align the datasets and alleviate the artifacts, we propose a new strategy for sampling image patches during training.
We further design a new adversarial training objective that facilitates enhancements that are geometrically and semantically consistent with the content of the input image.

\begin{figure*}[t]
	\centering
	\newcolumntype{P}[1]{>{\centering\arraybackslash}p{#1}}
	\begin{tabularx}{\linewidth}{@{}P{90mm}@{\hspace{1mm}}P{90mm}@{}}
		\myfootnotesize GTA & \myfootnotesize Ours (trained on KITTI)\\
		\begin{minipage}{90mm}
		\includegraphics[width=90mm]{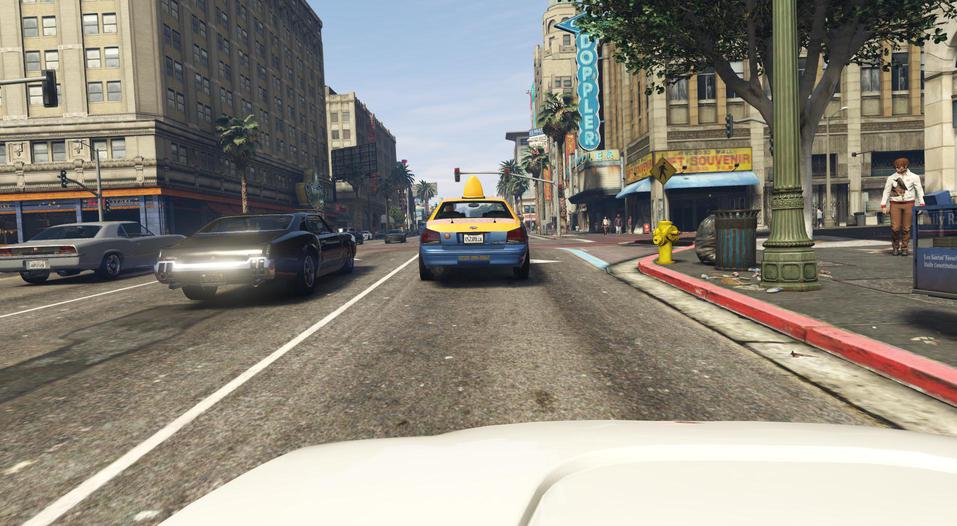}
		\vspace{1mm}
		\end{minipage}&
		\begin{minipage}{90mm}
		\includegraphics[width=90mm,trim={0cm 5.1cm 0cm 0cm},clip]{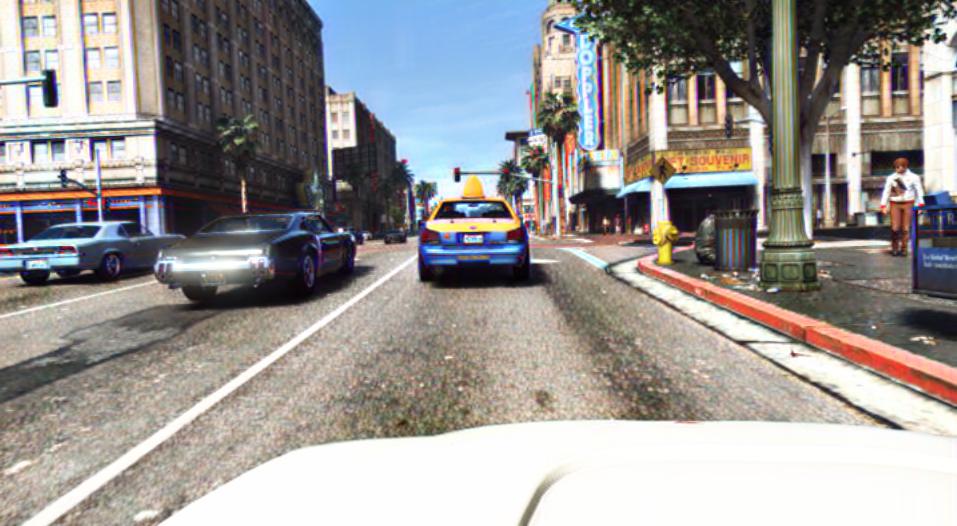}
		\begin{tabular}{@{}p{45mm}@{}p{45mm}@{}}
		\includegraphics[width=45mm]{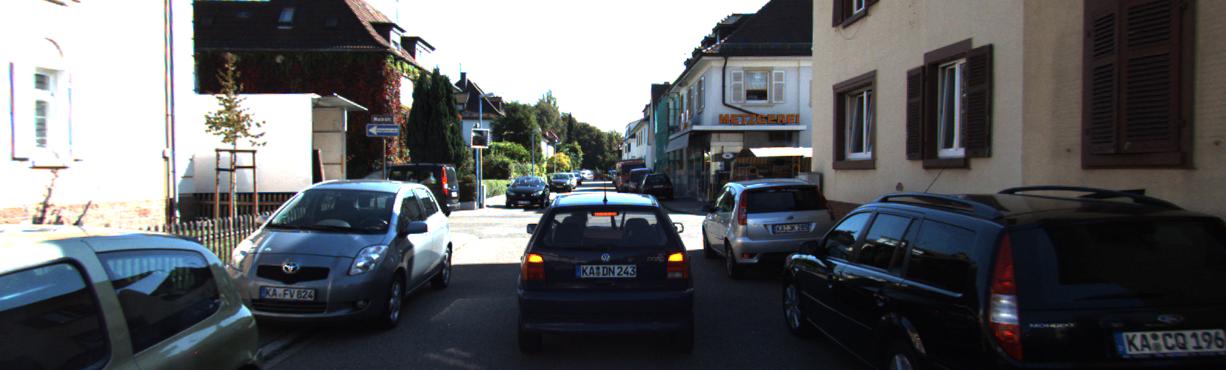} &
		\includegraphics[width=45mm]{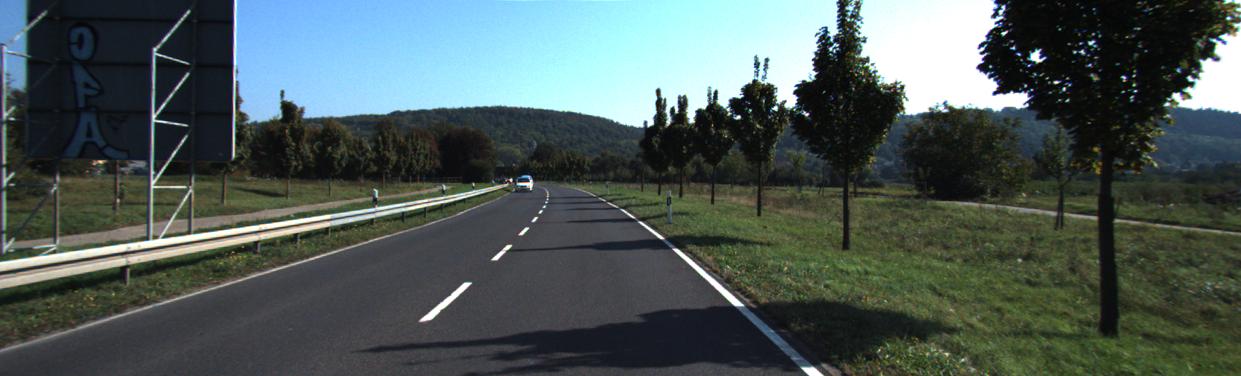} \\
		\end{tabular}
		\end{minipage}\\
		\myfootnotesize Ours (trained on Cityscapes) & \myfootnotesize Ours (trained on Mapillary Vistas)\\
		\begin{minipage}{90mm}
		\includegraphics[width=90mm,trim={0cm 5cm 0cm 0cm},clip]{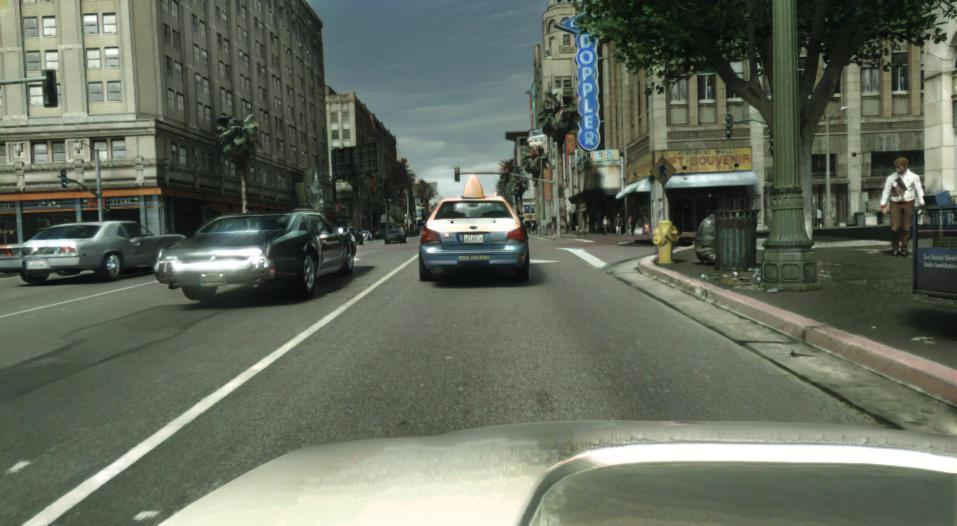} \\
		\begin{tabular}{@{}p{30mm}@{}p{30mm}@{}p{30mm}@{}}
		\includegraphics[width=30mm]{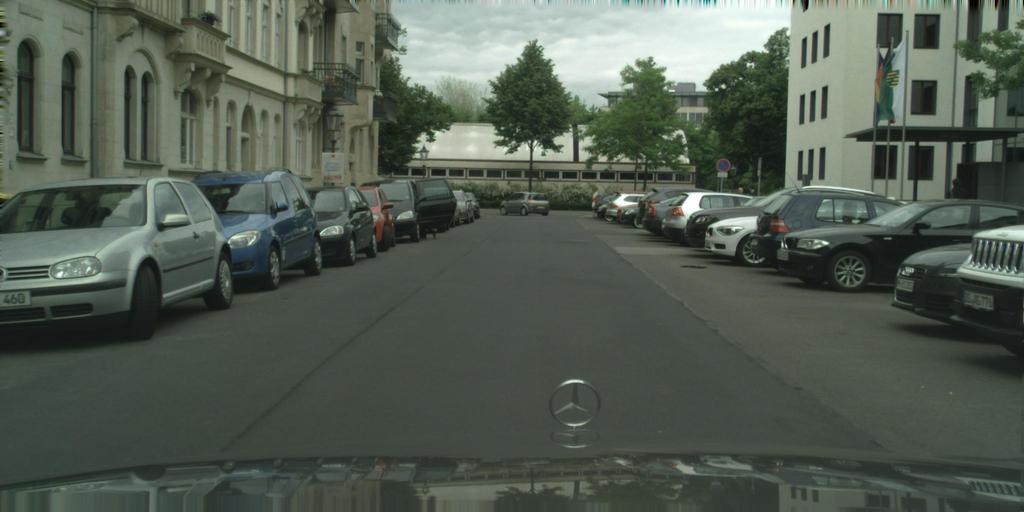} &
		\includegraphics[width=30mm]{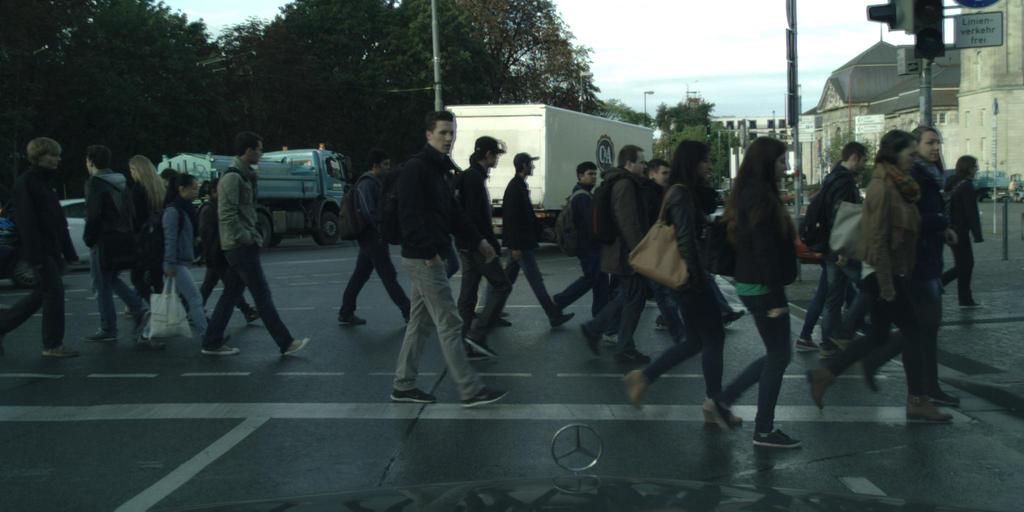} &
		\includegraphics[width=30mm]{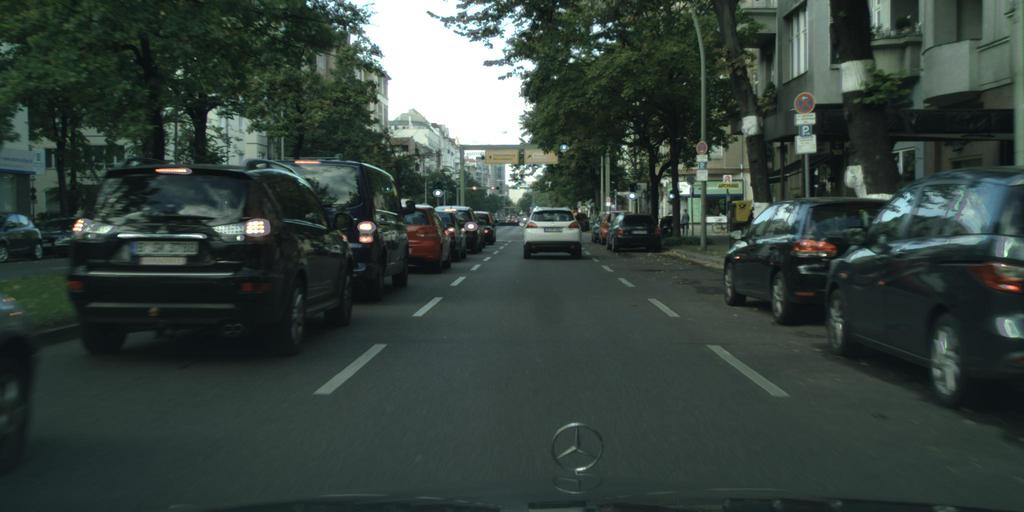} \\
		\end{tabular}
		\end{minipage} &
		\begin{minipage}{90mm}
		\includegraphics[width=90mm,trim={0cm 5.7cm 0cm 0cm},clip]{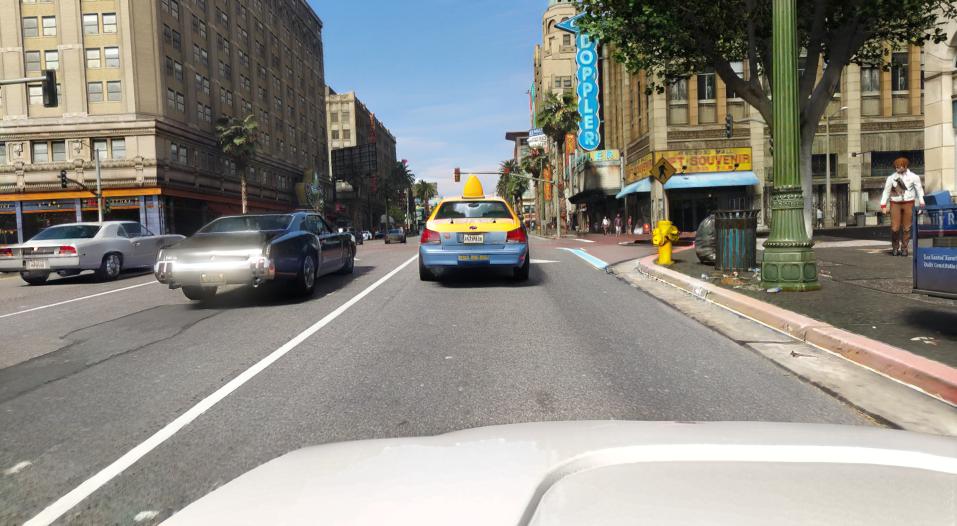} \\
		\begin{tabular}{@{}p{22.5mm}@{}p{22.5mm}@{}p{22.5mm}@{}p{22.5mm}@{}}
		\includegraphics[width=22.5mm]{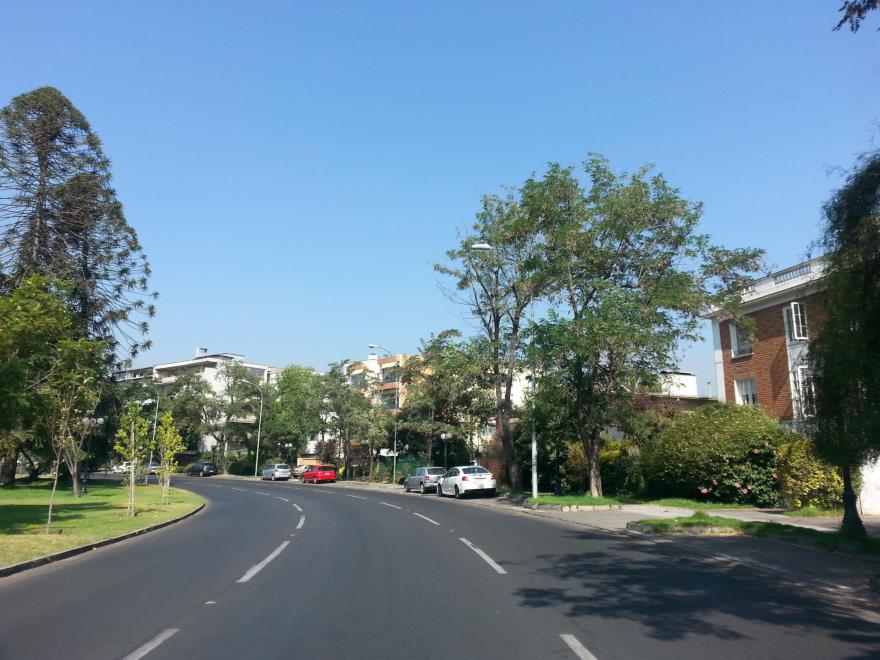} &
		\includegraphics[width=22.5mm]{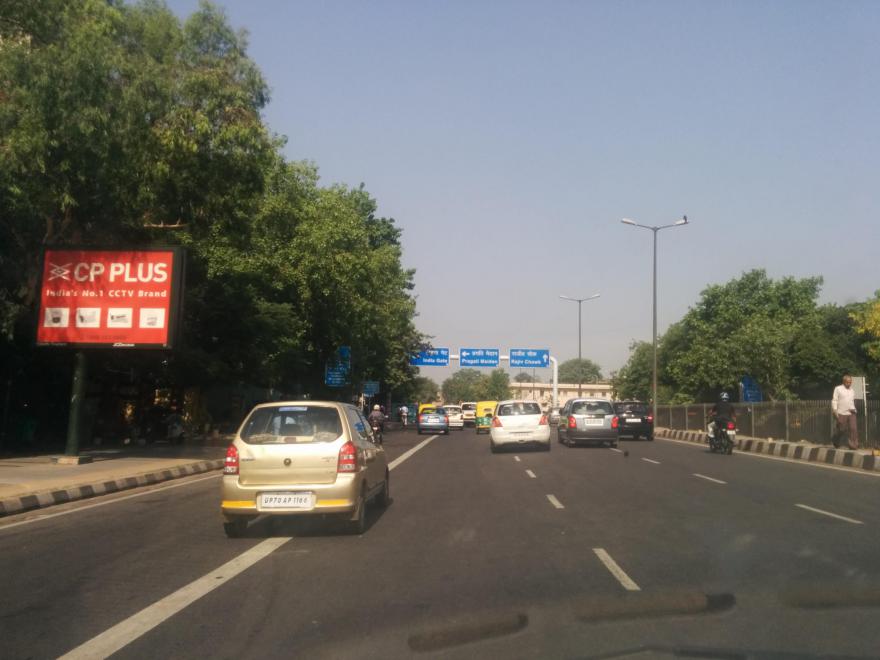} &
		\includegraphics[width=22.5mm]{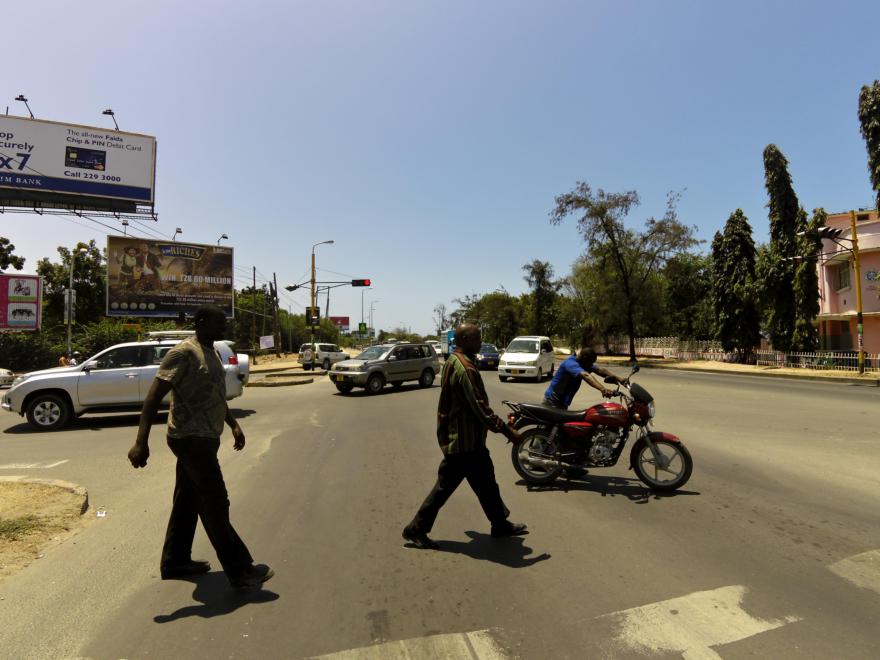} &
		\includegraphics[width=22.5mm]{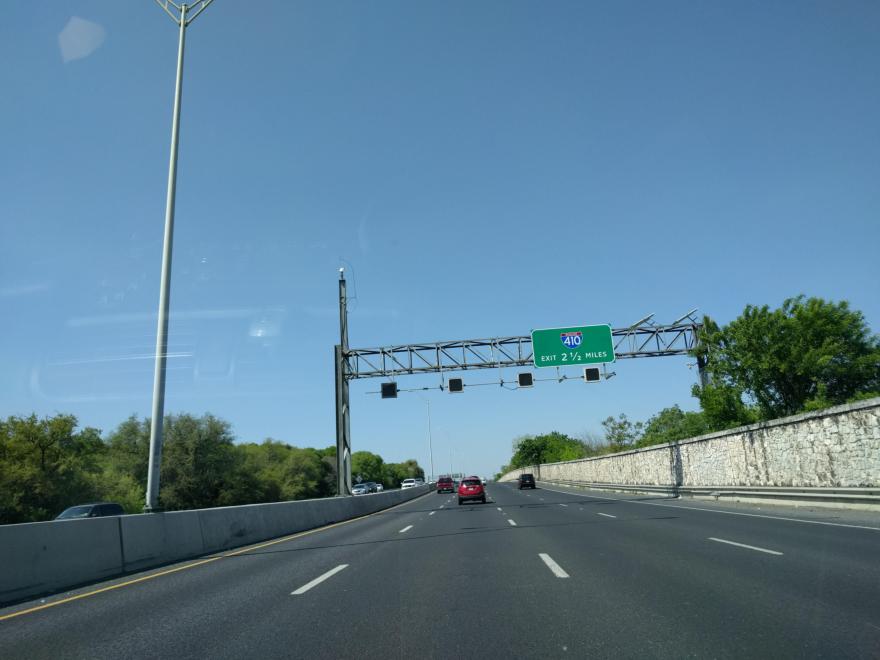} \\
		\end{tabular}
		\end{minipage} \\
	\end{tabularx}
	\caption{Targeting different real-world datasets. We train our method to enhance images from GTA (top left) with KITTI, Cityscapes, and Vistas as target datasets. Our method is able to reproduce the characteristic appearance of these datasets (\eg sensor noise in KITTI, saturation in Cityscapes, fine textures in Vistas) while keeping the structure of the original GTA images. Insets show sample images from the respective target datasets.}
	\label{fig:results_diverse_datasets}
\end{figure*}

Combining all of our contributions, our approach significantly enhances the photorealism of rendered images (\Fig~\ref{fig:teaser}).
It can add gloss to cars (\nth{1} row), green parched hills (\nth{2} row), and rebuild roads (\nth{3} row).
Training it with different real-world image collections (\eg Cityscapes~\cite{Cordts2016}, KITTI~\cite{Geiger2012}, or Mapillary Vistas~\cite{Neuhold2017}) expresses the corresponding visual styles in the output (\Fig~\ref{fig:results_diverse_datasets}).

Our analysis further suggests that standard metrics confound differences in style and content.
Motivated by this observation, we develop a new family of metrics that mitigate the effect of mismatched scene layouts and provide a finer-grained assessment of realism at multiple levels.

We compare the presented approach against a broad array of strong baselines that represent diverse perspectives on photorealism enhancement.
We also conduct a perceptual experiment to assess photorealism.
The results indicate that our approach consistently produces the most realistic results, by a wide margin.
In all experiments, our approach outperforms all baselines and sets a new state of the art in photorealism enhancement.

\section{Related work}\label{sec:related}
Photorealistic images can be synthesized by simulating all the physical processes involved in image formation.
However, this simulation is computationally expensive, may not be feasible at interactive rates, and requires physically accurate models of scenes, objects, materials, and lighting.
As a result, a variety of approximations have been developed that generate images that may not be physically correct, but nevertheless appear realistic to varying extents~\cite{Moller2018,Reinhard2013}.

Methods for conditional image synthesis aim to learn the complete image formation process from data~\cite{ChenKoltun2017,Isola2017,Park2019:CVPR,Wang2018:CVPR,Wang2018:Neurips,Wang2019,Zhu2020,Jiang2020}. These works often focus on synthesizing images from semantic label maps. As such, the synthesis is severely underconstrained.
Since geometric structure is only provided through the silhouettes of objects and their composition in the label map, substantial ambiguity remains, leading to visible artifacts and temporal inconsistency.
Furthermore, the reliance on semantic label maps requires annotated real-world data, which is extremely laborious to create at large scale~\cite{Cordts2016}.
Instead of trying to synthesize images, our approach enhances already rendered images, integrates scene information to produce geometrically and semantically consistent images, and does not require any annotation of real data.

Image-based rendering techniques can produce results that are indistinguishable from photographs by recycling real imagery of a scene for rendering novel views~\cite{Debevec1996,Shum2000review,Wood2000Surface,Buehler2001,Kopf2014First,Hedman2018,Broxton2020,Mildenhall2020Nerf,RieglerKoltun2021}.
However, they require capturing photos of the scene of interest beforehand and make it difficult to manipulate captured scenes afterward.
Furthermore, novel views need to be fairly close to the prerecorded camera trajectory to avoid artifacts.

Another promising direction combines traditional rendering with data-driven approaches.
Johnson \etal proposed to improve the realism of rendered images by transferring nearest neighbor patches from similarly structured photographs~\cite{Johnson2011}.
Liao \etal retrieved nearest neighbor patches in a learned feature space~\cite{Liao2017}.
Reinhard \etal matched color distributions of a source image to a target image~\cite{Reinhard2001} and showed that this can enhance the realism of computer graphics images. Subsequent work introduced increasingly sophisticated techniques for photographic style transfer between images~\cite{Pitie2005, Pitie2007, Luan2017, Mechrez2017, Li2017, Liu2017, Li2018:ECCVa, Yoo2019}.
While early CNN-based methods required costly optimization per image~\cite{Gatys2016},
improvements in feature normalization reduced computation~\cite{Ulyanov2016},
enabled transfer to new styles~\cite{HuangBelongie2017}, and increased semantic consistency~\cite{Park2019:CVPR,Jiang2020,Zhu2020}.
Recent work has focused on encoder/decoder structures, mixing content and style of source and target images~\cite{Li2018:ECCVa,Li2019}.
Common to all style-transfer and example-based approaches~\cite{Ma2019} is the dependence on a favorable reference image. Differences in content or layout between rendered and reference images hurt the quality of enhanced images.
This is especially problematic for enhancing dynamic content. Our approach learns the style from an image collection through adversarial training and does not require reference images for enhancement.

Our approach is aligned with work on unpaired image-to-image translation, which learns a mapping from one image collection to another~\cite{Zhu2017,Lee2018,Hoffman2018,Huang2018,Park2020}.
By training on large datasets, these methods do not depend on favorable reference images, and often learn to capture the unique styles of datasets very well.
However, as there are no direct correspondences across image collections (\eg through paired images), the methods need to learn suitable correspondences implicitly.
This is challenging and often produces images that appear realistic at first glance, but contain artifacts that are inconsistent with the input images.
Improving the consistency between input and output has received a lot of attention and led to additional constraints on this ill-defined problem.
Notable improvements came from cycle-consistency~\cite{Zhu2017,Yi2017}, custom attention modules~\cite{Kim2020}, temporal regularization~\cite{Park2019:MM}, modeling sensor noise~\cite{Carlson2019}, geometric constraints in the image plane~\cite{Fu2019}, constraints derived from depth maps~\cite{Bousmalis2017}, and contrastive losses~\cite{Park2020}.

Many approaches to image synthesis or translation have employed adversarial objectives, which involve a discriminator network that evaluates the realism of generated images.
The discriminator is commonly trained alongside a network performing the image generation~\cite{Isola2017,Zhu2017}.
One intention of this setup is for the discriminator to learn high-level semantic concepts to provide high-quality supervision to the generator network. However, with a simple binary classification objective, the discriminator may focus on low-level textures and patches instead, since these already provide discriminative features. To direct attention to high-level semantic content, a number of modifications have been proposed. For example, the binary real vs.\ fake decision can be accompanied by a classification objective~\cite{Odena2017,Tseng2020}.
Additional semantic segmentation maps can be concatenated to the input image~\cite{Mirza2014,Wang2018:CVPR}, projected to a high-dimensional feature space~\cite{Brock2019,Liu2019,Miyato2018:ICLRb}, or processed via a separate network stream~\cite{Ntavelis2020}. These works use ground-truth annotations for guiding the discriminator. While this provides ground-truth semantic information, it restricts the image collections available for training to densely annotated datasets compatible with the synthetic label maps at hand.
In our work, we leverage a robust semantic segmentation network~\cite{Lambert2020} to provide label maps that are approximately consistent for synthetic and real images. Previously, Hoffman et al.\ used a segmentation network for synthetic-to-real translation~\cite{Hoffman2018}. In contrast to prior work~\cite{Hoffman2018,Cherian2019}, we neither train the segmentation network on any of our datasets nor apply it to enhanced images. This is for two reasons. First, training a segmentation network on synthetic data is prone to overfitting. Thus a network trained in this way may not provide semantically consistent guidance for the transition from synthetic to real images during the training of the generator. Second, incorporating a segmentation network during training can result in generated images that are easy for the segmentation network to parse but are not necessarily more realistic.

Providing the discriminator with additional semantic information supports adaptive processing for different semantic categories. Incorporating the information into the learning objective is an even stronger type of supervision, as the distinction between different objects is explicitly learned. However, datasets that are commonly used in synthetic-to-real translation are limited in their diversity, especially in the context of urban driving~\cite{Richter2016,Ros2016,Cordts2016}. Thus high-capacity networks may overfit to spurious artifacts in the data, leading to suboptimal performance of the generator~\cite{Karras2020,Zhang2019}. Prior research addressed this overfitting via regularization~\cite{Jia2020,Mescheder2018} or augmentation~\cite{Karras2020}. We take inspiration from the neural patch discriminator of Li and Wand~\cite{Li2016} and train our discriminator on feature maps from a pretrained VGG network. In contrast to the single-image style transfer approach of Li and Wand, we use multiple feature maps extracted from different layers of the VGG network, and train a discriminator on each.

Prior work hypothesized that differences in scene layout may be a key contributing factor in suboptimal performance in image translation~\cite{Hoffman2018} or in training semantic segmentation models~\cite{Li2020, Peng2017}, and addressed this challenge via regularization~\cite{Jia2020}, semantic consistency losses~\cite{Hoffman2018,Li2018:BMVC}, and matching source and target images~\cite{Li2020}. Akin to the work of Li~\etal on semantic segmentation~\cite{Li2020}, we address the layout mismatch via semantics-aware sampling. However, Li~\etal train a dedicated network for matching samples across datasets at image level, which significantly reduces the samples available for training, and hence diversity. We devise a simpler sampling strategy that requires no training and operates at the patch level, leaving sufficiently many diverse samples for training.

Our work is inspired by hybrid approaches that integrate the conventional rendering pipeline in a more structured way, such that information about a scene generated during rendering can be subsequently exploited by a learning-based approach. Nalbach~\etal demonstrated that a CNN can learn to shade G-buffers rendered by a conventional rendering pipeline~\cite{Nalbach2017}.
AlHaija~\etal combined this approach with an adversarial loss to improve the realism of rendered images~\cite{AlHaija2018}.
We introduce a different network structure that better integrates the G-buffers, as well as a new discriminator architecture and training objective, leading to significantly better results.

Bi \etal developed a pipeline that enhanced the realism of low-quality renderings of indoor scenes~\cite{Bi2019}. Their approach relies on the availability of paired low-quality and high-quality renderings of the same scenes, involves a number of synthesis stages with different losses, and produces results of limited fidelity. Our work is different in several ways. We assume higher-quality input, as produced by modern computer games. We do not rely on the availability of rendered images at different levels of quality, but do utilize auxiliary buffers produced by rendering engines. Our approach also differs in the network architecture, training objectives, and the application domain. We demonstrate significantly better results, with temporal consistency and measurably higher realism than high-production-value commercial games.

\section{Method}\label{sec:method}
\begin{figure}[t]
	\centering
	\includegraphics[width=85mm]{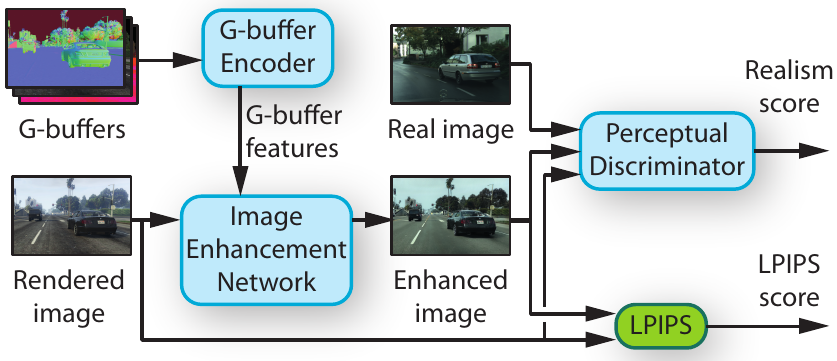}
	\caption{Our approach is built around an image enhancement network that transforms a rendered image. In addition to the image, the network ingests G-buffer feature tensors at multiple scales. The tensors represent rendering information from a conventional graphics pipeline, encoded by a G-buffer encoder network. Both networks are trained jointly via an LPIPS loss (to retain the structure of the rendered image) and a perceptual discriminator (to maximize the realism of the enhanced image).}
	\label{fig:overview}
\end{figure}

\begin{figure*}[t]
	\centering
	\includegraphics[width=152mm]{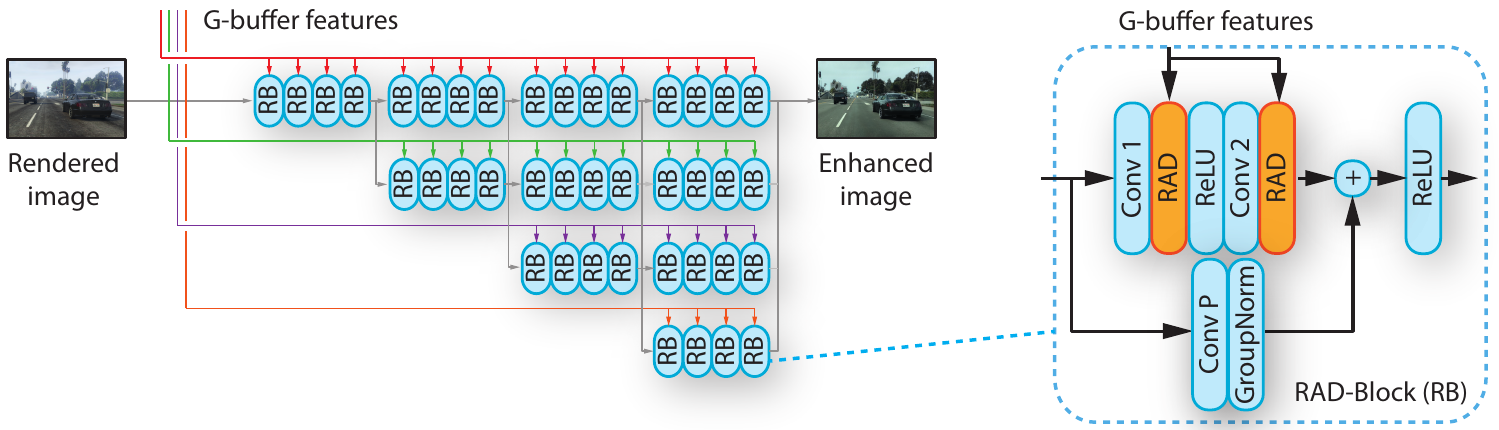}
	\caption{Image enhancement network. We replace the batch normalization layers within HRNet by rendering-aware denormalization (RAD), forming RAD blocks (RB). Each branch of the HRNet receives a G-buffer feature tensor at a matching scale (different scales are coded by color). Original feature streams are shown in {\color{Gray} gray}. We omit initial stem convolutions as well as transition and fusion layers for clarity.}
	\label{fig:hrnet}
\end{figure*}

\begin{figure*}[t]
	\centering
	\begin{tabular}{@{}c*{6}{@{\hspace{0mm}}c}@{}}
	\footnotesize Image &
	\footnotesize Normal &
	\footnotesize Depth &
	\footnotesize Albedo &
	\footnotesize Glossiness &
	\footnotesize Atmosphere &
	\footnotesize Segmentation \\
	\includegraphics[width=25mm]{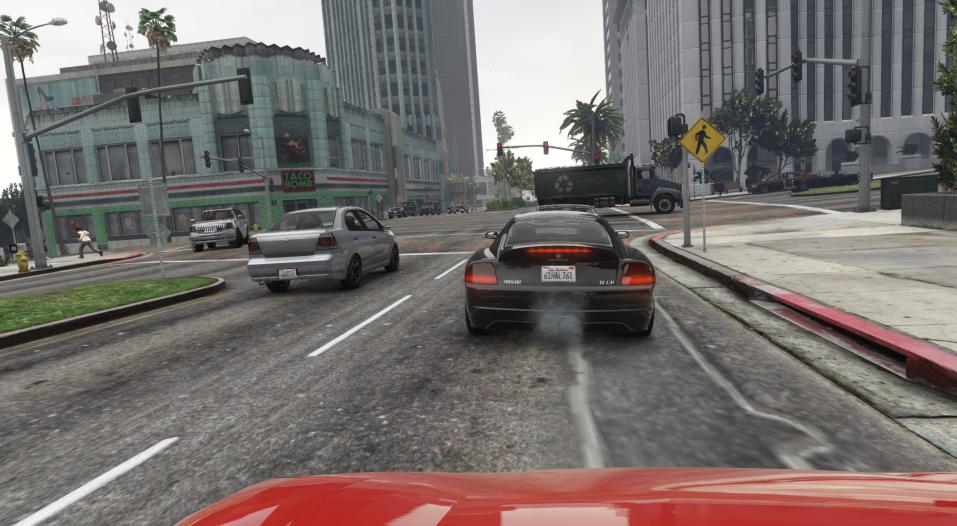} &
	\includegraphics[width=25mm]{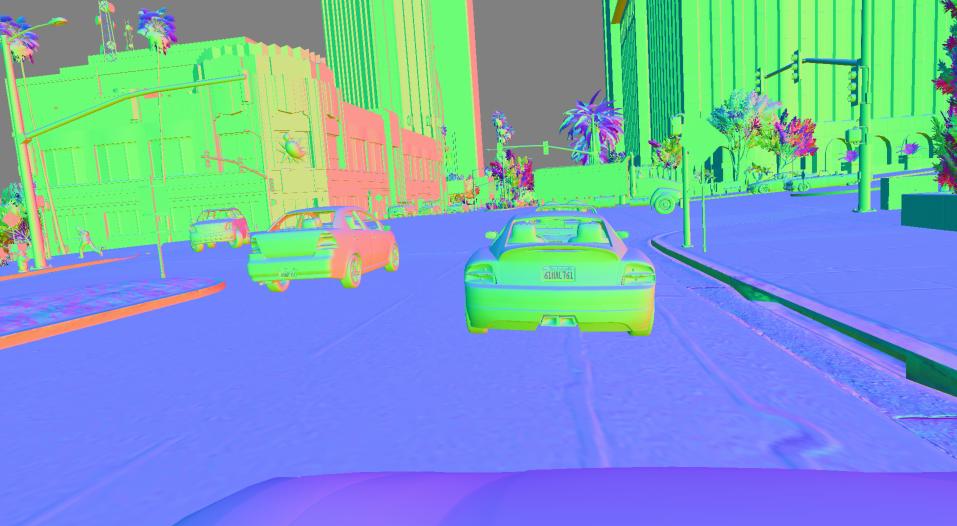} &
	\includegraphics[width=25mm]{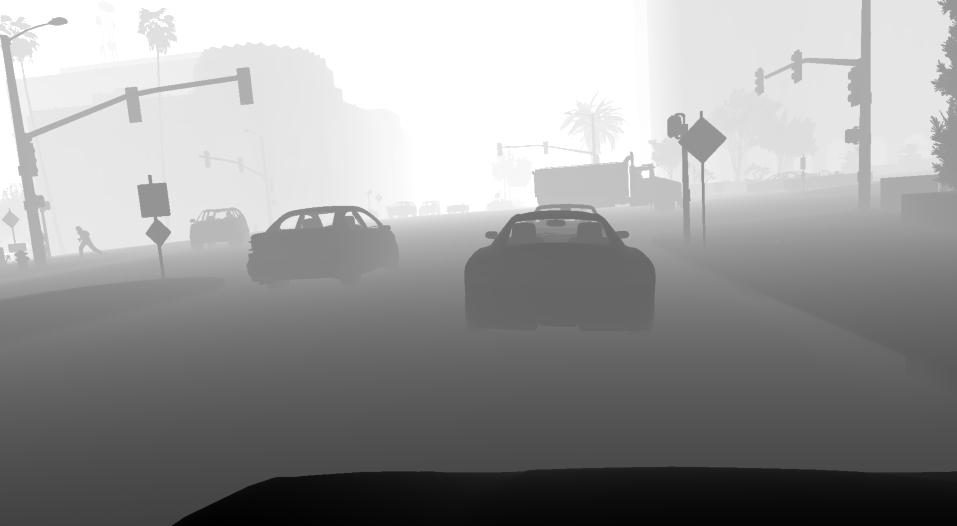} &
	\includegraphics[width=25mm]{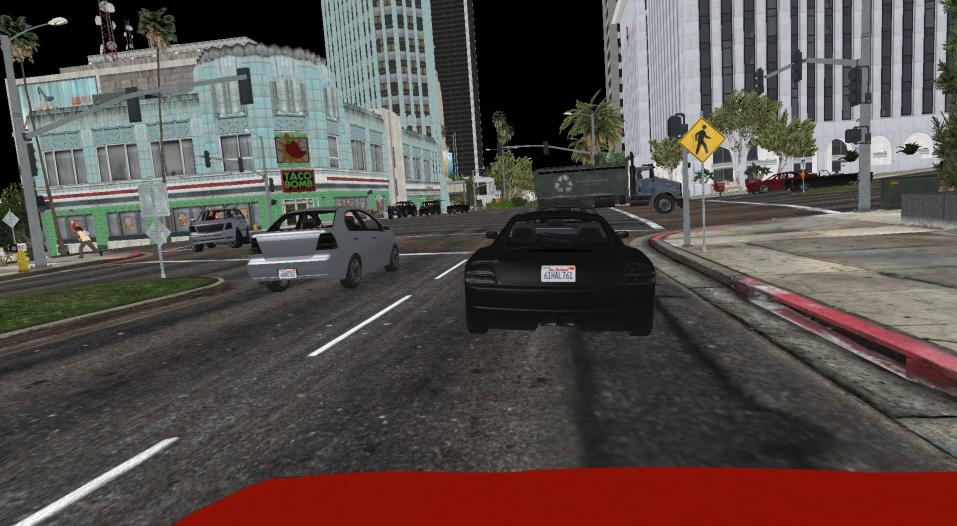} &
	\includegraphics[width=25mm]{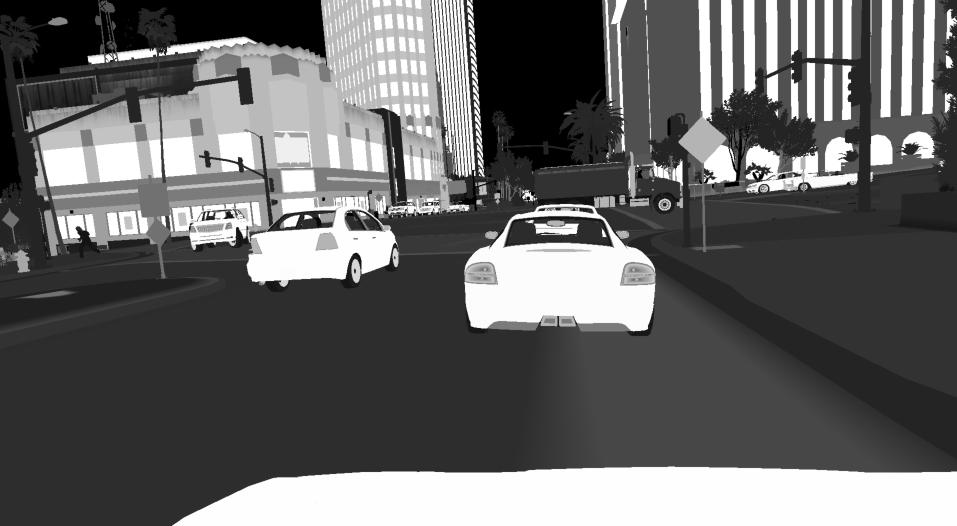} &
	\includegraphics[width=25mm]{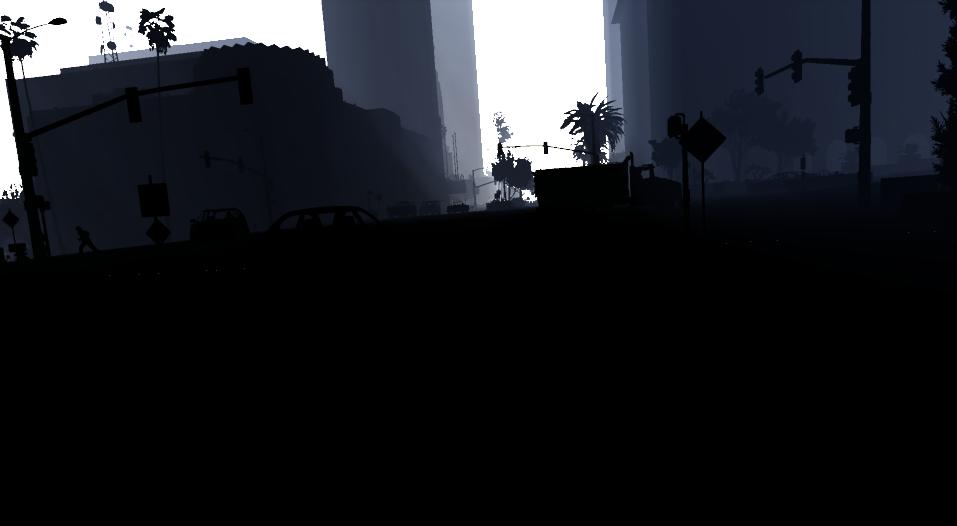} &
	\includegraphics[width=25mm]{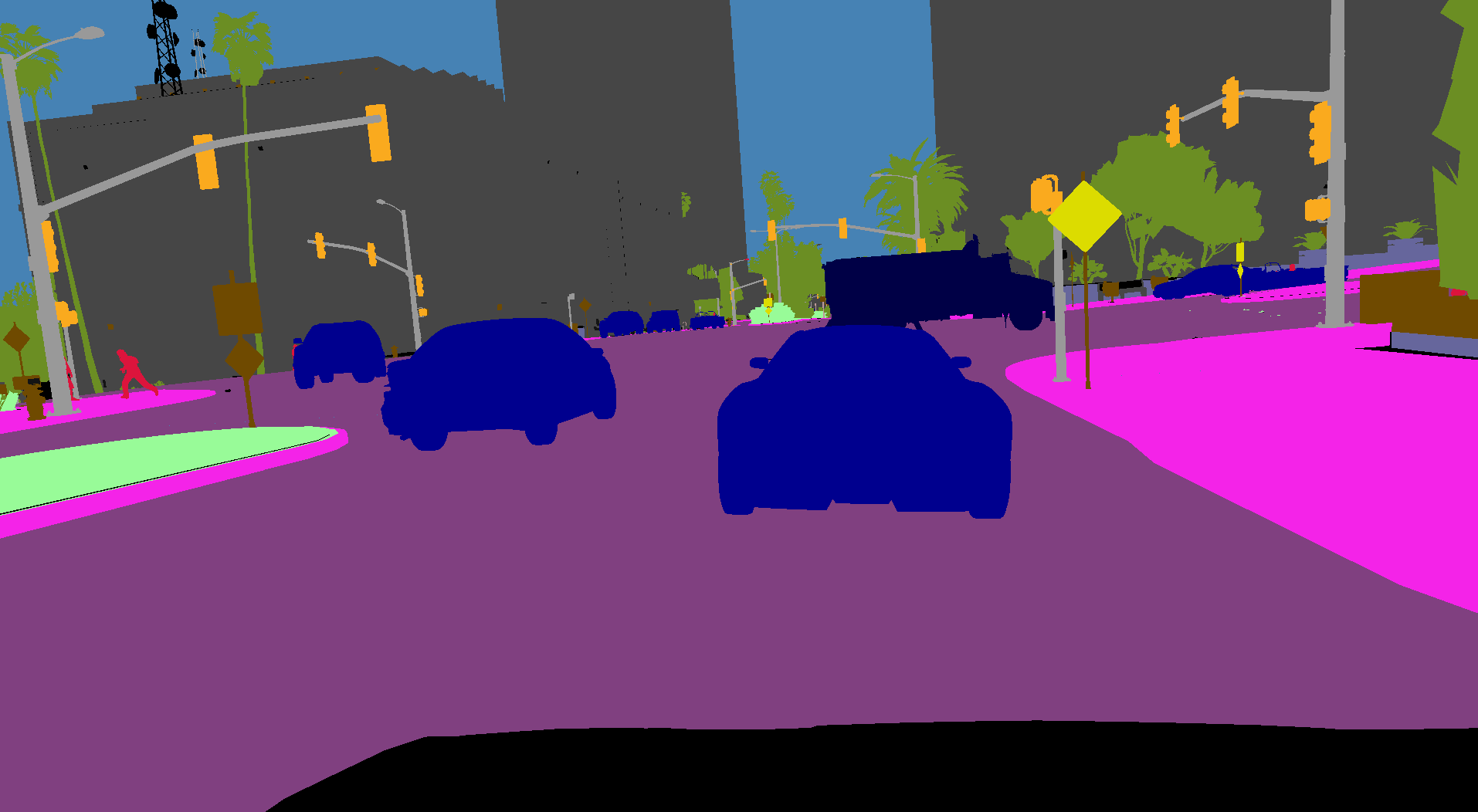} \\
	\end{tabular}
	\caption{For a rendered image (left), G-buffers represent information on geometry (\eg normal, depth), materials (\eg albedo, glossiness), and lighting (\eg atmosphere). A semantic segmentation, which can be derived from the G-buffers~\cite{Richter2016, Richter2017}, provides further high-level information on the scene.}
	\label{fig:gbuffers}
\end{figure*}

\subsection{Overview}\label{sec:architecture}
\Fig~\ref{fig:overview} provides an overview of our approach.
Our method consists of an image enhancement network, which takes as input a rendered image and outputs an enhanced image.
To facilitate the enhancement, we provide additional inputs to the network.
Specifically, we extract intermediate rendering buffers (G-buffers) from the graphics pipeline. These G-buffers provide information on the geometry, materials, and lighting in the scene.
The G-buffers are processed by a G-buffer encoder network, which outputs G-buffer features at multiple scales. This is described in \Sec~\ref{sec:gbuffers}.
The G-buffer features are then provided as input to the image enhancement network, where they are used to modulate image features.

The image enhancement network is based on HRNetV2, which demonstrated strong performance on a variety of dense prediction tasks~\cite{Wang2020} (see \Fig~\ref{fig:hrnet}).
The HRNet processes an image via multiple branches that operate at different resolutions.
Importantly, one feature stream is kept at relatively high resolution ($\tfrac{1}{4}$ of the input resolution) to preserve fine image structure.
We modify the HRNet architecture as follows.
First, we replace the initial strided convolutions by regular convolutions to have the network operate on the full resolution and preserve even finer detail.
Second, within the residual blocks in each branch we replace the batch normalization layers by rendering-aware denormalization (RAD) modules, described in \Sec~\ref{sec:gbuffers}. The modified blocks modulate the feature streams based on information extracted from the G-buffers.

The image enhancement network is trained with two objectives. First, an LPIPS loss~\cite{Zhang2018} penalizes large structural differences between the input and output images.
Second, a perceptual discriminator evaluates the realism of output images.
The discriminator, described in \Sec~\ref{sec:discriminator}, is trained to distinguish images enhanced by our network and real photographs from a target dataset.
During training, a specific sampling strategy for selecting real and synthetic image patches is used to significantly reduce artifacts that are commonly observed in prior work. This is described in \Sec~\ref{sec:structural_shift}.

\subsection{Leveraging conventional rendering pipelines}
\label{sec:gbuffers}
Many real-time rendering methods factor the rendering process into multiple passes.
A particularly popular approach is \emph{deferred shading} or \emph{deferred lighting}, which decouples visibility and shading computations by caching intermediate rendering results in image-sized G-buffers~\cite{Deering1988,Moller2018,Saito1990,Thibieroz2004}.
Although G-buffers generally contain no explicit semantic information, they are consistent with semantic entities~\cite{Richter2016, Shafaei2016}.
Since they capture geometry and material properties, using them as auxiliary input allows a network to condition the synthetic-to-real translation on the geometry, materials, and illumination of a scene.

\mypara{Extraction of G-buffers.}
We base our work on the popular game Grand Theft Auto V.
To obtain G-buffers from the game, we follow recent approaches to extracting rendering resources from computer games~\cite{Krahenbuhl2018,Richter2017,Richter2016}.
Specifically, we extract G-buffers that provide information about geometric structure (surface normals, depth), materials (shader IDs, albedo, specular intensity, glossiness, transparency), and lighting (approximate irradiance and emission, sky, bloom). Some of these are illustrated in \Fig~\ref{fig:gbuffers}.
We further augment this set with two quantities we derive from the G-buffers. First, we reflect for each pixel the view vector at the surface normal to obtain a reflection vector.
Second, we compute the dot product between the surface normal and the reflection vector.

While we extract an extensive set of G-buffers for providing comprehensive information about the scene to the generator network for best results, our experiments in \Sec~\ref{sec:ablation} confirm that substantial enhancements can be achieved even if only a subset of the buffers is available.

\mypara{G-buffer encoder.}
The G-buffers we extract mix one-hot encodings for material information, dense continuous values for normals, depth, and color, and sparse continuous information for bloom and sky buffers.
For some image regions the G-buffers are zero, depending on the objects being rendered. Sky regions, for example, contain no geometry and material information. And transparent regions and highlights are only present for a small number of objects.

\begin{figure}[thbp]
	\centering
	\includegraphics[width=85mm]{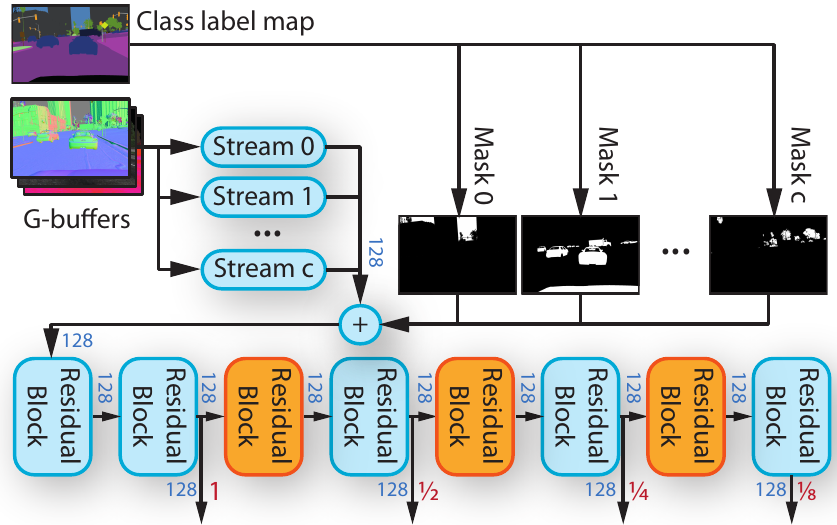}
	\caption{G-buffer encoder. The G-buffer encoder accounts for different data types and varying spatial density of the G-buffers. It processes them via multiple streams (0--c), which are fused into a joint representation in accordance with one-hot-encoded object IDs. The features are further transformed via residual blocks~(see \Fig~\ref{fig:residual}), of which the {\color{BurntOrange}orange} blocks downscale the tensors. The output scale of the feature tensors is {\color{red}red}, the channel width is {\color{Cerulean}blue}. Scales match with branches in the image enhancement network.}
	\label{fig:encoder}
\end{figure}

\begin{figure}[thbp]
	\centering
	\includegraphics[width=33.3mm]{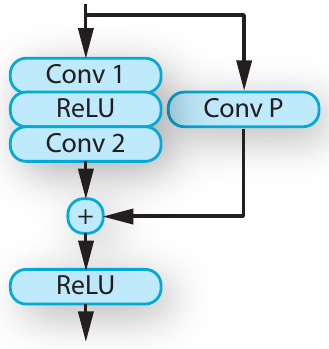}
	\caption{Residual blocks. We employ residual blocks in the G-buffer encoder and RAD modules. The blocks consist of convolutional layers (kernel size 3) with spectral normalization~\cite{Miyato2018:ICLRa} and ReLUs. Changes in channel width or downscaling are performed in \texttt{Conv 1} and \texttt{Conv P}. If channel width and resolution are constant, the projection via \texttt{Conv P} is omitted.}
	\label{fig:residual}
\end{figure}

To account for the different types of data in the G-buffers, we process the G-buffers via a G-buffer encoder~(\Fig~\ref{fig:encoder}).
The G-buffer encoder consists of multiple network streams that process the same set of G-buffers.
Each stream comprises two residual blocks, shown in \Fig~\ref{fig:residual}.

Let $f_c$ denote a feature tensor from a stream targeting object class $c$, and let $m_c$ denote a mask for objects of that class.
We then fuse the tensors via $\sum_c m_c \cdot f_c$.

The object IDs group multiple class labels from semantic segmentation maps (\eg all vehicle classes are mapped to the same ID to reduce the number of separate network streams).
This way, the streams can map information from the G-buffers differently for certain types of objects.
The fused feature tensors are further processed via residual blocks.
We extract a feature tensor before each downsampling residual block to obtain tensors at multiple scales.
The feature tensors are ingested by the image enhancement network via RAD modules.

\begin{figure}
	\centering
	\includegraphics[width=85mm]{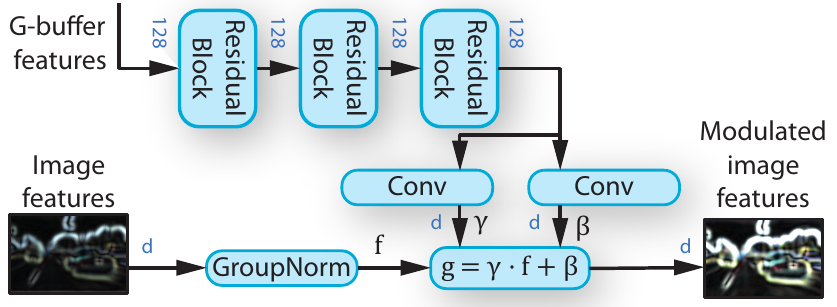}
	\caption{Rendering-aware denormalization (RAD) modulates image feature tensors via encoded geometry, material, lighting, and semantic information from a conventional rendering pipeline. The image features are normalized via group normalization, then scaled and shifted via per-element weights $\gamma, \beta$. The weights are learned and adapt to G-buffer features received from the G-buffer encoder (\Fig~\ref{fig:encoder}). To better adapt the weights, we transform the G-buffer features via three residual blocks within each RAD module.}
	\label{fig:normalization}
\end{figure}

\mypara{Rendering-Aware Denormalization (RAD).}
Our rendering-aware denormalization modules take inspiration from recent work on modulating feature tensors based on external information~\cite{Jiang2020, Karras2019, Park2019:CVPR, Zhu2020}.
In contrast to prior work, which conditions on semantic classes~\cite{Park2019:CVPR}, style features~\cite{Jiang2020,Zhu2020}, optical flow~\cite{Mallya2020}, or noise~\cite{Karras2019}, our modules learn weights from a comprehensive scene representation.
Specifically, our modules transform a feature tensor received from the G-buffer encoder network via 2 residual blocks (\Fig~\ref{fig:residual}).
The transformed features are used to learn elementwise scale and shift weights $\gamma$ and $\beta$.
The weights represent the parameters of an affine transformation of normalized image features (\Fig~\ref{fig:normalization}).

Controlled experiments reported in \Sec~\ref{sec:ablation} confirm that RAD modules deliver better results than SPADE~\cite{Park2019:CVPR} when applied to ingesting G-buffer features into the image enhancement network.

\subsection{Perceptual discriminator}
\label{sec:discriminator}
\begin{figure}
	\centering
	\includegraphics[width=85mm]{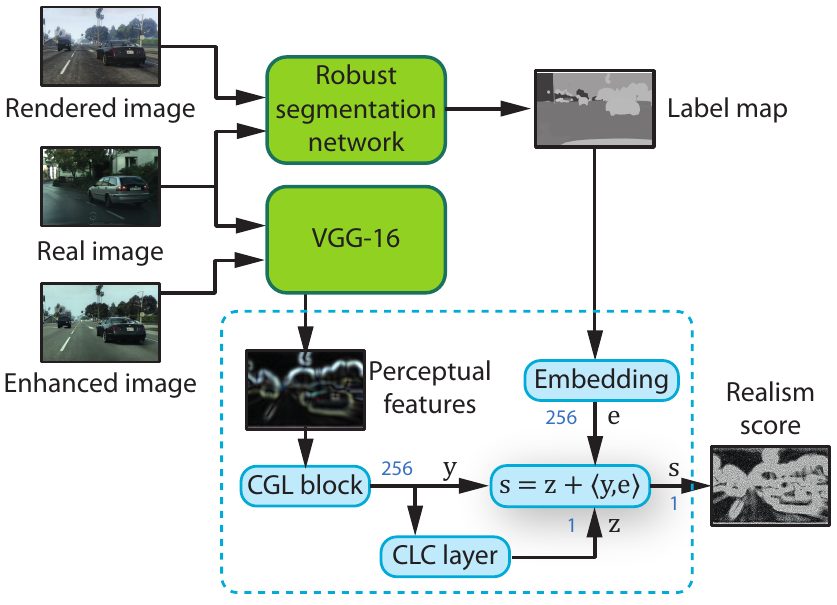}
	\caption{Perceptual discriminator. Realism of enhanced images is evaluated via a perceptual discriminator. It consists of {\color{LimeGreen}pretrained} robust segmentation (MSeg~\cite{Lambert2020}) and perceptual (VGG~\cite{SimonyanZisserman2015}) networks. These provide high-level semantic information via label maps and perceptual feature tensors. The maps and tensors are ingested by discriminator networks, which produce a realism score map. For clarity, we only show a single discriminator network, indicated by the dashed rectangle.}
	\label{fig:discriminator}
\end{figure}

During training of the image enhancement network, the realism of enhanced images is evaluated via a perceptual discriminator.
The discriminator consists of a robust semantic segmentation network, a perceptual feature extraction network, and multiple discriminator networks~(\Fig~\ref{fig:discriminator}).
We employ MSeg~\cite{Lambert2020} for semantic segmentation and VGG-16~\cite{SimonyanZisserman2015} for perceptual feature extraction.
Both networks are pretrained and are not optimized during training of the image enhancement network.
We apply the segmentation network to real images from a target dataset and unmodified rendered images.
This provides compatible semantic information for real and synthetic images.
It also enables training on datasets without ground-truth annotations.
In practice, we apply the segmentation network once to all images and cache results.

Prior work trained a segmentation network on synthetic data and applied it during training to ensure semantic consistency~\cite{Hoffman2018}. We avoid this as networks trained on synthetic data can generalize poorly to real data. Thus segmentations from such a network can differ substantially when applied to original and enhanced images. We also avoid backpropagating through the segmentation network as this may result in images that are easy to segment but not necessarily realistic.

Applying the VGG feature extraction network to real and enhanced images provides perceptual features at several levels of abstraction.
We extract feature tensors from the \texttt{relu} layers of the VGG and train a discriminator network for each level.
This way each network specializes on a different perceptual level.

The discriminator networks (\Fig~\ref{fig:discriminator}) each consist of a stack of five Convolution-GroupNorm-LeakyReLU (CGL) layers, which produces a 256-dimensional feature tensor $y$, and a Convolution-LeakyReLU-Convolution (CLC) layer, which projects the feature tensor down to a single-channel map $z$. The feature tensor $y$ is further fused with an embedding tensor $e$ via an inner product. The embedding tensor contains a 256-dimensional embedding per pixel, learned from the label maps discussed above. The inner product of features and embeddings was used in prior work~\cite{Miyato2018:ICLRb,Brock2019,Liu2019}.

Further details on the architecture of the discriminator networks are provided in the supplementary material.

\subsection{Layout differences cause artifacts}
\label{sec:structural_shift}
In an adversarial setup as we employ it, a discriminator is trained to classify images, assigning a \emph{real} or \emph{fake} label to each image or pixel.
During training, the discriminator will pick up any feature that allows it to easily discriminate real and fake images.
For example, if sensor noise is present in real images, but not in synthetic ones, the discriminator will quickly learn to correctly label a noisy patch as \emph{real}.
Backpropagating the gradient from the discriminator to the generator encourages the generator to add noise to synthetic images, making them appear more realistic.

\begin{figure}[thb]
	\centering
	\renewcommand{\arraystretch}{0.5}
	\begin{tabular}{@{}c@{\hspace{0.4mm}}c*{3}{@{\hspace{0.2mm}}c}@{}}
		&\myfootnotesize sky & \myfootnotesize vegetation & \myfootnotesize building & \myfootnotesize road \\
		\raisebox{0.05\linewidth}[0pt][0pt]{\rotatebox[origin=c]{90}{\footnotesize GTA}} &
		\includegraphics[width=2.1cm]{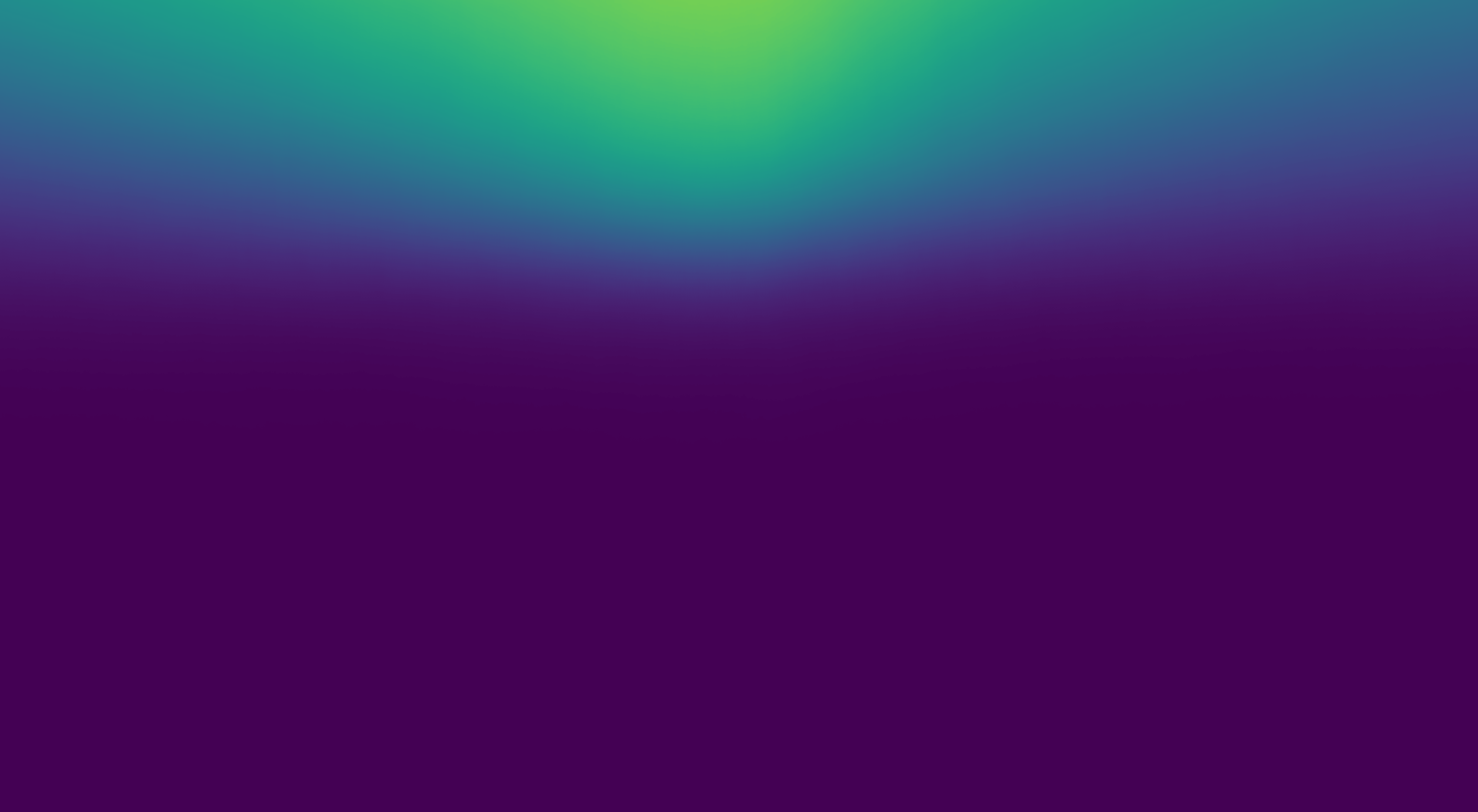}&
		\includegraphics[width=2.1cm]{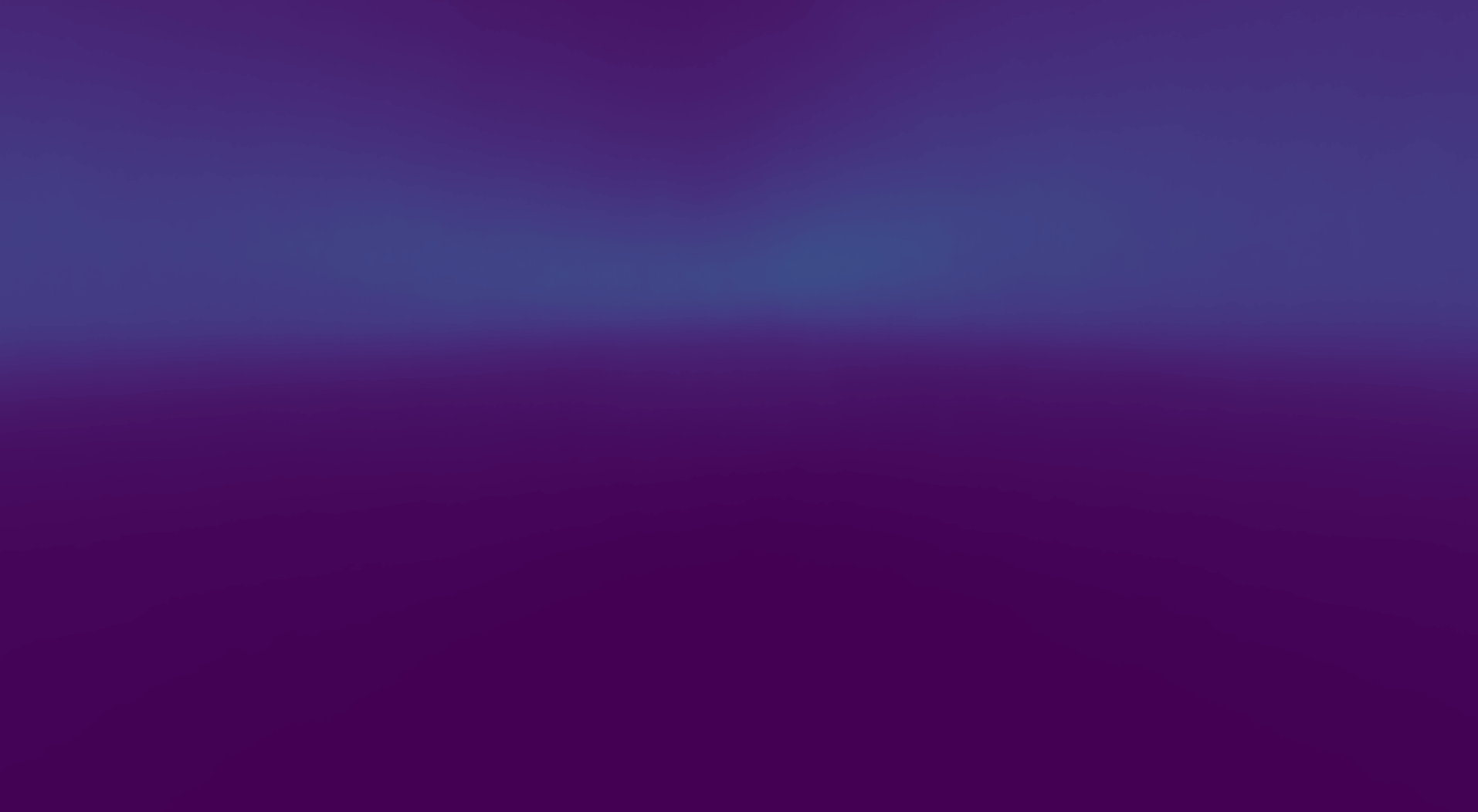}&
		\includegraphics[width=2.1cm]{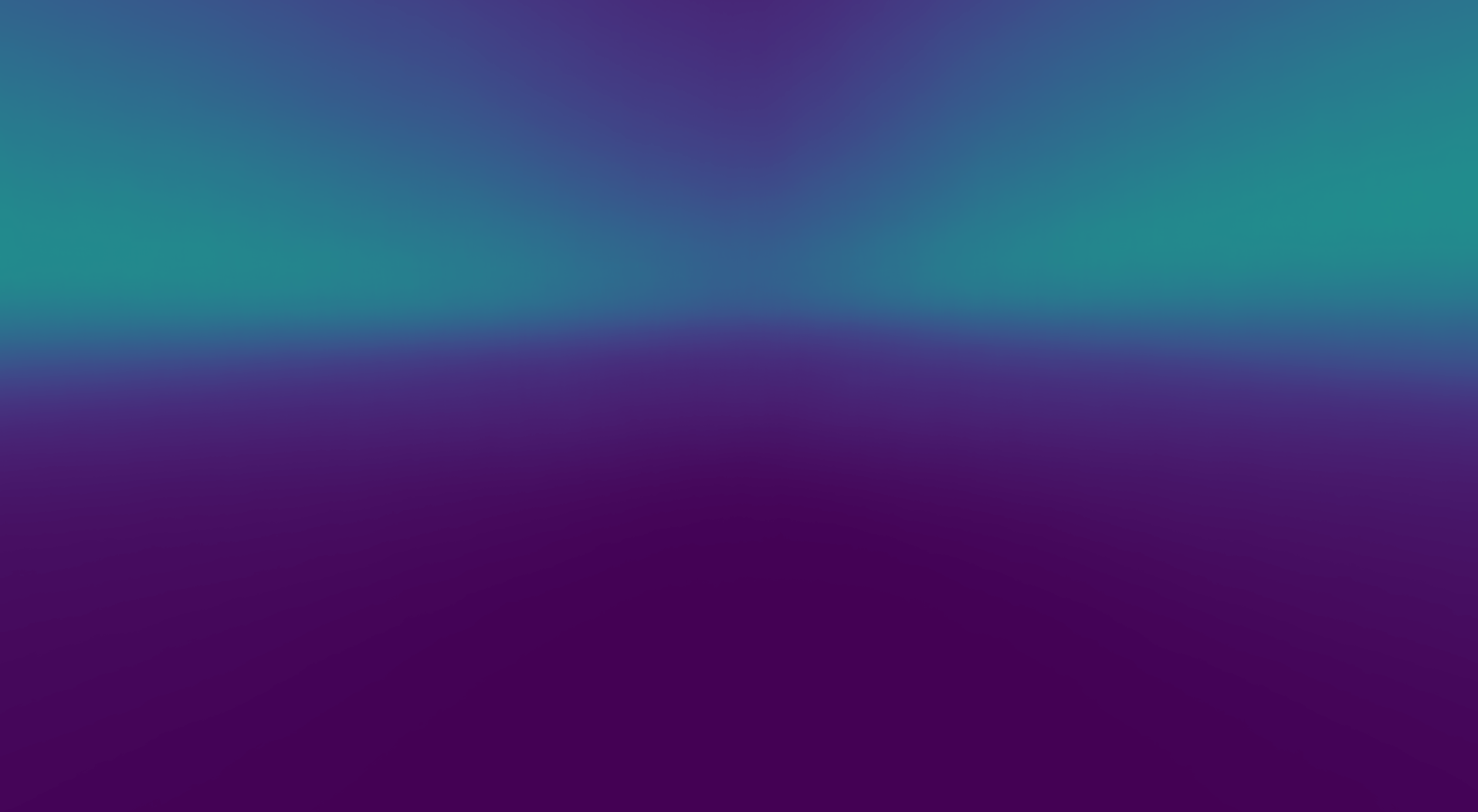}&
		\includegraphics[width=2.1cm]{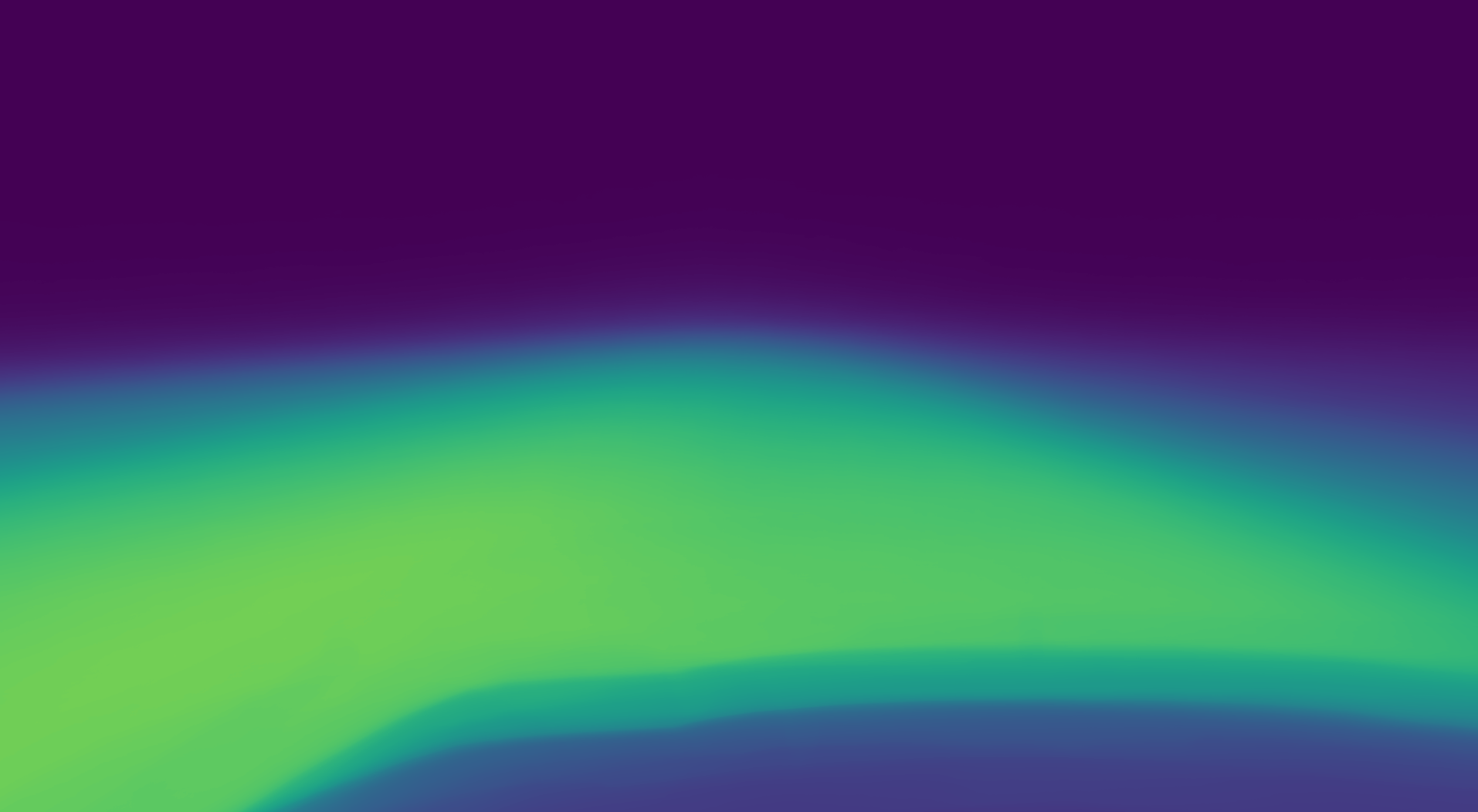}\\
		\raisebox{0.05\linewidth}[0pt][0pt]{\rotatebox[origin=c]{90}{\footnotesize CS}} &
		\includegraphics[width=2.1cm]{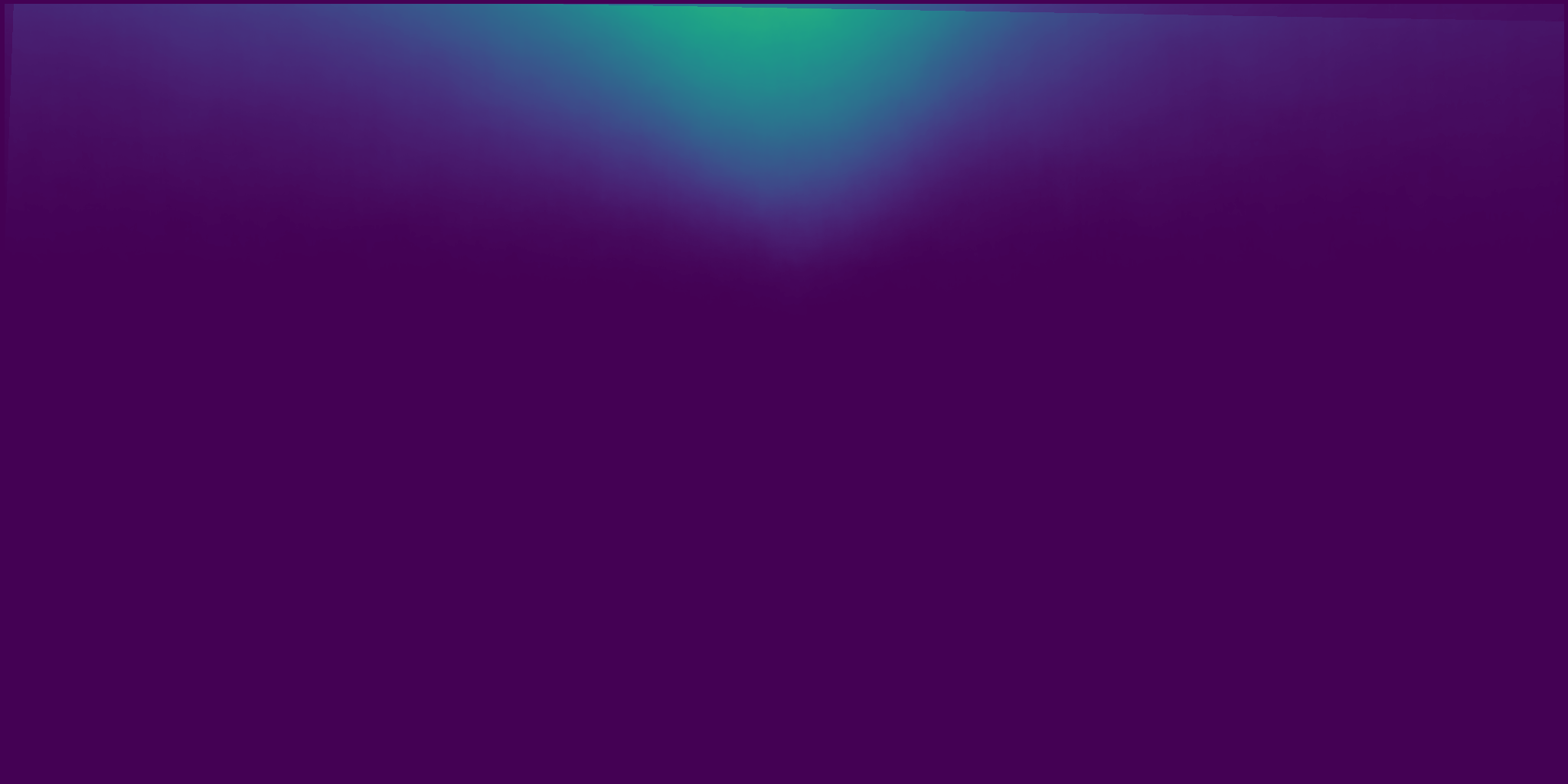}&
		\includegraphics[width=2.1cm]{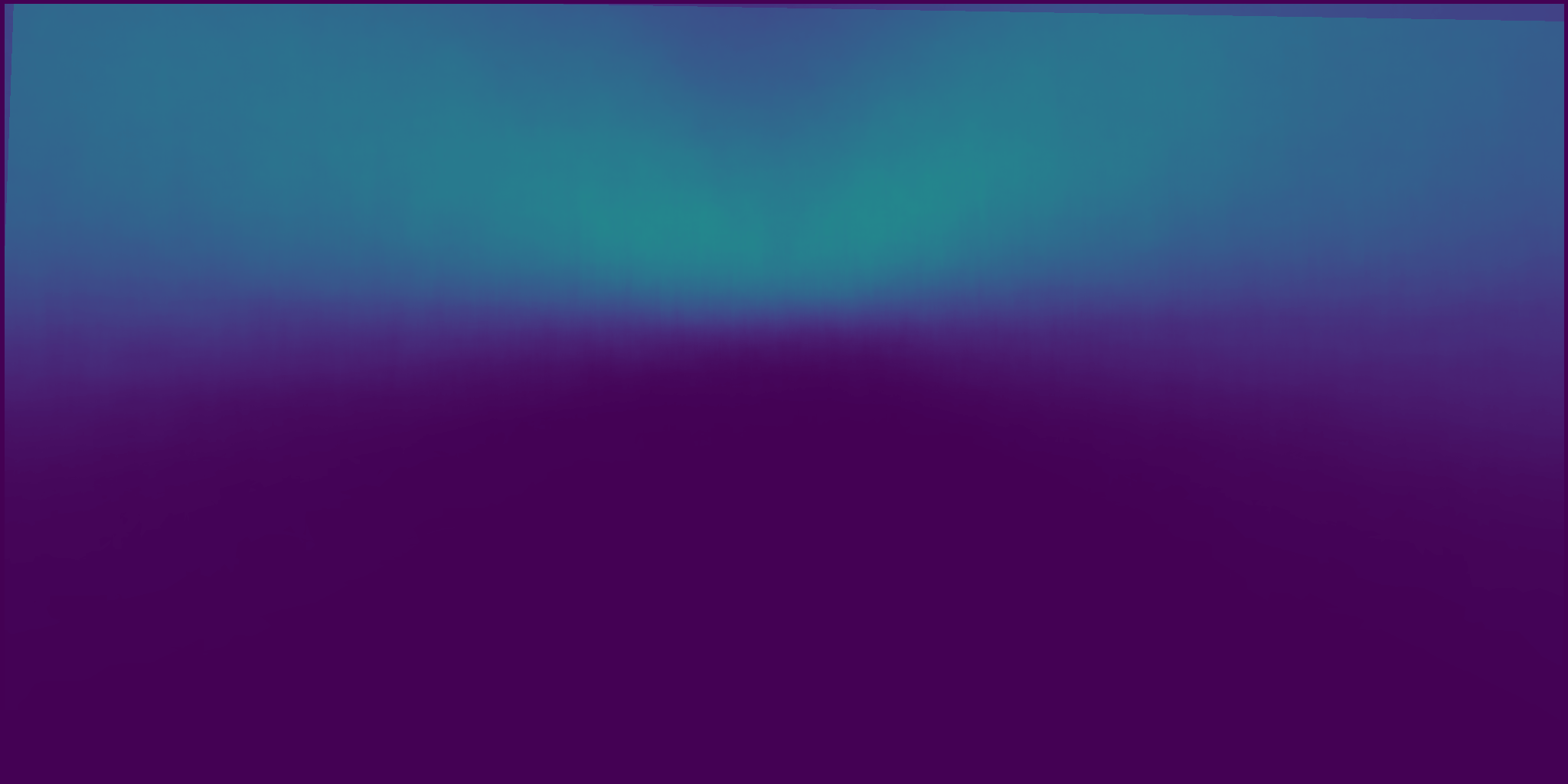} &
		\includegraphics[width=2.1cm]{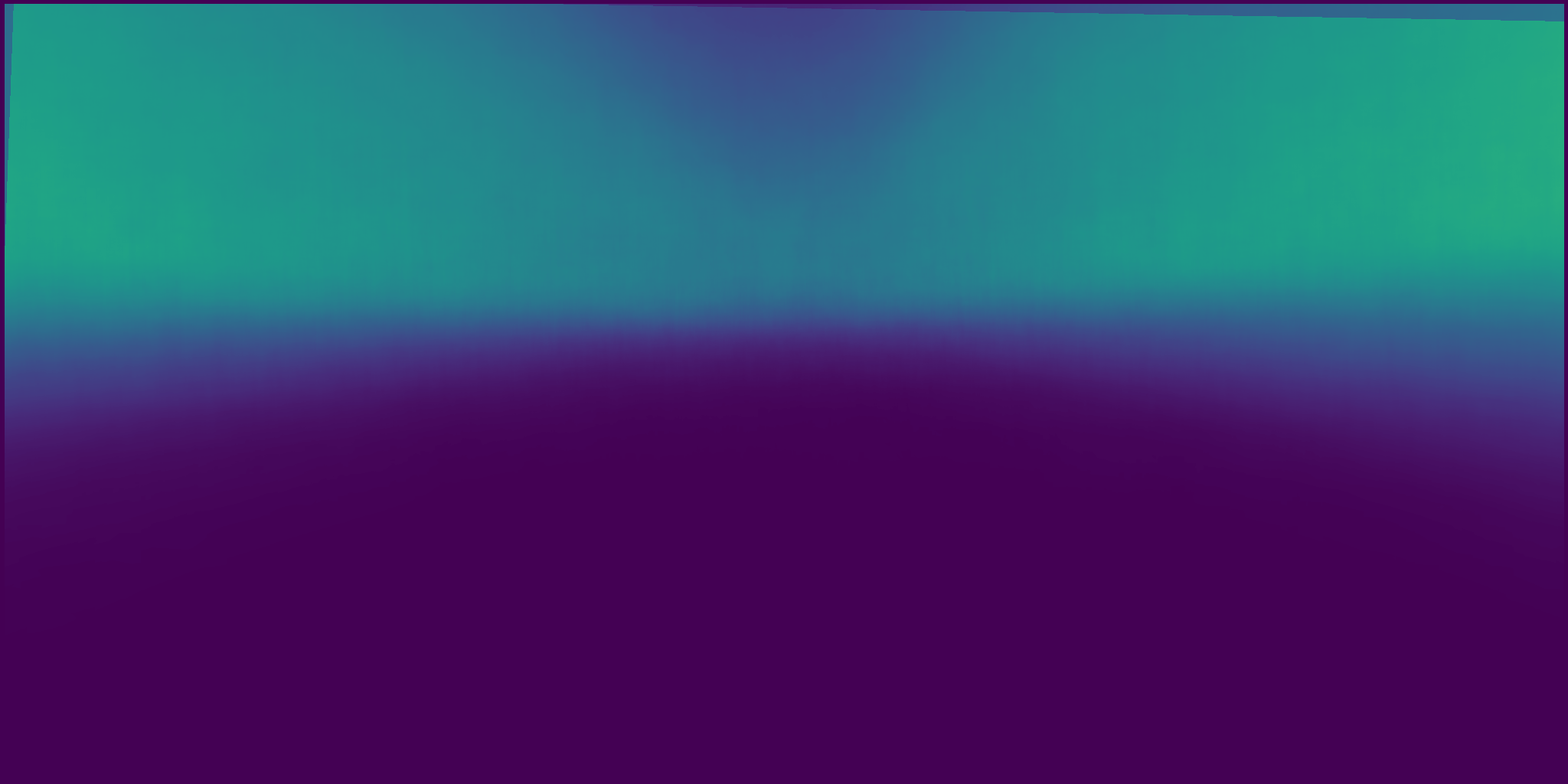}&
		\includegraphics[width=2.1cm]{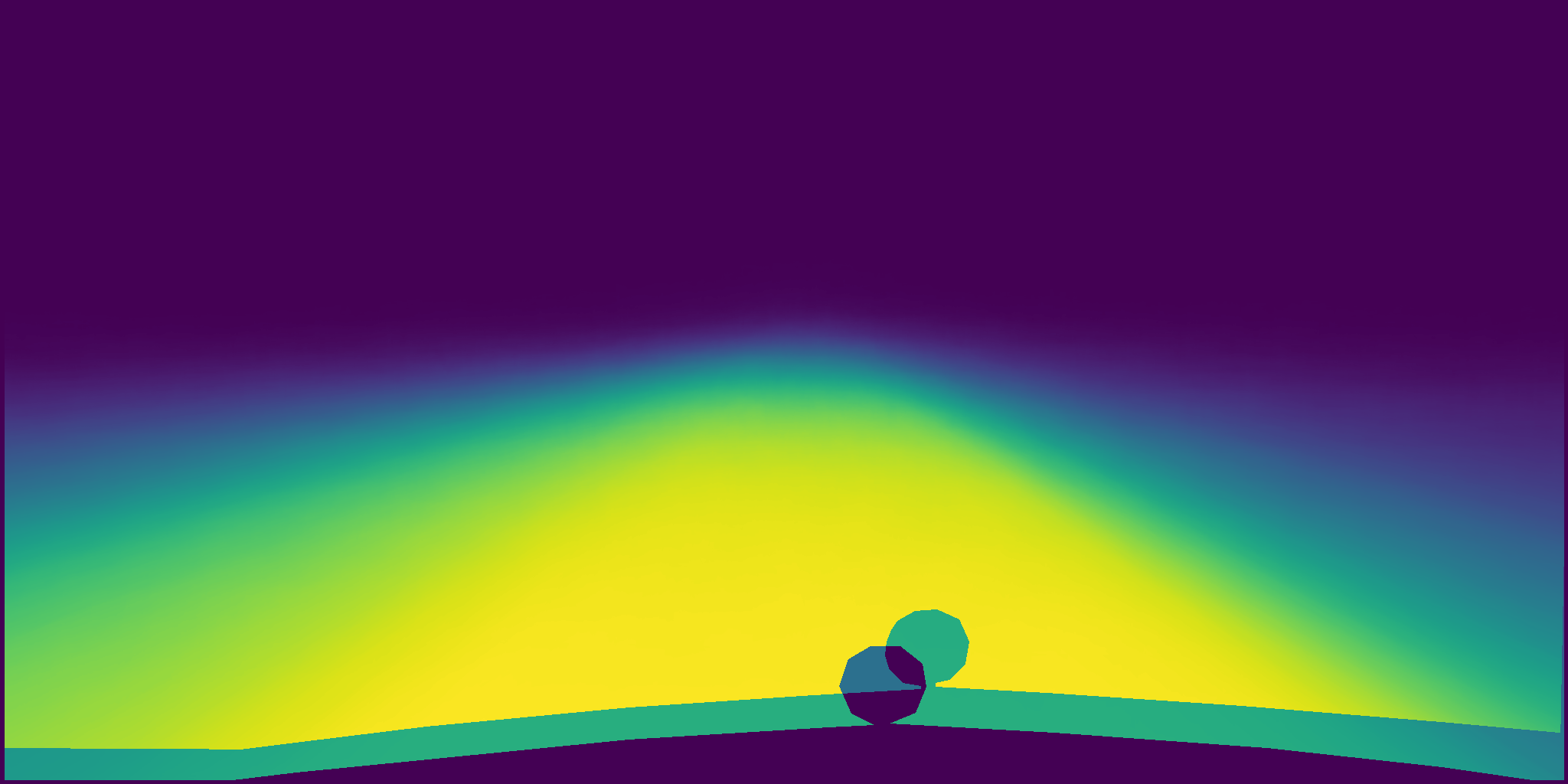}\\
	\end{tabular}
	\caption{Scene layouts are different in GTA (top) and Cityscapes (bottom). This affects the probability of seeing certain classes at specific positions. For example, GTA is more likely to have sky at the top of the image, while Cityscapes is more likely to have trees there and a star logo on the hood of the car. \Fig~\ref{fig:gan_artifacts} illustrates the impact of this structural mismatch on photorealism enhancement.}
	\label{fig:class_distribution}
\end{figure}

A problem arises when fake and real images can be distinguished by relying on spurious features.
For example, the probability of seeing sky at the top of an image from GTA V is much higher than for Cityscapes, as can be seen in the probability density maps in \Fig~\ref{fig:class_distribution}.
Conversely, at the same position it is much more likely to find trees in Cityscapes than in GTA.
Thus, a classifier trained on uniformly sampled images from both datasets may easily identify (real) images from Cityscapes by checking the top of the image.
If the top contains some texture resembling trees, it is more likely to be real.
Putting this discriminator to work in an adversarial training setup will push the generator to place trees in the sky.

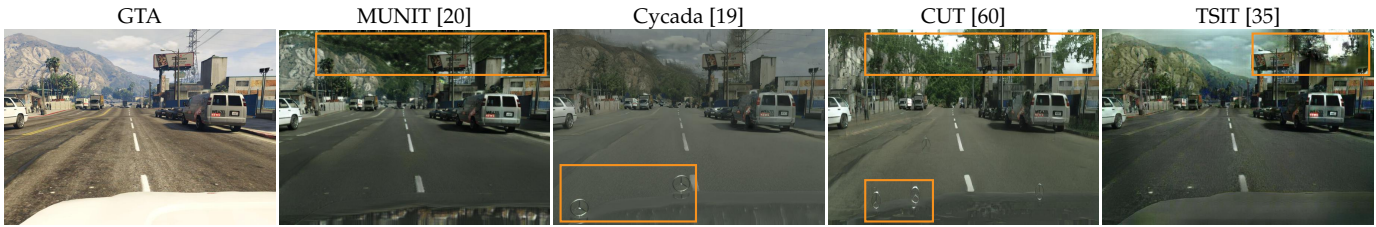
\begin{figure*}[tb]
	\centering
	\renewcommand{\arraystretch}{0.75}
	\begin{tabular}{@{}c*{4}{@{\hspace{0.5mm}}c}@{}}
		\myfootnotesize GTA & \myfootnotesize MUNIT~\cite{Huang2018} & \myfootnotesize Cycada~\cite{Hoffman2018} & \myfootnotesize CUT~\cite{Park2020} & \myfootnotesize TSIT~\cite{Jiang2020}\\
		\includegraphics[width=3.6cm,trim={6cm 0cm 2cm 0cm},clip]{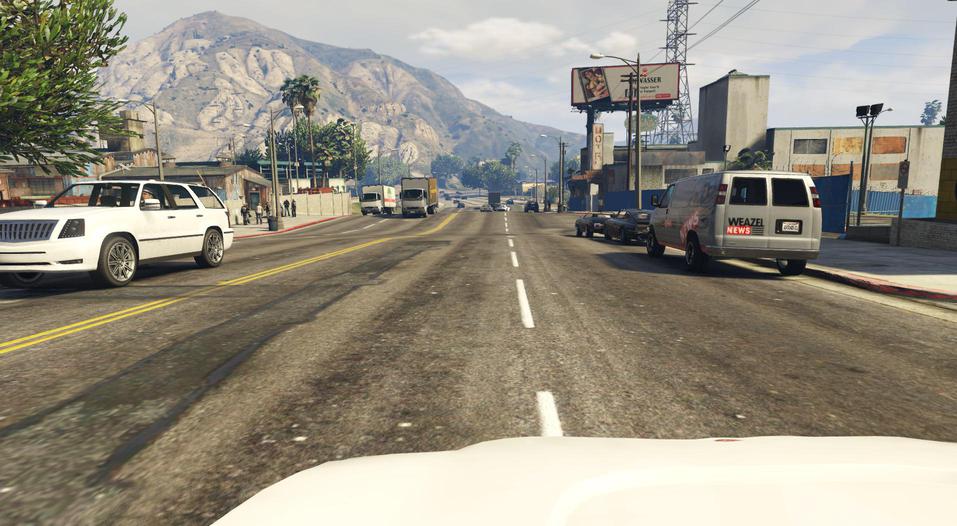}&
		\begin{tikzpicture}
			\node[anchor=south west,inner sep=0] (image) at (0,0) {\includegraphics[width=3.6cm,trim={6cm 0cm 2cm 0cm},clip]{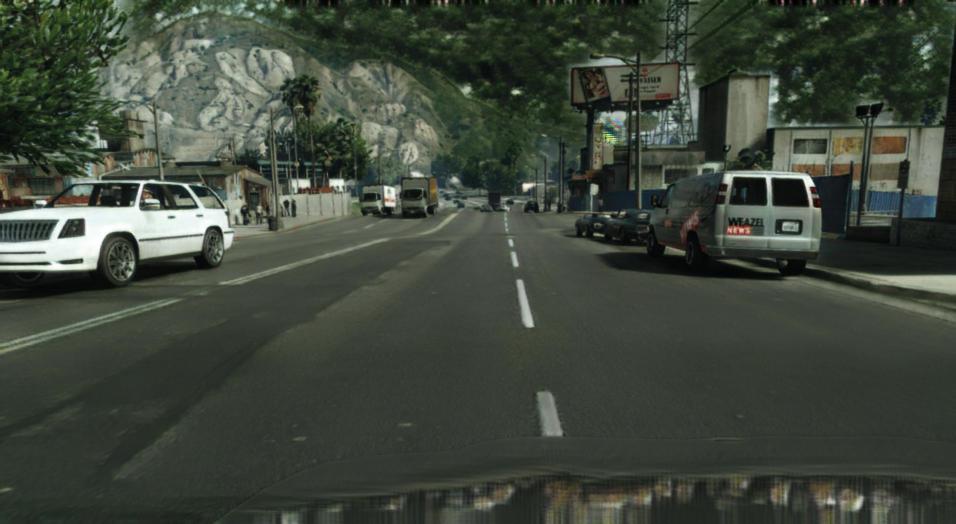}};
			\draw[BurntOrange,thick] (0.5, 2) rectangle (3.55, 2.55);
		    \end{tikzpicture}&
		\begin{tikzpicture}
			\node[anchor=south west,inner sep=0] (image) at (0,0) {\includegraphics[width=3.6cm,trim={6cm 0cm 2cm 0cm},clip]{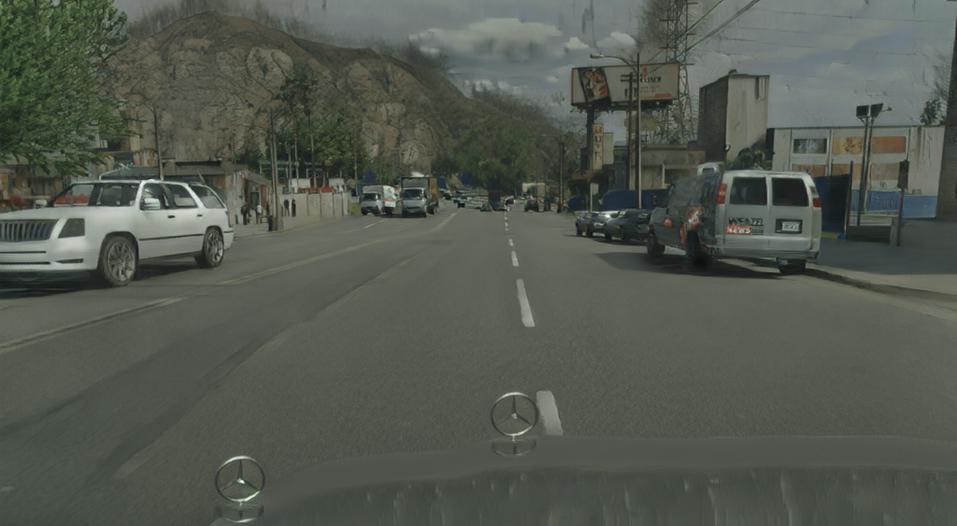}};
			\draw[BurntOrange,thick] (0.1, 0.05) rectangle (1.9, 0.8);
		    \end{tikzpicture}&
		\begin{tikzpicture}
			\node[anchor=south west,inner sep=0] (image) at (0,0) {\includegraphics[width=3.6cm,trim={6cm 0cm 2cm 0cm},clip]{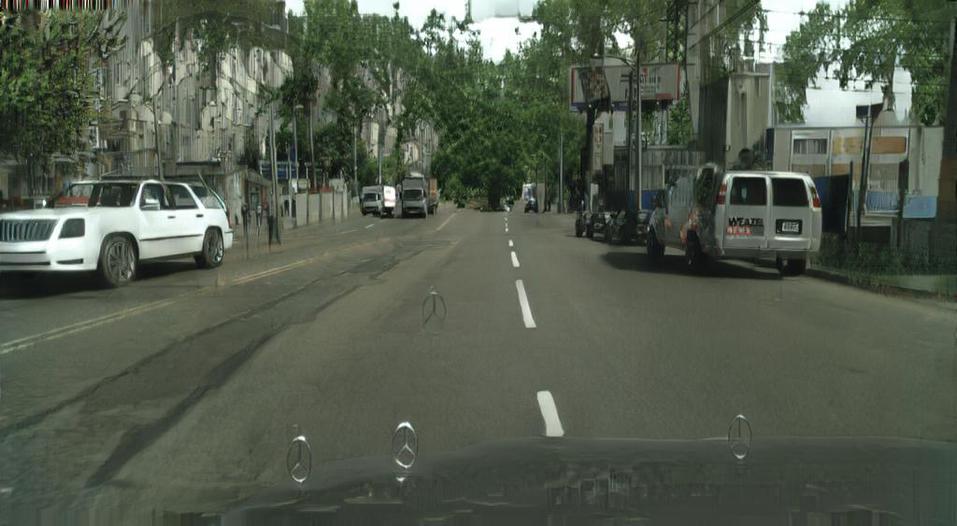}};
			\draw[BurntOrange,thick] (0.5, 2) rectangle (3.55, 2.55);
			\draw[BurntOrange,thick] (0.5, 0.05) rectangle (1.4, 0.6);
		    \end{tikzpicture}&
		\begin{tikzpicture}
			\node[anchor=south west,inner sep=0] (image) at (0,0) {\includegraphics[width=3.6cm,trim={6cm 0cm 2cm 0cm},clip]{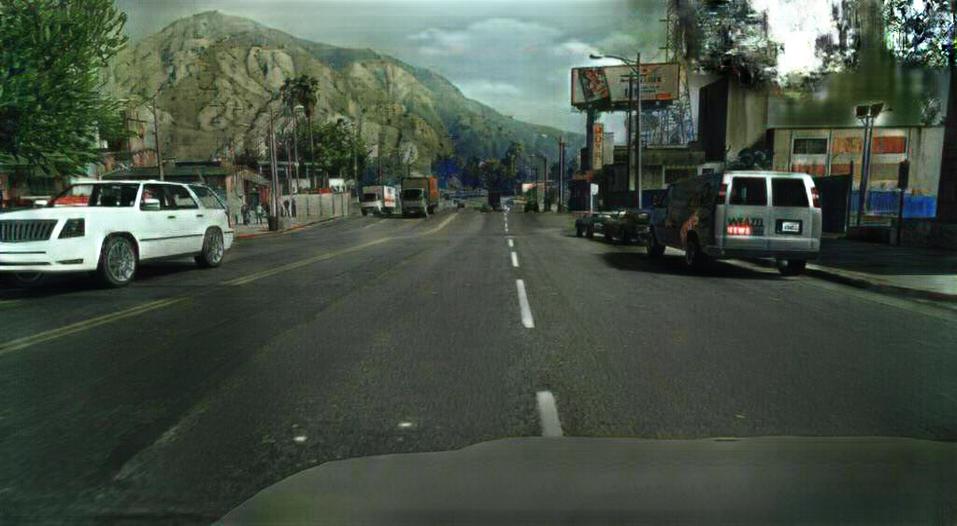}};
			\draw[BurntOrange,thick] (2, 2) rectangle (3.55, 2.55);
		    \end{tikzpicture}\\
	\end{tabular}
	\caption{When transforming images from GTA (left) to match the style of Cityscapes, GANs commonly hallucinate objects such as trees in the sky or star logos at the bottom (remaining columns, highlighted).}
	\label{fig:gan_artifacts}
\end{figure*}

Confirming evidence for this hypothesis comes from recent methods that employ GANs to translate GTA V to Cityscapes -- they indeed hallucinate trees at the top or star logos at the bottom of images, as shown in \Fig~\ref{fig:gan_artifacts}.

\mypara{Sampling matching patches.}
Our analysis suggests that randomly sampled images from GTA and Cityscapes differ in their layout in expectation, although they may contain the same type and quantity of objects~\cite{Richter2017}.
This suggests that the standard strategy of comparing image patches that are as large as possible to maximize context~\cite{Hoffman2018, Huang2018, Jia2020, Park2020} is suboptimal for enhancing realism.
We propose a different sampling strategy.

First, we shrink the crop size to an area of only about $7\%$ of the full image.
This contrasts with prior work which ingests $30$--$50\%$ of an image as a single patch.
Experiments in \Sec~\ref{sec:ablation} confirm that this simple modification already improves results significantly.

Second, we match sampled patches across datasets to balance the distribution of objects presented to the discriminator.
Specifically, we process patches from synthetic images with a VGG network~\cite{SimonyanZisserman2015}, pretrained on ImageNet~\cite{Deng2009}, and extract feature tensors at the last \texttt{relu} layer.
We crop patches at a width of 196 pixels, which corresponds to the receptive field of VGG at this layer. We thus obtain a $1 \times 1 \times 512$ dimensional feature vector per patch.
Let $\phi(p_i)$ denote the feature vector computed from patch $p_i$.
We consider two patches as matching if they have a cosine similarity above 0.5:
\begin{align}
	\mathcal{P}_{\mathit{match}}(p_i) = \left\{p_j \in \mathcal{P}_{\mathit{real}} \Bigg| \frac{\phi(p_i)\cdot\phi(p_j)}{\lVert\phi(p_i)\rVert \lVert\phi(p_j)\rVert} > 0.5 \right\},
\end{align}
where $\mathcal{P}_{\mathit{real}}$ is the set of patches extracted from real images.

\begin{figure}[tb]
	\centering
	\renewcommand{\arraystretch}{0.75}
	\begin{tabular}{@{}c*{5}{@{\hspace{1mm}}c}@{}}
		\raisebox{0.085\linewidth}[0pt][0pt]{\rotatebox[origin=c]{90}{\footnotesize GTA}} &
		\includegraphics[width=0.18\linewidth]{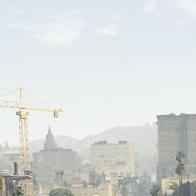}&
		\includegraphics[width=0.18\linewidth]{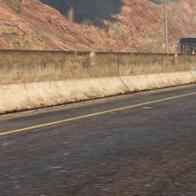}&
		\includegraphics[width=0.18\linewidth]{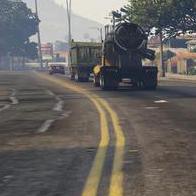}&
		\includegraphics[width=0.18\linewidth]{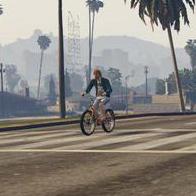}&
		\includegraphics[width=0.18\linewidth]{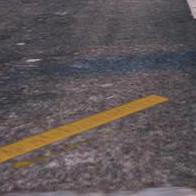}\\[0pt]
		\raisebox{0.085\linewidth}[0pt][0pt]{\rotatebox[origin=c]{90}{\footnotesize Cityscapes}} &
		\includegraphics[width=0.18\linewidth]{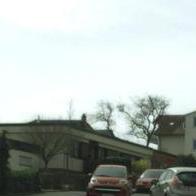}&
		\includegraphics[width=0.18\linewidth]{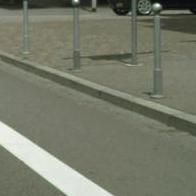}&
		\includegraphics[width=0.18\linewidth]{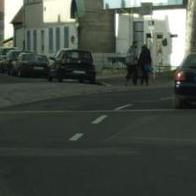}&
		\includegraphics[width=0.18\linewidth]{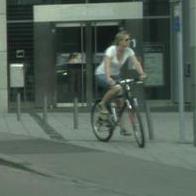}&
		\includegraphics[width=0.18\linewidth]{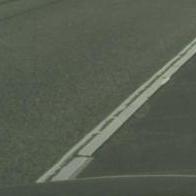}\\
	\end{tabular}
	\caption{Matching patches across datasets. We show patches sampled from GTA (top) and corresponding patches from Cityscapes (bottom).}
	\label{fig:matched_patches}
\end{figure}

For efficiency, we unit-normalize $\phi(p_i)$ and compute the $L_2$ distance via FAISS~\cite{Johnson2017}.
Examples of matching patches across GTA and Cityscapes are shown in \Fig~\ref{fig:matched_patches}.

\subsection{Implementation \& training details}
\label{sec:implementation}
We train the generator and discriminator with an $L_2$ loss. The generator is further regularized with an LPIPS loss (${\lambda = 5}$)~\cite{Zhang2018}.
We train all networks with Adam~\cite{Kingma2015} (${\beta_1 = 0.9}$, ${\beta_2 = 0.999}$, weight decay $0.0001$). The learning rate is set to 0.0001 and halved every 100K iterations. We clip gradient magnitudes greater than 1000 for all networks. All discriminators are regularized with a gradient penalty on real data ($\gamma = 0.06$)~\cite{Mescheder2018}. We train with batch size 1 for 1M iterations (controlled experiments 600K).

Adversarial training can become unstable if discriminators become too strong and cease to provide meaningful gradients to the generator~\cite{Arjovsky2017,Karras2020}.
In addition, we have encountered training difficulties that stem from the use of multiple discriminators. For example, we have observed that discriminators that operate on low-level VGG features tend to learn faster, and are thus prone to dominating the training and overfitting.
To stabilize training and prevent overfitting of individual discriminator networks, we throttle the training speed of each discriminator by randomly skipping the backward pass for this discriminator with a certain probability.
The probability of skipping the backward pass is determined by the current performance of the discriminator.
Intuitively, the better a discriminator becomes at judging realism, the closer it is to overfitting, and the less frequently it will be updated by our training protocol.
This strategy keeps all discriminators roughly at the same level of performance as training progresses.
A similar mechanism was used in ABC-GAN~\cite{Susmelj2017} to balance generator and discriminator training.
We use this idea to balance multiple discriminators.
Our computation of the probability of skipping the backward pass for a given discriminator is inspired by the augmentation probability of Karras~\etal~\cite{Karras2020}. (Different from Karras~\etal, we take into account both real and fake samples in computing the update heuristic.)

\section{Evaluation}\label{sec:evaluation}
\subsection{Metrics}\label{sec:metrics}
A number of metrics for evaluating the realism of generated images have been proposed. The most common are the Inception Score (IS)~\cite{Salimans2016}, the Fr{\'e}chet Inception Distance (FID)~\cite{Heusel2017}, and the Kernel Inception Distance (KID)~\cite{Binkowski2018}. Among these, the KID has been shown to be superior~\cite{Binkowski2018}, and we use it in our evaluation for this reason.

However, our analysis in \Sec~\ref{sec:structural_shift} on the structural shift across datasets implies that quality assessment using the KID may be misleading due to mismatched scene layouts.
The KID compares features extracted from the \texttt{pool3} layer of an inception network~\cite{Szegedy2016}, which corresponds to high-level semantic concepts.
Thus, roughly speaking, the KID measures distance between semantic structure, but not necessarily a difference in perceived realism.
This is problematic, since in enhancing the photorealism of synthetic images we aim to retain the scene structure and semantic content of the source image, rather than shift them towards scene structures that may be more common in the real-world dataset.
Put another way, we can trivially minimize the KID by reproducing a real image from the target dataset and ignoring rendered images altogether.
Thus, preserving semantic content poses a lower bound on the KID, a level below which the KID \emph{should not} be driven.
Overall, the KID objective is misaligned with the broader photorealism enhancement objective and is a problematic metric for this reason.

\begin{figure}[tb]
	\centering
	\renewcommand{\arraystretch}{0.75}
	\begin{tabular}{@{}c*{5}{@{\hspace{1mm}}c}@{}}
		\raisebox{0.085\linewidth}[0pt][0pt]{\rotatebox[origin=c]{90}{\footnotesize GTA}} &
		\includegraphics[width=0.18\linewidth]{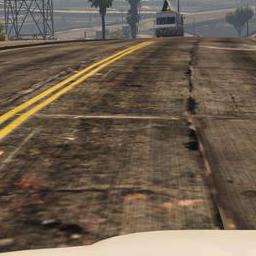}&
		\includegraphics[width=0.18\linewidth]{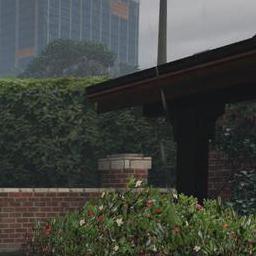}&
		\includegraphics[width=0.18\linewidth]{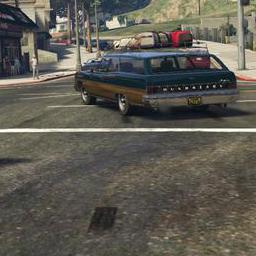}&
		\includegraphics[width=0.18\linewidth]{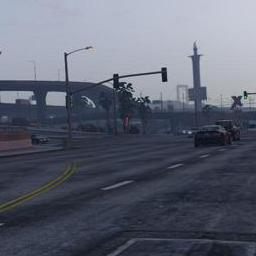}&
		\includegraphics[width=0.18\linewidth]{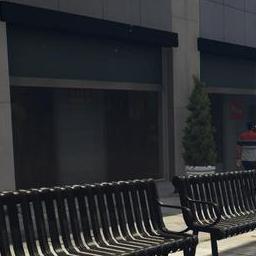}\\[0pt]
		\raisebox{0.085\linewidth}[0pt][0pt]{\rotatebox[origin=c]{90}{\footnotesize Cityscapes}} &
		\includegraphics[width=0.18\linewidth]{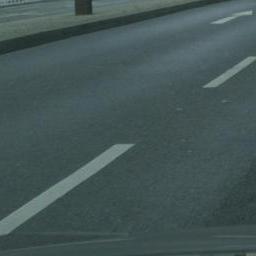}&
		\includegraphics[width=0.18\linewidth]{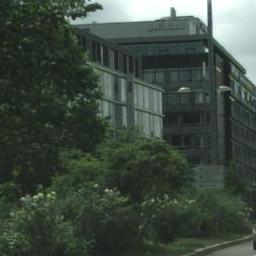}&
		\includegraphics[width=0.18\linewidth]{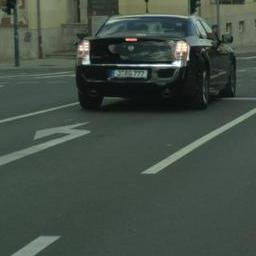}&
		\includegraphics[width=0.18\linewidth]{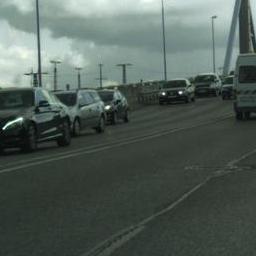}&
		\includegraphics[width=0.18\linewidth]{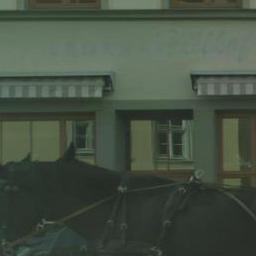}\\
	\end{tabular}
	\caption{Semantically corresponding patches sampled for calculating our sKVD metrics.}
	\label{fig:kvd}
\end{figure}

We propose a different set of metrics that alleviate this problem. Our metrics build on the KID, but incorporate some key changes.
In order to better assess the difference between images at several perceptual levels, we replace the features from the inception network with features extracted at different layers of VGG, since this architecture has been widely used for assessing perceptual image quality~\cite{ChenKoltun2017,Zhang2018}.
To address the problem of mismatched layouts in the source and target datasets, we align the distribution of patches for which we extract features.
Specifically, we extract quadratic patches of $\tfrac{1}{8}$ of the image size from the semantic label maps of source and target datasets.
We downsample these patches to a resolution of $16 \times 16$ to obtain a 256-dimensional vector.
The vectors obtained in this way correspond to ground-truth semantic descriptions of the patches.
For each such vector from the synthetic dataset, we find the nearest neighbor in the set of vectors from the real dataset. We retain pairs of vectors with more than $50\%$ matching entries.
This way, we obtain a set of semantically corresponding patches from the two datasets (\Fig~\ref{fig:kvd}).

More formally, let $\sigma$ denote the vector encoding of a patch and $[\cdot]$ the Iverson bracket. We sample the nearest neighbor patch for $p_i$ by
\begin{align}
	\mathit{NN}(p_i) = \argmax_{p_j} \sum_k \left[\sigma_k(p_i) = \sigma_k(p_j)\right],
\end{align}
where $k$ iterates over elements in the vector encoding.

Following the construction of KID~\cite{Binkowski2018}, we define our metric as the squared maximum mean discrepancy (MMD) between features from a VGG-16, computed on corresponding patches.\footnote{To compute the MMD, we use the same polynomial kernel as Binkowski~\etal~\cite{Binkowski2018}.}
The different feature representations of the VGG give rise to a family of metrics, characterized by the layer at which feature maps are extracted.
We extract features at \texttt{relu1-2}, \texttt{relu2-2}, \texttt{relu3-3}, \texttt{relu4-3}, \texttt{relu5-3}, and term the corresponding metrics $\skvd_{*}$ for \emph{semantically aligned Kernel VGG Distance} with a subscript indicating the corresponding VGG \texttt{relu} layer.
\begin{figure*}[t]
	\centering
	\newcolumntype{P}[1]{>{\centering\arraybackslash}p{#1}}
	\renewcommand{\arraystretch}{0.5}
	\begin{tabular}{@{}P{7.5cm}@{}P{3cm}@{}P{7.5cm}@{}}
		\footnotesize GTA & & \footnotesize Ours \\
		\begin{minipage}{7.5cm}
			\begin{tikzpicture}
			\node[anchor=north west,inner sep=0] (image) at (0,0) {\includegraphics[width=7.5cm]{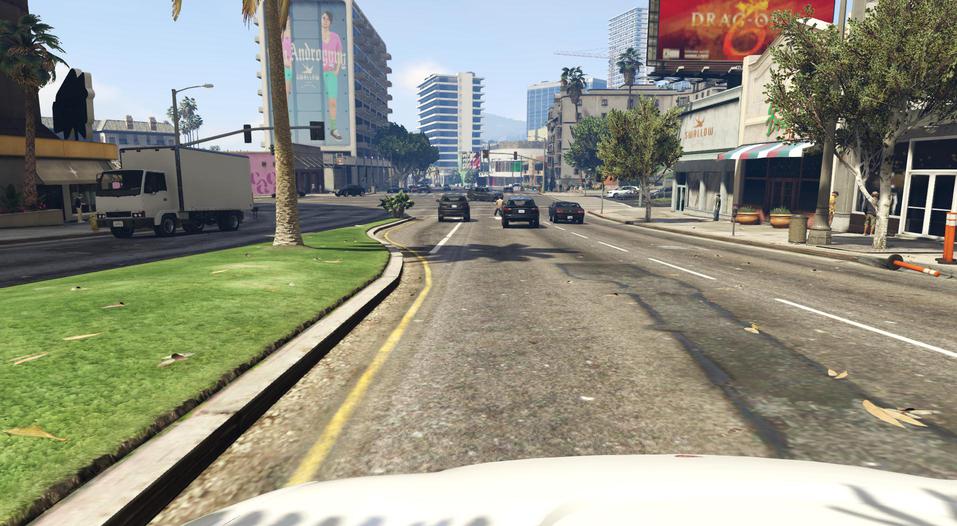}};
			\begin{scope}[x={(image.north east)},y={(image.south west)}]
				\draw[BurntOrange,thick] (0.38, 0.76) rectangle (0.58, 1);
		    \end{scope}
			\end{tikzpicture}
		\end{minipage}
		\vspace{0.1mm}
		&
		\begin{minipage}{3cm}
		\includegraphics[width=3cm,trim={13cm 0cm 14cm 13.91cm},clip,cfbox=BurntOrange 0.5mm -0.5mm]{figures/comparison/gta/05137.jpg}
		\includegraphics[width=3cm,trim={13cm 0cm 14cm 13.91cm},clip,cfbox=Cerulean 0.5mm -0.5mm]{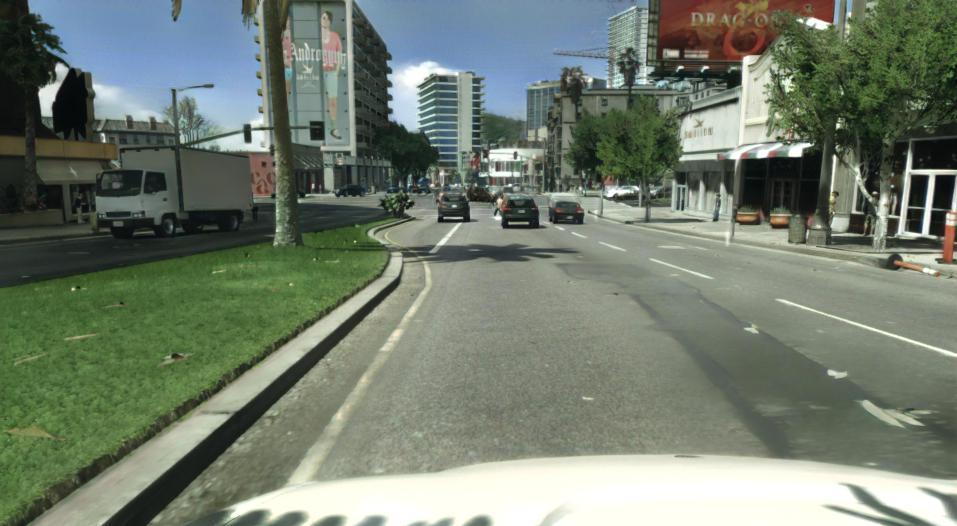}
		\end{minipage}
		\vspace{0.1mm}
		&
		\begin{minipage}{7.5cm}
			\begin{tikzpicture}
				\node[anchor=north west,inner sep=0] (image) at (0,0) {\includegraphics[width=7.5cm]{figures/comparison/ours/05137.jpg}};
				\begin{scope}[x={(image.north east)},y={(image.south west)}]
					\draw[Cerulean,thick] (0.38, 0.76) rectangle (0.58, 1);
			    \end{scope}
				\end{tikzpicture}
		\end{minipage}
		\vspace{0.1mm}\\
		\footnotesize Color transfer & & \footnotesize WCT2 \\
		\begin{minipage}{7.5cm}
			\begin{tikzpicture}
			\node[anchor=north west,inner sep=0] (image) at (0,0) {\includegraphics[width=7.5cm]{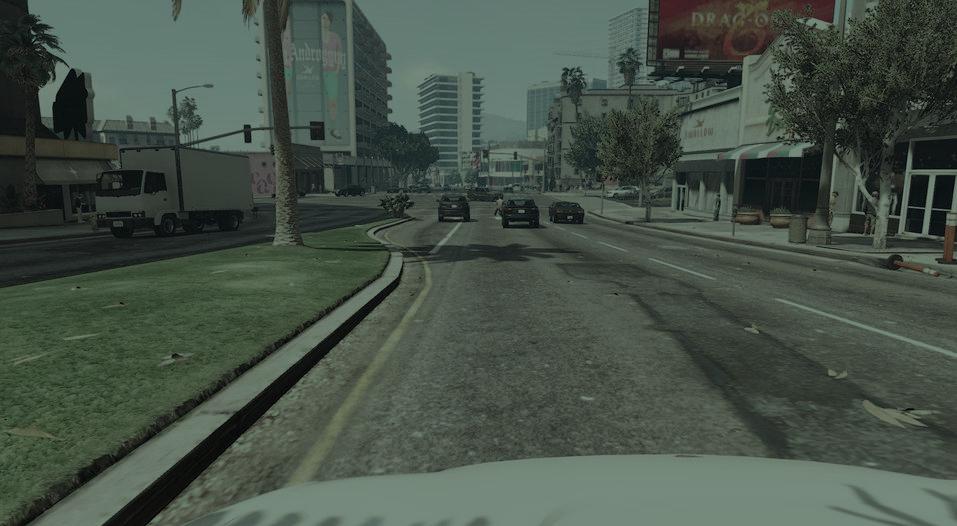}};
			\begin{scope}[x={(image.north east)},y={(image.south west)}]
				\draw[BurntOrange,thick] (0.38, 0.76) rectangle (0.58, 1);
		    \end{scope}
			\end{tikzpicture}
		\end{minipage}
		\vspace{0.1mm}
		&
		\begin{minipage}{3cm}
		\includegraphics[width=3cm,trim={13cm 0cm 14cm 13.91cm},clip,cfbox=BurntOrange 0.5mm -0.5mm]{figures/comparison/colortransfer/05137.jpg}
		\includegraphics[width=3cm,trim={13cm 0cm 14cm 13.91cm},clip,cfbox=Cerulean 0.5mm -0.5mm]{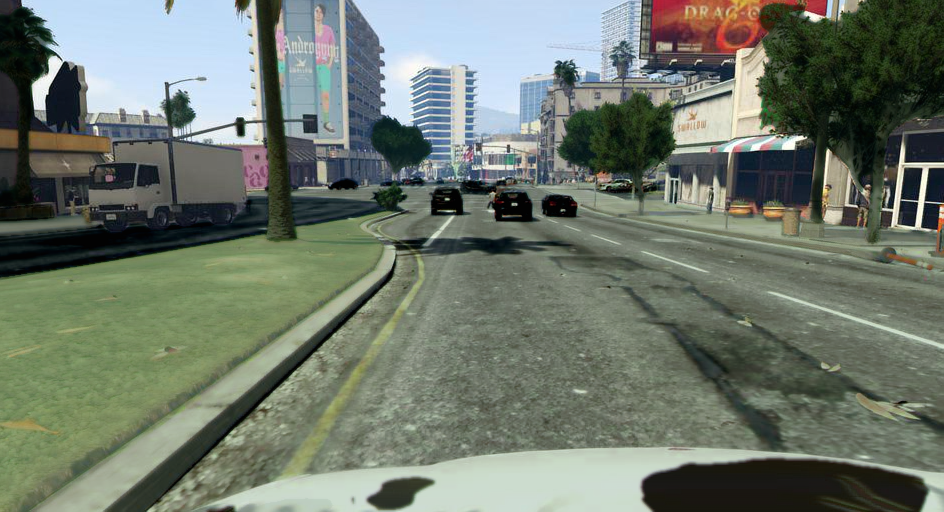}
		\end{minipage}
		\vspace{0.1mm}
		&
		\begin{minipage}{7.5cm}
			\begin{tikzpicture}
				\node[anchor=north west,inner sep=0] (image) at (0,0) {\includegraphics[width=7.5cm]{figures/comparison/wct2/05137.png}};
				\begin{scope}[x={(image.north east)},y={(image.south west)}]
					\draw[Cerulean,thick] (0.38, 0.76) rectangle (0.58, 1);
			    \end{scope}
				\end{tikzpicture}
		\end{minipage}
		\vspace{0.1mm}\\
		\footnotesize SPADE & & \footnotesize MUNIT \\
		\begin{minipage}{7.5cm}
			\begin{tikzpicture}
			\node[anchor=north west,inner sep=0] (image) at (0,0) {\includegraphics[width=7.5cm]{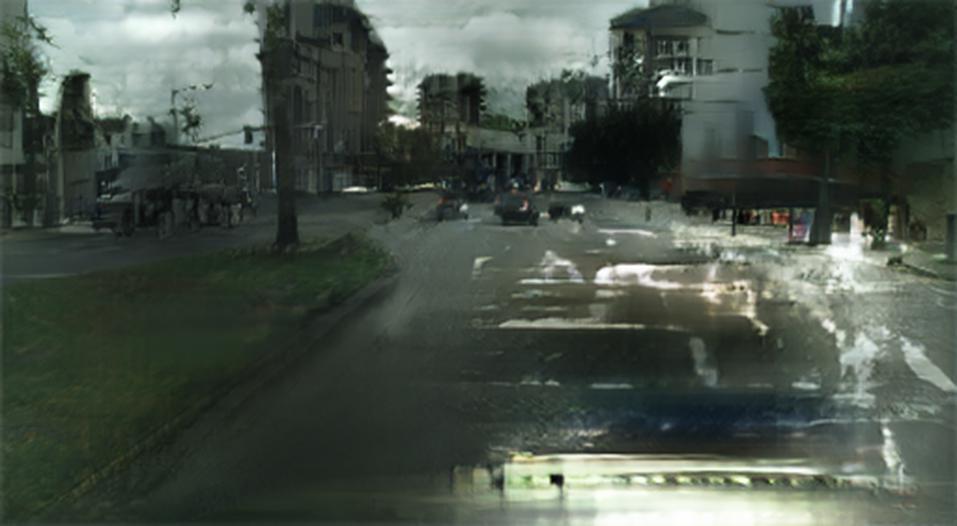}};
			\begin{scope}[x={(image.north east)},y={(image.south west)}]
				\draw[BurntOrange,thick] (0.38, 0.76) rectangle (0.58, 1);
		    \end{scope}
			\end{tikzpicture}
		\end{minipage}
		\vspace{0.1mm}
		&
		\begin{minipage}{3cm}
		\includegraphics[width=3cm,trim={13cm 0cm 14cm 13.91cm},clip,cfbox=BurntOrange 0.5mm -0.5mm]{figures/comparison/spade/05137.jpg}
		\includegraphics[width=3cm,trim={13cm 0cm 14cm 13.91cm},clip,cfbox=Cerulean 0.5mm -0.5mm]{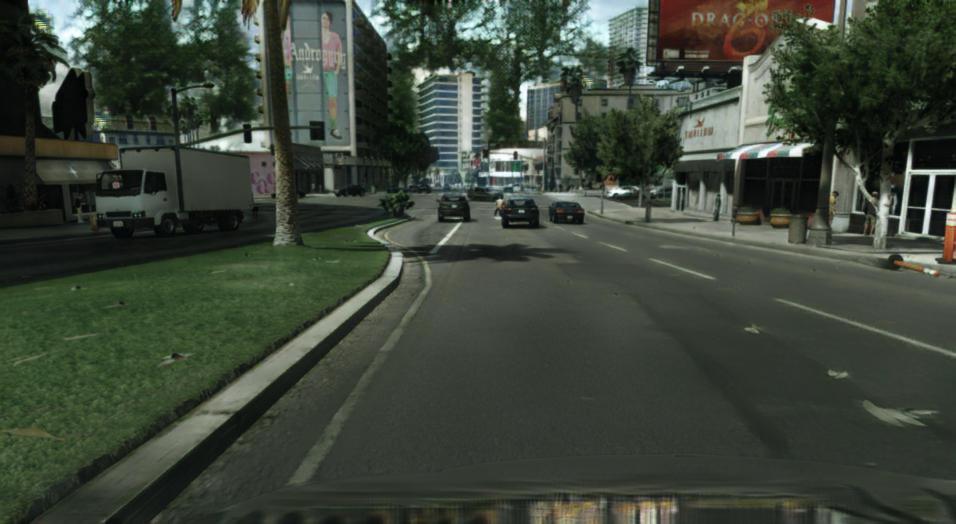}
		\end{minipage}
		\vspace{0.1mm}
		&
		\begin{minipage}{7.5cm}
			\begin{tikzpicture}
				\node[anchor=north west,inner sep=0] (image) at (0,0) {\includegraphics[width=7.5cm]{figures/comparison/munit/05137.jpg}};
				\begin{scope}[x={(image.north east)},y={(image.south west)}]
					\draw[Cerulean,thick] (0.38, 0.76) rectangle (0.58, 1);
			    \end{scope}
				\end{tikzpicture}
		\end{minipage}
		\vspace{0.1mm}\\
		\footnotesize CUT & & \footnotesize Cycada \\
		\begin{minipage}{7.5cm}
			\begin{tikzpicture}
			\node[anchor=north west,inner sep=0] (image) at (0,0) {\includegraphics[width=7.5cm]{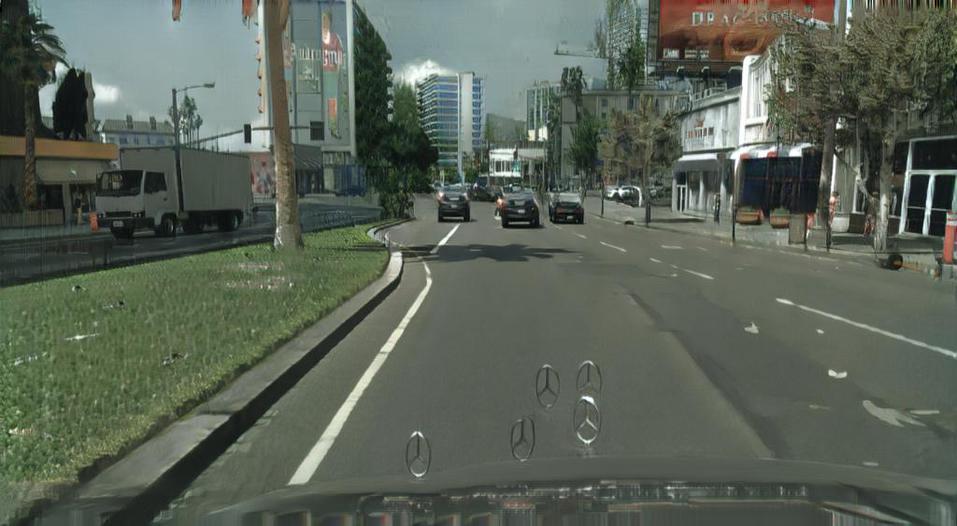}};
			\begin{scope}[x={(image.north east)},y={(image.south west)}]
				\draw[BurntOrange,thick] (0.38, 0.76) rectangle (0.58, 1);
		    \end{scope}
			\end{tikzpicture}
		\end{minipage}
		\vspace{0.1mm}
		&
		\begin{minipage}{3cm}
		\includegraphics[width=3cm,trim={13cm 0cm 14cm 13.91cm},clip,cfbox=BurntOrange 0.5mm -0.5mm]{figures/comparison/CUT/05137.jpg}
		\includegraphics[width=3cm,trim={13cm 0cm 14cm 13.91cm},clip,cfbox=Cerulean 0.5mm -0.5mm]{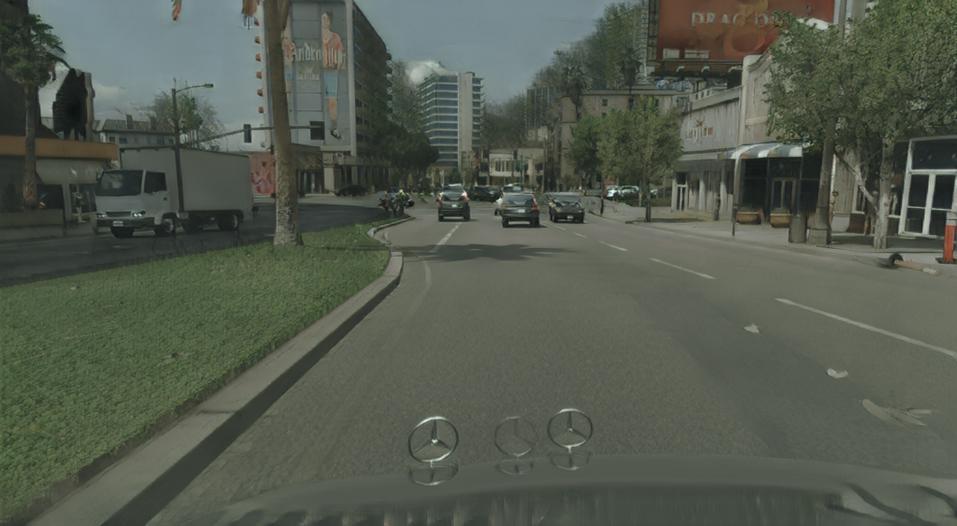}
		\end{minipage}
		\vspace{0.1mm}
		&
		\begin{minipage}{7.5cm}
			\begin{tikzpicture}
				\node[anchor=north west,inner sep=0] (image) at (0,0) {\includegraphics[width=7.5cm]{figures/comparison/cycada/05137.jpg}};
				\begin{scope}[x={(image.north east)},y={(image.south west)}]
					\draw[Cerulean,thick] (0.38, 0.76) rectangle (0.58, 1);
			    \end{scope}
				\end{tikzpicture}
		\end{minipage}
		\vspace{0.1mm}\\
		\end{tabular}
	\caption{We compare our results to original GTA images and a number of baselines. Our results are structurally consistent with the input. ColorTransfer does not modify textures. WCT2 strongly depends on a favorable reference image. SPADE fails due to differences in the scene layouts. MUNIT hallucinates trees, and Cycada and CUT hallucinate star logos.	Additional results for all baselines are shown in the supplement.}
	\label{fig:prior_work}
\end{figure*}

\subsection{Comparison to prior work}\label{sec:eval_baselines}
For the comparison to prior work, we select a number of baselines that represent multiple lines of work that can be applied to photorealism enhancement. For methods that require semantic segmentation labels as input, we provide maps for synthetic and real images predicted by MSeg~\cite{Lambert2020}, the same robust segmentation network we employ in the discriminator of our method (see the supplement for examples). We report overall results in \Tab~\ref{tab:prior} and show examples in \Fig~\ref{fig:prior_work} and in the supplement.

\mypara{Color transfer.}
We compare against classic work on color transfer. Namely, we evaluate the seminal work of Reinhard~\etal~\cite{Reinhard2001} (Color Transfer) and the color distribution transfer (CDT) of Piti{\'e} \etal~\cite{Pitie2007}. Modifications by these methods are restricted to colors of individual pixels. While this prevents enhancements of textures, it also prevents the introduction of artifacts common to more aggressive learning-based approaches, thus keeping the resulting images fairly close to the original input (\Fig~\ref{fig:prior_work}). Consequently, the largest improvements can be observed in the metrics for low-level features (sKVD$_{1-2}$ and sKVD$_{2-2}$), where the gains match the level of more recent deep learning approaches.

\begin{table*}[thb]
	\newcolumntype{Y}{S[table-format=2.2,table-auto-round]}
	\newcolumntype{Z}{>{\footnotesize\color{gray}}S[table-format=2.1,table-auto-round]}
	\newcolumntype{A}{>{\footnotesize\color{gray}}c}
	\centering
	\small
	\begin{tabularx}{\linewidth}{@{\hspace{0mm}}c@{\hspace{3mm}}X@{\hspace{3mm}}Y@{\hspace{0.5mm}}A@{\hspace{0.1mm}}Z|*{5}{@{\hspace{3mm}}Y@{\hspace{0.5mm}}A@{\hspace{0.1mm}}Z}}
		\toprule
											& Method 								& \multicolumn{3}{c}{KID} & \multicolumn{3}{c}{sKVD$_{1-2}$} & \multicolumn{3}{c}{sKVD$_{2-2}$} & \multicolumn{3}{c}{sKVD$_{3-3}$} & \multicolumn{3}{c}{sKVD$_{4-3}$}& \multicolumn{3}{c}{sKVD$_{5-3}$}\\
		\midrule
											& GTA     								& 41.44121 		& $\pm$	& 1.534779 & 330.2787 & $\pm$ & 6.535653 & 699.0819 & $\pm$ & 32.49241 & 917.3354 & $\pm$ & 53.4539 & 56.89031 & $\pm$ & 2.477653 & 11.01485 & $\pm$ & 0.3831605\\
		\midrule
		\multirow{2}{*}{Color transfer} 	& Color transfer \cite{Reinhard2001} 	& 41.31224 		& $\pm$	& 1.566259 & 36.644 & $\pm$ & 3.006222 & 301.0199 & $\pm$ & 16.74726 & 210.283 & $\pm$ & 13.17212 & 21.73576 & $\pm$ & 0.8445657 & 4.860138 & $\pm$ & 0.2036666\\
		                                  	& CDT \cite{Pitie2007}              	& 40.92414 		& $\pm$	& 1.646442 & 56.32768 & $\pm$ & 3.932661 & 459.2891 & $\pm$ & 24.73689 & 329.9116 & $\pm$ & 20.01842 & 27.15921 & $\pm$ & 1.075054 & 5.063038 & $\pm$ & 0.1894589 \\[5pt]
		\multirow{2}{*}{Photo style transfer}  	& PhotoWCT \cite{Li2018:ECCVa} 		& 56.66614 & $\pm$	& 1.858625 & \bfseries 4.508253 & $\pm$ & \bfseries 0.4784876 & 46.64516 & $\pm$ & 3.331458 & 94.36127 & $\pm$ & 9.225842 & 13.83324 & $\pm$ & 1.077324 & 3.657362 & $\pm$ & 0.2506997 \\
											& WCT2 \cite{Yoo2019}  					& 29.67029 		& $\pm$	& 1.416437 & 33.32558 & $\pm$ & 3.646628 & 160.0709 & $\pm$ & 15.18206 & 172.7016 & $\pm$ & 19.24181 & 16.15364 & $\pm$ & 0.9600473 & 3.304822 & $\pm$ & 0.1902612 \\[5pt]

		Conditional image synthesis					& SPADE \cite{Park2019:CVPR} 			& 77.39194 		& $\pm$	& 1.866913 & 11.56492 & $\pm$ & 0.8328438 & 153.0965 & $\pm$ & 4.309927 & 244.2762 & $\pm$ & 12.42884 & 44.96653 & $\pm$ & 1.856472 & 11.78003 & $\pm$ & 0.4154863 \\[5pt]
		\multirow{4}{*}{\shortstack{Image-to-image\\ translation}}
											& TSIT \cite{Jiang2020}					& 27.96616		& $\pm$ & 1.458685 	& 38.79533 & $\pm$ & 5.184864 & 291.1027 & $\pm$ & 37.08504 & 330.1205 & $\pm$ & 40.57694 & 24.87774 & $\pm$ & 1.729798 & 4.659616 & $\pm$ & 0.3404401 \\
											& CUT \cite{Park2020} 					& 21.17947 		& $\pm$	& 1.392516 & 19.44036 & $\pm$ & 2.367301 & 149.2503 & $\pm$ & 14.57768 & 214.5493 & $\pm$ & 24.15001 & 21.31399 & $\pm$ & 1.816288 & 5.082276 & $\pm$ & 0.2851401 \\
											& MUNIT \cite{Huang2018} 				& 15.72107 		& $\pm$	& 0.9023296 & 42.4135 & $\pm$ & 3.278709 & 172.7454 & $\pm$ & 15.43535 & 207.2705 & $\pm$ & 21.40878 & 18.39966 & $\pm$ & 1.132026 & 2.754651 & $\pm$ & 0.1698883 \\
											& Cycada \cite{Hoffman2018} 			& 12.29517 		& $\pm$	& 0.8542726 & 28.39047 & $\pm$ & 1.845692 & 50.70144 & $\pm$ & 4.408521 & 57.73772 & $\pm$ & 5.162635 & 6.381161 & $\pm$ & 0.4239078 & 2.221395 & $\pm$ & 0.1667504 \\
		\midrule
											& Ours 	& \bfseries 10.95085 & $\pm$	& \bfseries 0.754779 &  6.128937 & $\pm$ &  1.057475 & \bfseries 11.11894 & $\pm$ & \bfseries 2.373916 & \bfseries 15.45489 & $\pm$ & \bfseries 4.007644 & \bfseries 3.610976 & $\pm$ & \bfseries 0.488827 & \bfseries 1.271319 & $\pm$ & \bfseries 0.1239705 \\
		\bottomrule
	\end{tabularx}
	\caption{Comparison to prior work. All methods were trained on the Cityscapes dataset. Performance reported as Kernel Inception Distance $\times 1000$ (KID) and semantically aligned Kernel VGG Distance $\times 1000$ (sKVD). Lower is better. {\color{Gray}Gray} values indicate standard deviation.}
	\label{tab:prior}
\end{table*}

\mypara{Photo style transfer.}
We compare against a closed-form solution for fast photographic style transfer (PhotoWCT)~\cite{Li2018:ECCVa} and the state of the art, based on wavelet transforms (WCT2)~\cite{Yoo2019}.
Both photo style transfer approaches require a style image and semantic segmentation maps for both source and style image. While color transfer methods apply transformations to individual pixel colors, photo style transfer methods perform transformations in learned higher-level feature spaces guided by semantic segmentations, and thus modify images more strongly. However, photo style transfer methods rely on a favorable style image that matches that input synthetic image. When the input image changes, as happens during interactive exploration of a synthetic environment, photo style transfer can produce unrealistic color shifts or temporal instability.

\mypara{Conditional image synthesis.}
We compare against a representative approach to conditional image synthesis, SPADE~\cite{Park2019:CVPR}, as it dominates preceding approaches (\eg \cite{ChenKoltun2017, Isola2017, Wang2018:CVPR}). We use a model pretrained for synthesizing urban street scenes, provided by the authors. (The model is pretrained on segmentation ground-truth from Cityscapes, which is compatible with GTA.)
SPADE considerably underperforms other methods. This can be explained by two factors. First, synthesizing a photo from only a semantic segmentation map is more challenging than modifying a given image. Second, since SPADE is trained to synthesize images from the Cityscapes dataset (and does so quite well), the distribution shift in scene layouts between Cityscapes and GTA takes this model far outside its training distribution.

\mypara{Image-to-image translation.}
We compare against Cycada~\cite{Hoffman2018}, which was specifically designed for adapting synthetic images to real photos. To this end, it augments the commonly used pixel-level cycle-consistency with a feature-level cycle-consistency and a semantic consistency loss. The latter leverages corresponding semantic labels to preserve the content of the synthetic images.
As Cycada was originally trained on GTA V and Cityscapes already, we use images provided by the authors.
We further compare against MUNIT~\cite{Huang2018}, CUT~\cite{Park2020}, and TSIT~\cite{Jiang2020}.
MUNIT extends the CycleGAN~\cite{Zhu2017} architecture to multi-modal translation and adds a domain-invariant perceptual loss. CUT and TSIT abandon the cycle consistency constraint. CUT builds instead on contrastive learning, and TSIT uses an exemplar style image for transferring style features.

We find that among the various types of baselines, image-to-image translation methods perform best, and within this group Cycada demonstrates the best results.
Although Cycada makes use of more explicit semantic information than other methods that use a perceptual loss, it still hallucinates extraneous objects (see \Fig~\ref{fig:gan_artifacts}).
A possible reason may be its segmentation network, which is pretrained on unmodified synthetic images and is not updated during training of the image synthesis network.
Thus Cycada implicitly assumes that the drift between synthetic and real images will be limited and the segmentation network will continue to work reliably.
Another possible reason is training on large image patches (see \Sec~\ref{sec:structural_shift} for the associated discussion).

\subsection{Perceptual experiment}
\begin{figure}[thb]
	\centering
	\includegraphics[width=85mm]{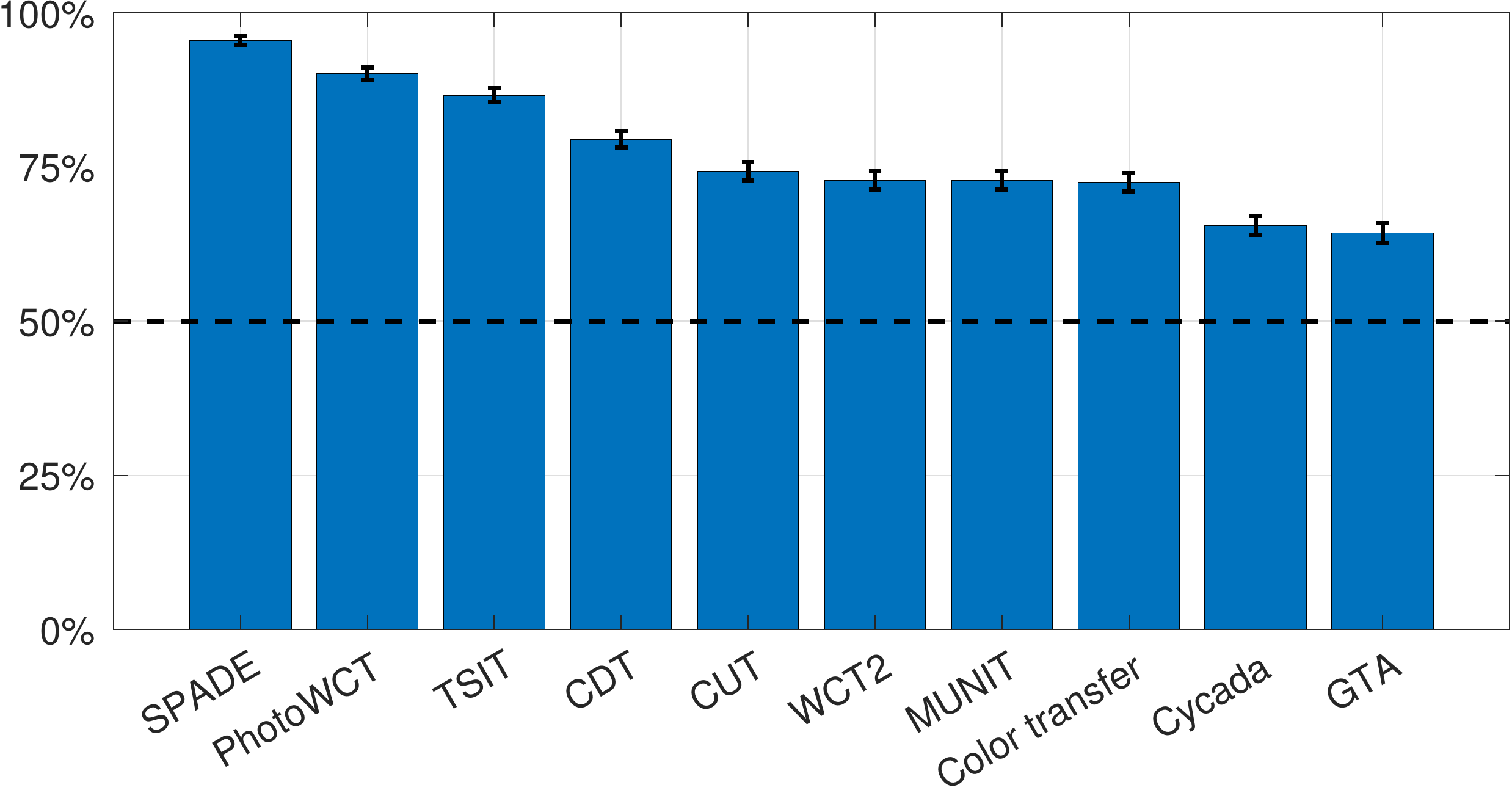}
	\caption{Perceptual experiment. Randomized blind unlimited-time pairwise comparisons of corresponding images. Values indicate percentage of comparisons in which our method was considered more realistic than the alternative. Chance is at 50\%, indicated by the dashed line. Error bars indicate 95\% confidence intervals.}
	\label{fig:mturk}
\end{figure}

We additionally compare our approach to all baselines in a large-scale crowdsourced perceptual experiment.
The setup of the perceptual experiment follows prior work and is based on pairwise comparisons with no time limit~\cite{ChenKoltun2017}.
We sample 500 images from GTA and compare enhancements from our method to baselines.
For each image, our method is paired with each baseline. The pairs are presented to crowd workers.
Each pairing is shown 10 times in total (across workers), resulting in 50,000 comparisons in all.
The assignment of pairs to workers, their order of presentation, and the left-right order within pairs are all randomized.
To filter unreliable workers we include sentinel tasks.
The results are shown in \Fig~\ref{fig:mturk}.
We find that images enhanced by our method are consistently considered more realistic than all baseline methods.
The results are statistically significant for all methods (error bars indicate 95\% confidence intervals).

\subsection{Controlled experiments}
\label{sec:ablation}
To assess the effect of specific ideas in our approach, we conduct a set of controlled experiments.
We evaluate our sampling strategy, the importance of G-buffers, architectures for ingesting G-buffers, and different setups for the adversarial loss.
The results are shown in \Tab~\ref{tab:results_controlled}.
The first column identifies the experimental question and the second row lists the baseline condition. The first row provides reference metrics for unmodified GTA images. The last row provides the metrics for our full approach.
\begin{table*}[thb]
	\newcolumntype{Y}{S[table-format=2.2,table-auto-round]}
	\newcolumntype{Z}{>{\footnotesize\color{gray}}S[table-format=2.1,table-auto-round]}
	\newcolumntype{A}{>{\footnotesize\color{gray}}c}
	\centering
	\small
	\begin{tabularx}{\linewidth}{@{\hspace{0mm}}c@{\hspace{4mm}}X*{5}{@{\hspace{3mm}}Y@{\hspace{0.5mm}}A@{\hspace{0.1mm}}Z}@{\hspace{0mm}}}
		\toprule
		Experiment & Method & \multicolumn{3}{c}{sKVD$_{1-2}$} & \multicolumn{3}{c}{sKVD$_{2-2}$} & \multicolumn{3}{c}{sKVD$_{3-3}$} & \multicolumn{3}{c}{sKVD$_{4-3}$}& \multicolumn{3}{c}{sKVD$_{5-3}$}\\
		\midrule
		& GTA     					& 330.2787 & $\pm$ & 6.535653 & 699.0819 & $\pm$ & 32.49241 & 917.3354 & $\pm$ & 53.4539 & 56.89031 & $\pm$ & 2.477653 & 11.01485 & $\pm$ & 0.3831605 \\
		\midrule
		\multirow{3}{*}{How to sample?} & Ours (Unif. sampl., crop 400) & 43.80217 & $\pm$ & 3.996178 & 72.94973 & $\pm$ & 10.63883 & 61.67003 & $\pm$ & 11.9271 & 9.501995 & $\pm$ & 1.030827 & 1.940683 & $\pm$ & 0.2258251 \\
		& Ours (Unif. sampl., crop 256) & 10.49606 & $\pm$ & 1.605082 & 52.59904 & $\pm$ & 7.940396 & 70.98389 & $\pm$ & 13.32213 & 9.455297 & $\pm$ & 1.03067 & 2.477174 & $\pm$ & 0.264027 \\
		& Ours (Unif. sampl., crop 196) &  2.629832 &  $\pm$ &  0.561474 & 12.59793 & $\pm$ & 1.822784 & 18.07637 & $\pm$ & 4.53156 & 4.299577 & $\pm$ & 0.4892785 & 1.327728 & $\pm$ & 0.1359656 \\
		\midrule
		\multirow{2}{*}{Do G-buffers help?} & Ours (No G-buffer)  			&  5.843171 & $\pm$ &  0.7752008 & 23.07527 & $\pm$ & 3.121762 & 34.7575 & $\pm$ & 4.50609 & 6.336807 & $\pm$ & 0.5089464 & 1.957303 & $\pm$ & 0.1675012  \\
		& Ours (VIPER)  			& 1.590144 & $\pm$ & 0.2800436 & 15.70158 & $\pm$ & 1.18074 & 20.73757 & $\pm$ & 2.735168 & 5.817426 & $\pm$ & 0.5067069 & 1.603497 & $\pm$ & 0.1527528  \\
		\midrule
		\multirow{2}{*}{How to ingest G-buffers?} & Ours (Concat)	  			& 9.804612 & $\pm$ & 1.519726 & 47.06616 & $\pm$ & 6.993734 & 90.2788 & $\pm$ & 13.00831 & 9.259353 & $\pm$ & 0.8533835 & 2.339759 & $\pm$ & 0.2066608 \\
		& Ours (SPADE)	  			& 32.52044 & $\pm$ & 3.172519 & 192.2097 & $\pm$ & 22.30852 & 296.1047 & $\pm$ & 35.11985 & 23.54026 & $\pm$ & 1.818164 & 3.978143 & $\pm$ & 0.2740352 \\
		\midrule
		\multirow{3}{*}{Which discriminator?} & Ours (PatchGAN)  			& 45.45944 & $\pm$ & 4.033944  & 42.28199  & $\pm$ & 6.383632  & 43.97908 & $\pm$ & 7.806136 & 7.903254 & $\pm$ & 0.5764676 & 2.397025 & $\pm$ & 0.1683094 \\
		& Ours (No projection)		&  3.26578 & $\pm$ & 0.4139223 &  9.886326 & $\pm$ & 0.8688925 & 19.0776  & $\pm$ & 3.447557 & 4.941162 & $\pm$ & 0.4655654 & 1.580883 & $\pm$ & 0.1383063 \\
		& Ours (No adaptive backprop) & 8.089039 & $\pm$ & 1.021271 & 20.83244 & $\pm$ & 1.943034 & 20.07248 & $\pm$ & 2.12357 & 4.076078 & $\pm$ & 0.2982875 &  1.070638 & $\pm$ &  0.1046267  \\
		\midrule
		& Ours    					& 6.128937 & $\pm$ & 1.057475 &  11.11894 & $\pm$ &  2.373916 &  15.45489 & $\pm$ &  4.007644 &  3.610976 & $\pm$ &  0.488827 & 1.271319 & $\pm$ & 0.1239705 \\
		\bottomrule
	\end{tabularx}
	\caption{Controlled experiments. Each specific idea in our approach outperforms the respective baselines. In each condition, we train for 600K iterations on GTA and Cityscapes. Lower is better. {\color{Gray}Gray} values indicate standard deviation.}
	\label{tab:results_controlled}
\end{table*}

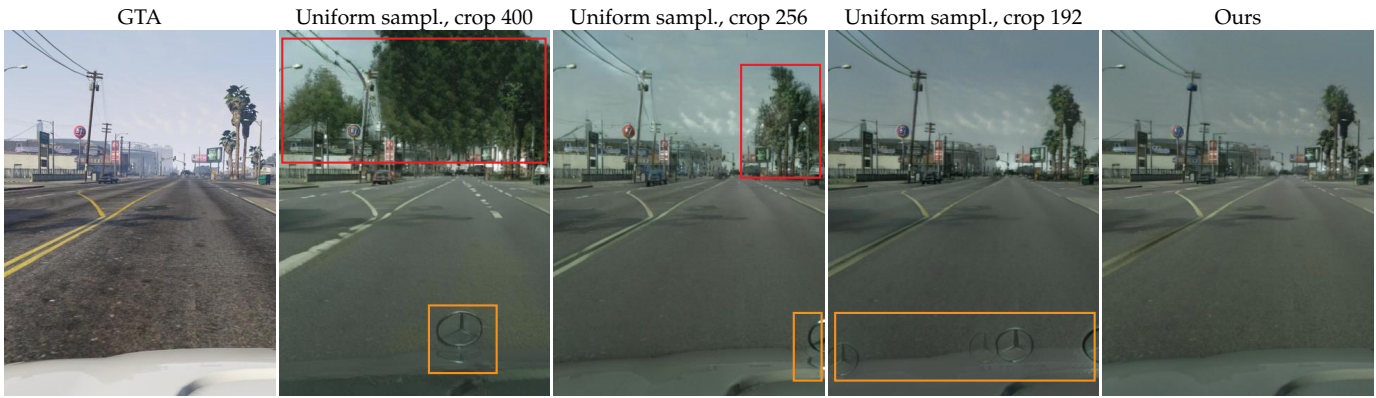
\begin{figure*}[t]
	\centering
	\renewcommand{\arraystretch}{0.5}
	\begin{tabular}{@{}c*{4}{@{\hspace{0.5mm}}c}@{}}
	\myfootnotesize GTA & \myfootnotesize Uniform sampl., crop 400 & \myfootnotesize Uniform sampl., crop 256 & \myfootnotesize Uniform sampl., crop 192 & \myfootnotesize Ours\\
	\includegraphics[width=3.6cm,trim={10cm 0cm 10cm 0cm},clip]{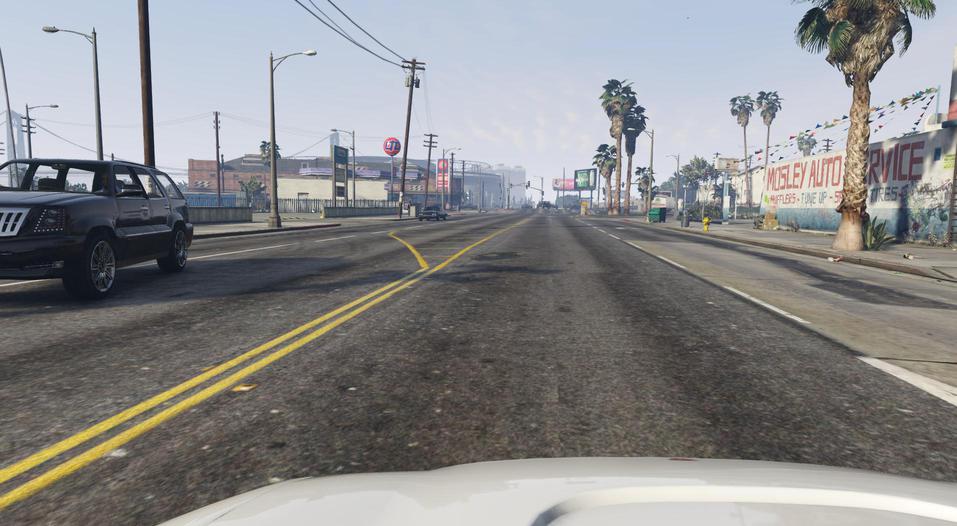} &
	\begin{tikzpicture}
			\node[anchor=south west,inner sep=0] (image) at (0,0) {\includegraphics[width=3.6cm,trim={10cm 0cm 10cm 0cm},clip]{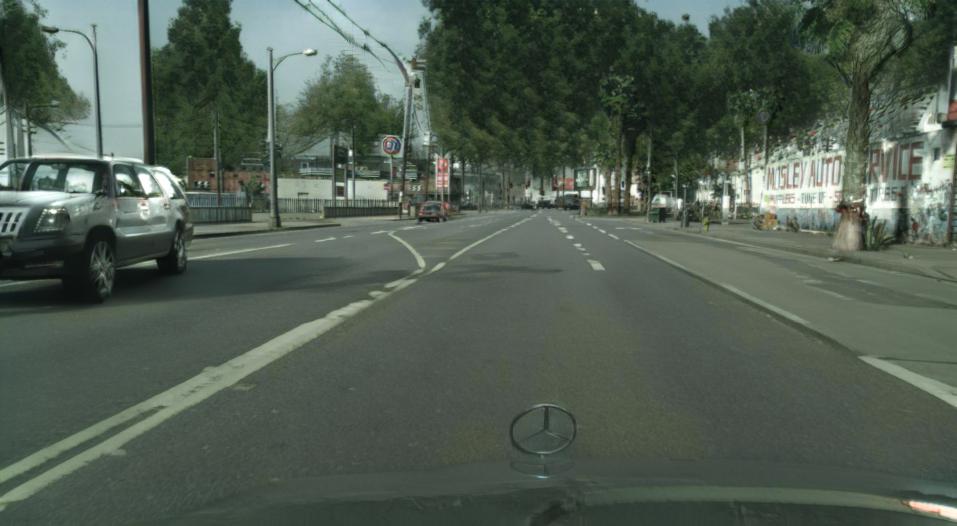}};
			\draw[Red,thick] (0.05, 3.1) rectangle (3.55, 4.75);
			\draw[BurntOrange,thick] (2, 0.3) rectangle (2.9, 1.2);
		    \end{tikzpicture}&
	\begin{tikzpicture}
			\node[anchor=south west,inner sep=0] (image) at (0,0) {\includegraphics[width=3.6cm,trim={10cm 0cm 10cm 0cm},clip]{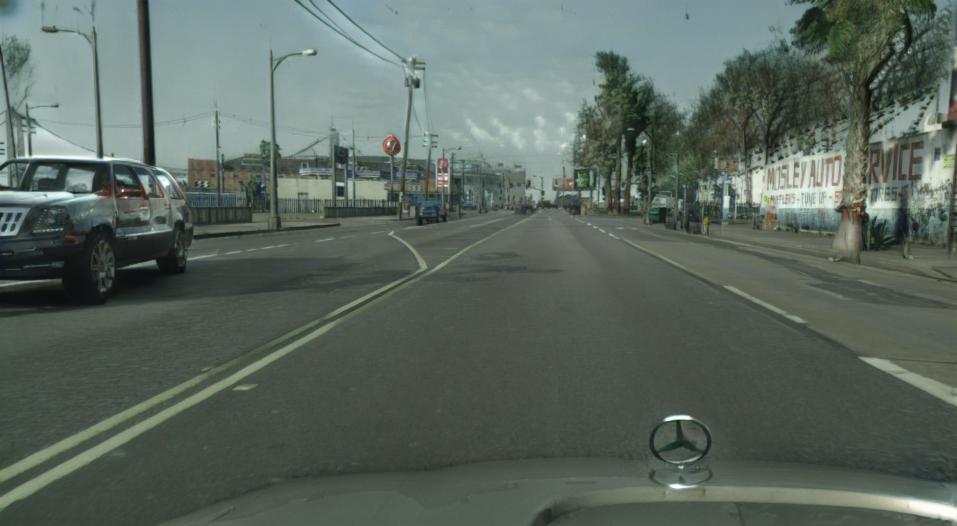}};
			\draw[Red,thick] (2.5, 2.9) rectangle (3.55, 4.4);
			\draw[BurntOrange,thick] (3.2, 0.2) rectangle (3.57, 1.1);
		    \end{tikzpicture}&
	\begin{tikzpicture}
			\node[anchor=south west,inner sep=0] (image) at (0,0) {\includegraphics[width=3.6cm,trim={10cm 0cm 10cm 0cm},clip]{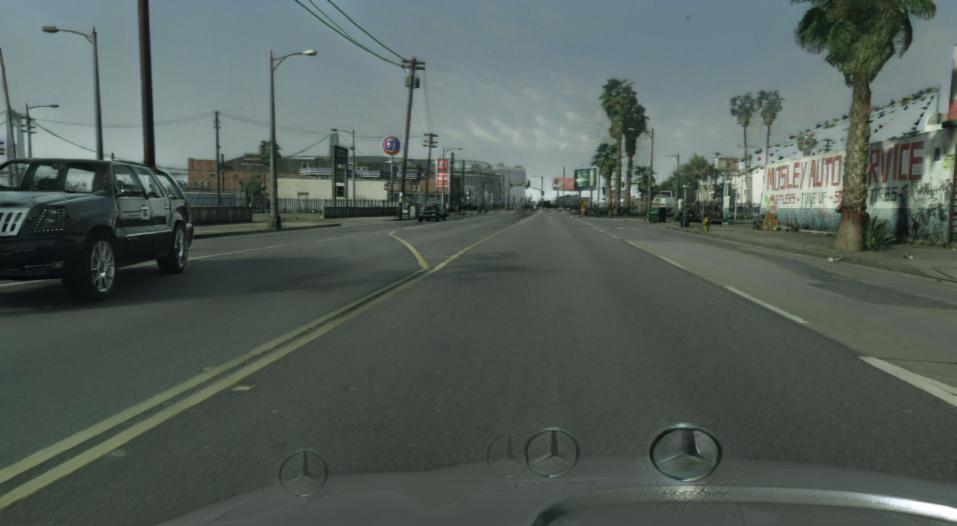}};
			\draw[BurntOrange,thick] (0.1, 0.2) rectangle (3.55, 1.1);
		    \end{tikzpicture}&
	\includegraphics[width=3.6cm,trim={10cm 0cm 10cm 0cm},clip]{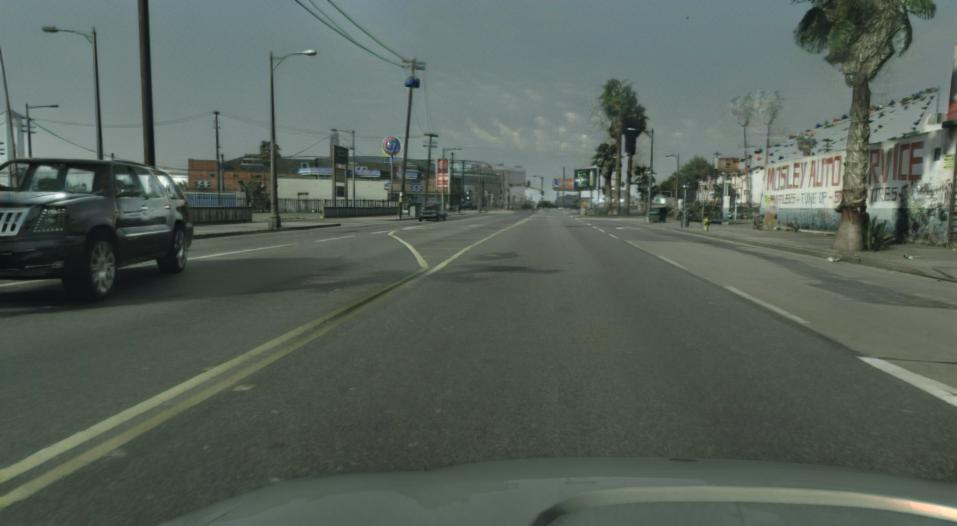}\\
	\end{tabular}
	\caption{Uniform sampling (columns 2--4) and larger patch sizes (columns 2 and 3) lead to greater mismatch between source and target datasets, inducing artifacts such as hallucinated {\color{Red}trees} and {\color{BurntOrange}stars}. These can be avoided with our sampling strategy (right).}
	\label{fig:results_cropsize}
\end{figure*}

\mypara{How to sample?}
To assess how the patch sampling strategy affects photorealism enhancement, we compare uniform sampling with different patch sizes (196, 256, 400) to the sampling of matching patch pairs (Ours).
The results are consistent with our hypothesis that sampling at smaller patch sizes reduces the mismatch between source and target datasets. We observe stronger hallucination artifacts for larger patch sizes (\Fig~\ref{fig:results_cropsize}, columns 2 \& 3).
Sampling at smaller patch sizes reduces the sKVD considerably. Sampling matching patches further reduces sKVD at medium to high levels of abstraction, while slightly increasing sKVD at the lowest level~(\Tab~\ref{tab:results_controlled}, How to sample?). This may be explained by the benefits of diversity when uniformly sampling patches, offset by the distribution mismatch for higher levels.

\mypara{Do G-buffers help?}
The set of available G-buffers depends on the rendering method used by the game and the method for capturing G-buffers.
For example, recording at video rates and lack of deep integration with the game engine limited the set of G-buffers recorded for the VIPER dataset~\cite{Richter2017}. (The set of G-buffers available for VIPER includes normal, reflection, depth, normal $\cdot$ view, and semantic segmentation.) To investigate the effect of the available set of G-buffers on photorealism enhancement, we compare three conditions: not using G-buffers at all (No G-buffer), the limited set of buffers available for VIPER~\cite{Richter2017} (VIPER), and our full set.

We find that without any G-buffers, enhanced images are significantly less realistic at all but the lowest level.
Without any auxiliary information provided by the G-buffers, the network seems to focus much more on low-level features.
Adding the buffers from VIPER improves realism at all levels, with the strongest effects observed at low and medium levels.
With our full set, the $\skvd_{1-2}$ is higher, which suggests that the network allocates more capacity to enhancing mid- and high-level features, for which sKVD is reduced.

\mypara{How to ingest G-buffers?}
We investigate several strategies for ingesting the G-buffers into the image enhancement network.
The first is to simply append them to the rendered image (Concat).
This corresponds to the strategy employed by AlHaija~\etal.\cite{AlHaija2018}.
As this variant does not treat G-buffers in any special way, RAD modules are not required, and we use instance normalization instead. This condition uses a standard HRNet architecture for image enhancement (no RAD modules or RAD blocks).
In the second condition, we replace our RAD modules by SPADE modules~\cite{Park2019:CVPR} (SPADE).
The third condition is our full approach, which uses our RAD modules (last row of \Tab~\ref{tab:results_controlled}).

\begin{figure*}[htb]
	\centering
	\renewcommand{\arraystretch}{0.5}
	\begin{tabular}{@{}c*{2}{@{\hspace{0.5mm}}c}@{}}
	\myfootnotesize GTA & \myfootnotesize Ours (SPADE) & \myfootnotesize Ours \\
	\includegraphics[width=0.33\linewidth]{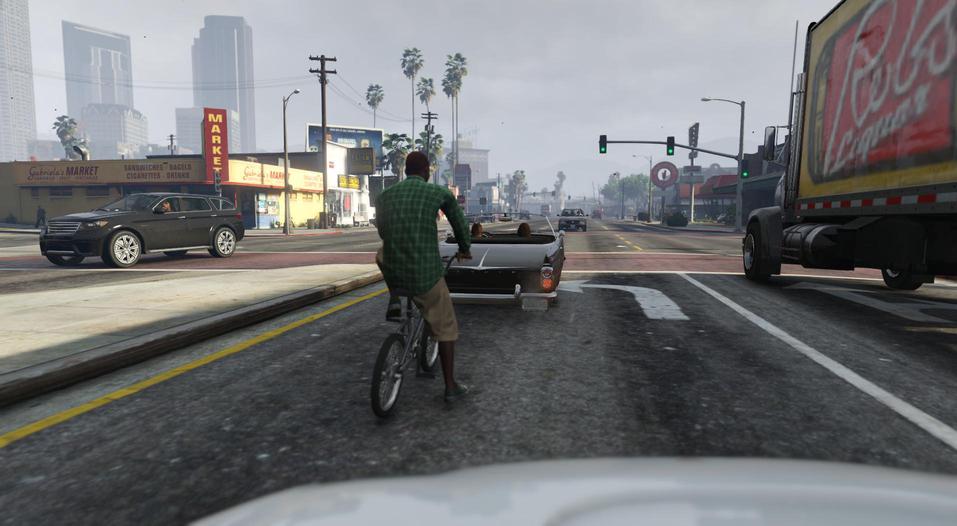} &
	\includegraphics[width=0.33\linewidth]{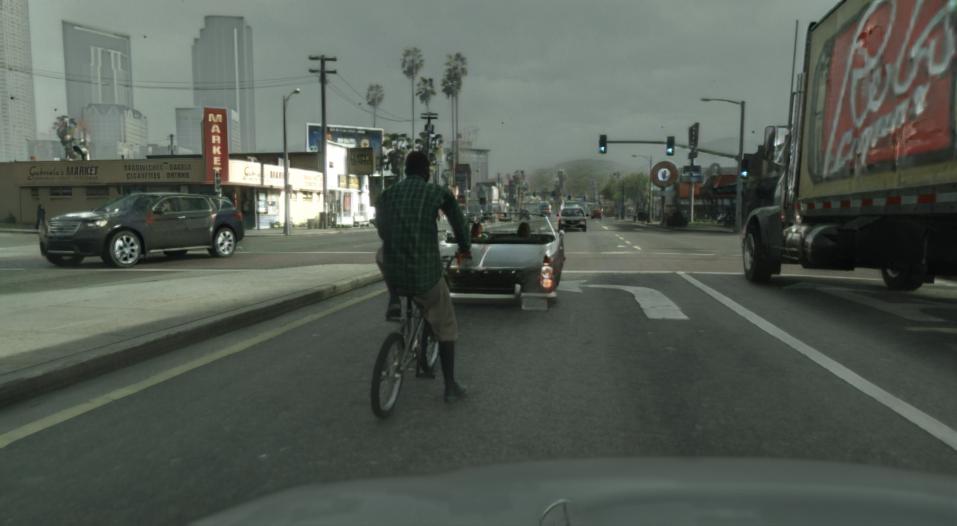} &
	\includegraphics[width=0.33\linewidth]{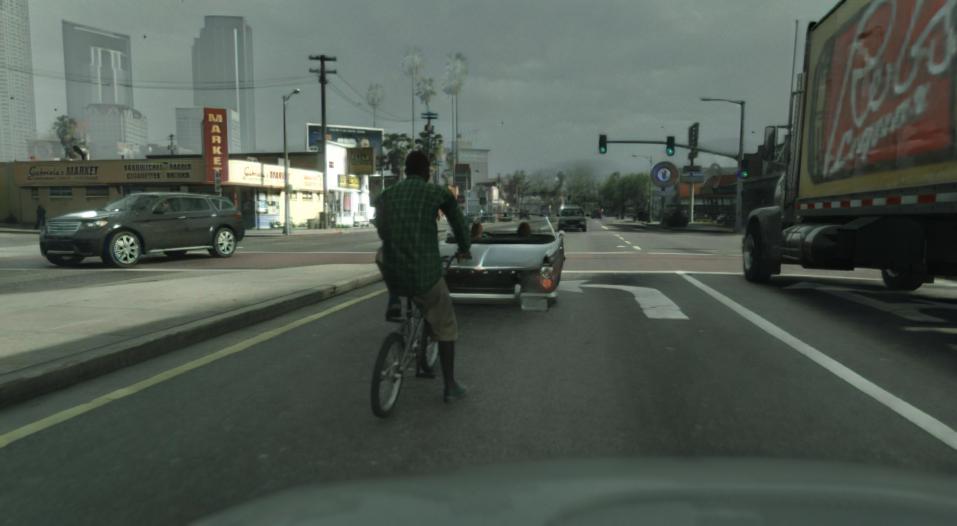} \\
	\includegraphics[width=0.33\linewidth]{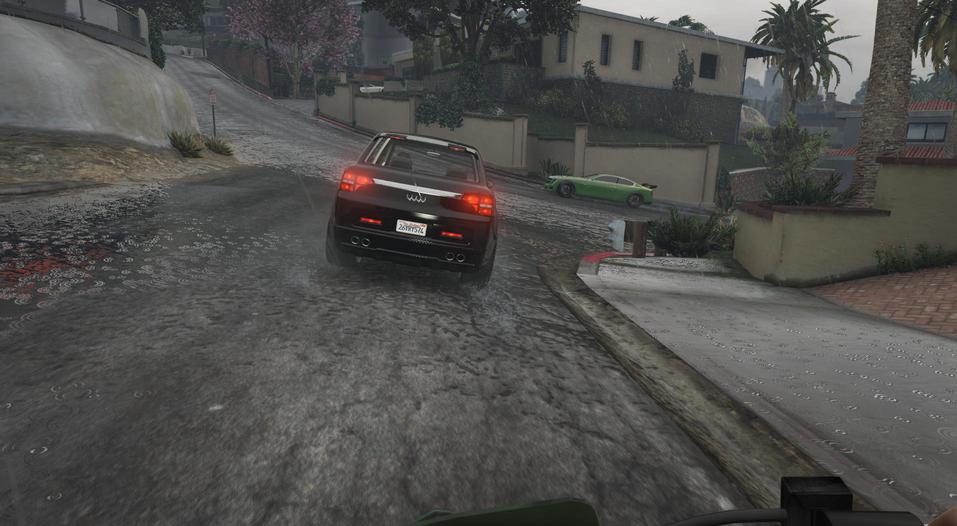} &
	\includegraphics[width=0.33\linewidth]{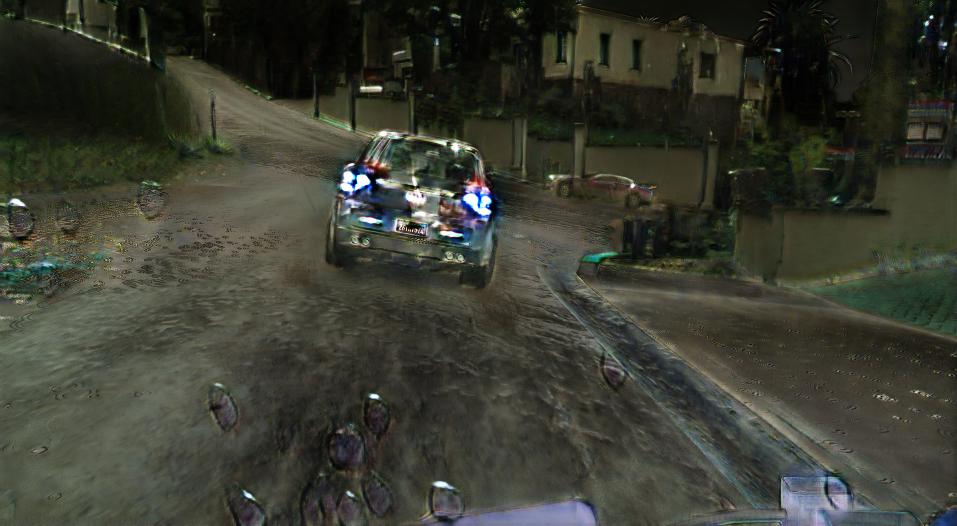} &
	\includegraphics[width=0.33\linewidth]{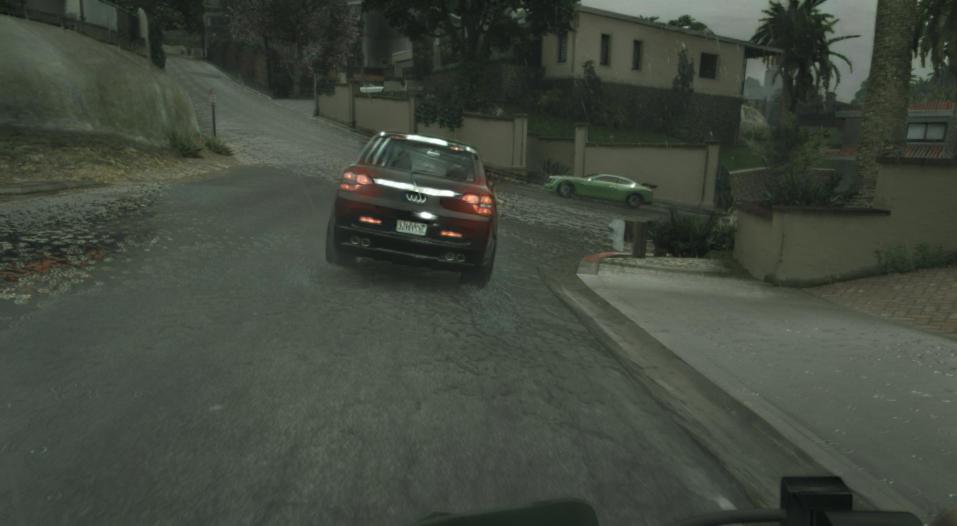} \\
	\end{tabular}
	\caption{Enhancement of GTA images (left) using SPADE modules (middle) is highly volatile across the dataset (compare the top and bottom rows). Our RAD modules yield more consistent results (right).}
	\label{fig:gbuf_architecture}
\end{figure*}

The results indicate that simple concatenation yields better results than the SPADE modules.
The results with SPADE modules are volatile across the dataset, ranging from realistic images to complete failures with strong artifacts and color shifts (\Fig~\ref{fig:gbuf_architecture}, middle column).
In contrast, our results with RAD modules are of consistently high quality (\Fig~\ref{fig:gbuf_architecture}, right column).
This is confirmed by all metrics in \Tab~\ref{tab:results_controlled}.

\mypara{Which discriminator?}
We now assess the importance of specific ideas in the perceptual discriminator.
As a baseline we pick the PatchGAN discriminator since it is standard in image-to-image translation~\cite{Huang2018,Isola2017,Jiang2020,Park2019:CVPR,Wang2018:CVPR,Wang2018:Neurips,Zhu2020}. Specifically, the PatchGAN condition uses four discriminator networks that each ingest images at a different scale. The network stems share the same architecture as our discrminator networks, except for the projection of segmentations, which is not part of the PatchGAN architecture.
In the next condition, we remove the robust segmentation network from our approach (No projection). In the third condition, we train the discriminator without adaptively throttling the different discriminator networks as described in \Sec~\ref{sec:implementation} (No adaptive backprop). The fourth condition is our full approach (last row of \Tab~\ref{tab:results_controlled}).
\begin{figure*}[htb]
	\centering
	\renewcommand{\arraystretch}{0.5}
	\begin{tabular}{@{}c*{2}{@{\hspace{0.5mm}}c}@{}}
	\myfootnotesize GTA & \myfootnotesize Ours (PatchGAN) & \myfootnotesize Ours\\
	\includegraphics[width=6cm,trim={2cm 5cm 7cm 2cm},clip]{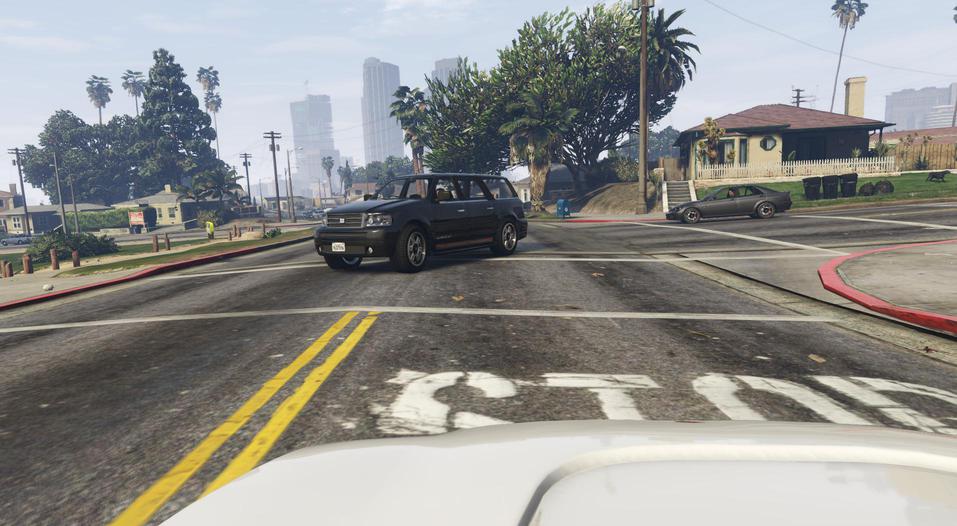} &
	\begin{tikzpicture}
			\node[anchor=south west,inner sep=0] (image) at (0,0) {\includegraphics[width=6cm,trim={2cm 5cm 7cm 2cm},clip]{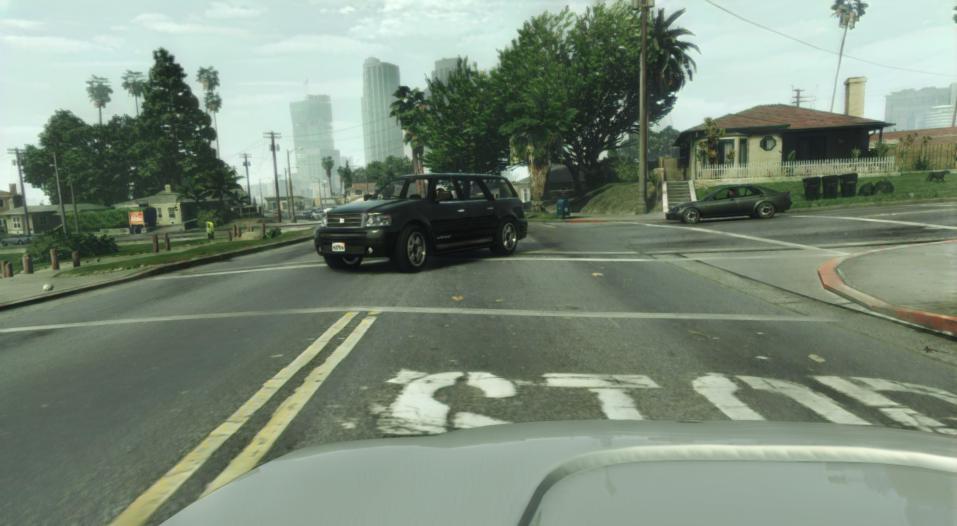}};
			\draw[BurntOrange,thick] (2.0, 0.9) rectangle (4.1, 2);
		    \end{tikzpicture}&
	\begin{tikzpicture}
			\node[anchor=south west,inner sep=0] (image) at (0,0) {\includegraphics[width=6cm,trim={2cm 5cm 7cm 2cm},clip]{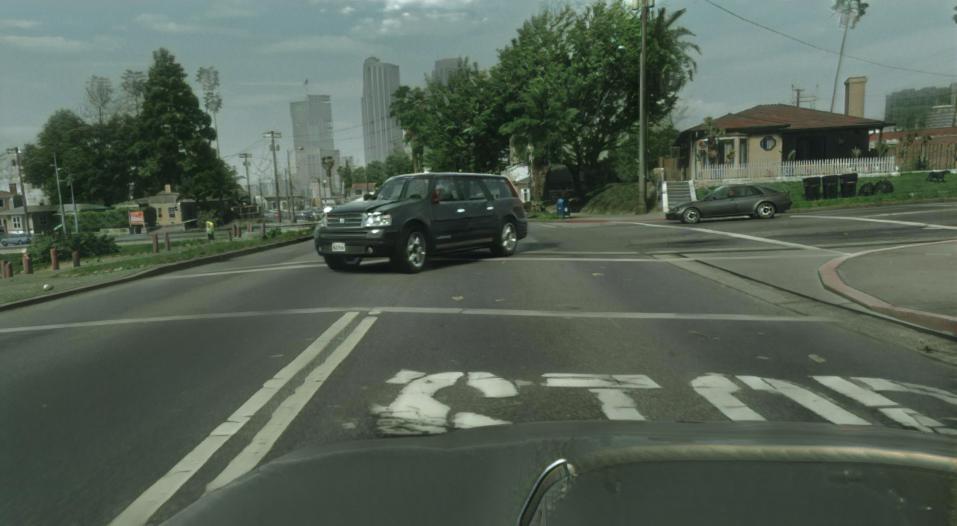}};
			\draw[BurntOrange,thick] (2.0, 0.9) rectangle (4.1, 2);
		    \end{tikzpicture}\\
	\end{tabular}
	\caption{Our perceptual discriminator (right) is more effective than a PatchGAN discriminator (middle). Note the stronger enhancements such as lights and reflections on the car on the right.}
	\label{fig:results_patchgan}
\end{figure*}

\begin{figure*}[htb]
	\centering
	\renewcommand{\arraystretch}{0.5}
	\begin{tabular}{@{}c*{2}{@{\hspace{0.5mm}}c}@{}}
	\myfootnotesize GTA & \myfootnotesize Ours (No projection) & \myfootnotesize Ours\\
	\includegraphics[width=6cm,trim={0cm 0cm 0cm 2.5cm},clip]{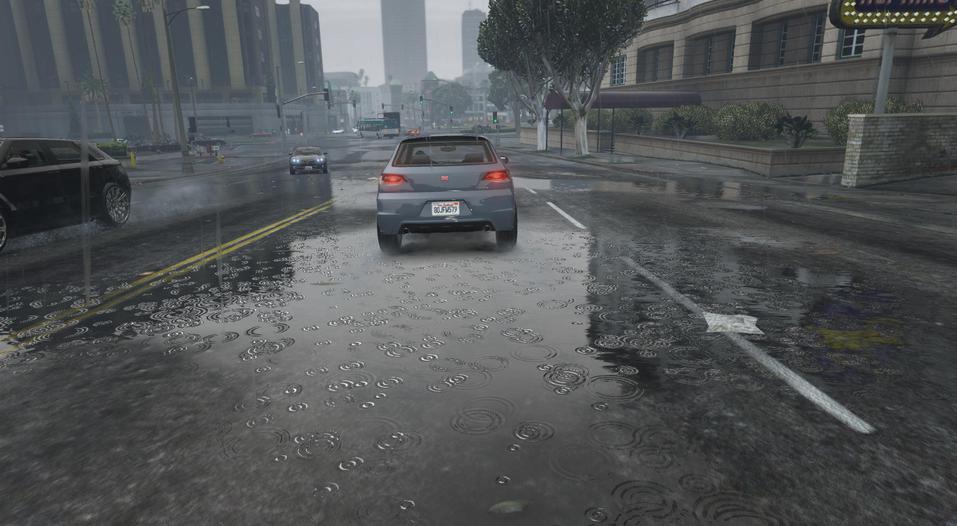} &
	\begin{tikzpicture}
			\node[anchor=south west,inner sep=0] (image) at (0,0) {\includegraphics[width=6cm,trim={0cm 0cm 0cm 2.5cm},clip]{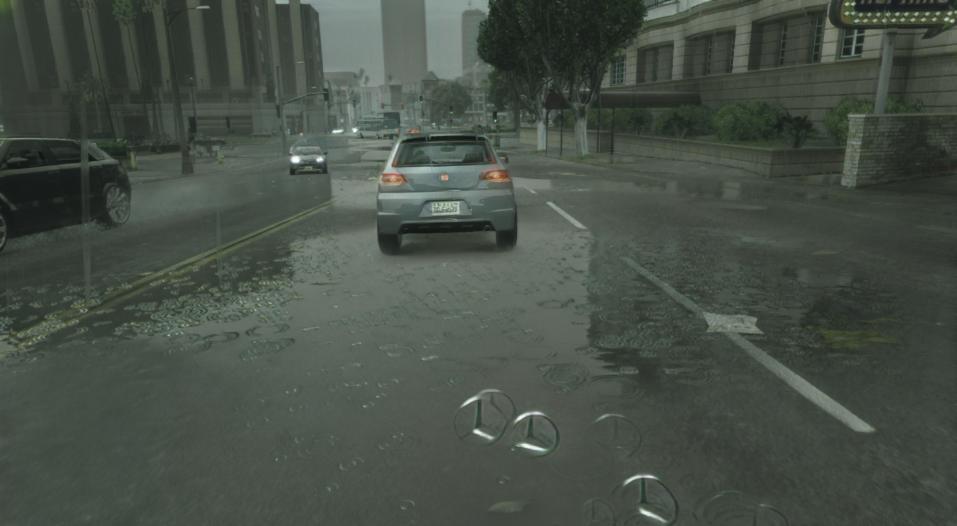}};
			\draw[BurntOrange,thick] (2.8, 0.05) rectangle (4.4, 1);
		    \end{tikzpicture}&
	\begin{tikzpicture}
			\node[anchor=south west,inner sep=0] (image) at (0,0) {\includegraphics[width=6cm,trim={0cm 0cm 0cm 2.5cm},clip]{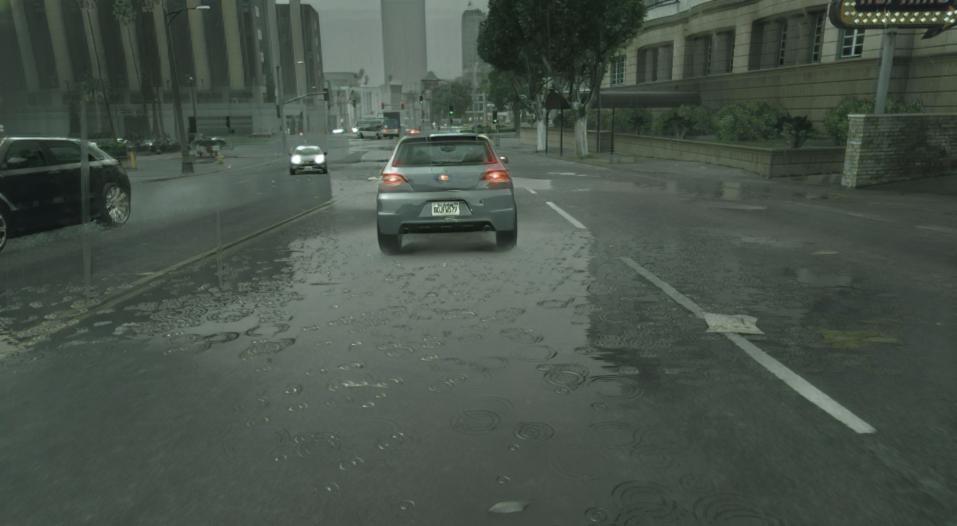}};
			\draw[BurntOrange,thick] (2.8, 0.05) rectangle (4.4, 1);
		    \end{tikzpicture}\\
	\end{tabular}
	\caption{Adding the projection of semantic embeddings to the discriminator removes artifacts.}
	\label{fig:results_projection}
\end{figure*}

As reported in \Tab~\ref{tab:results_controlled}, the results with the PatchGAN discriminator are significantly less realistic. This is illustrated in \Fig~\ref{fig:results_patchgan}.
The projection layer and adaptive backpropagation both help, but at different perceptual levels. Removing adaptive backpropagation hurts sKVD at all but the highest level.
Removing the projection layer increases sKVD at the higher levels. The effect of the projection layer is illustrated in \Fig~\ref{fig:results_projection}.
The combination of projection and adaptive backpropagation is beneficial when all levels are taken into account.

\section{Conclusion}\label{sec:discussion}
Our approach significantly enhances the realism of rendered images. This is confirmed by a comprehensive evaluation of our method against strong baselines.
Intuitively, our method achieves the strongest and most consistent results for objects and scenes that have clear correspondences in the real dataset; our method excels at road textures, cars, and vegetation. Objects and scenes that are less common in the real images (\eg close-up pedestrians) are modified less convincingly.
Overall, our approach produces high-quality enhancements that are geometrically and semantically consistent with the input images while matching the style of the respective dataset. (Additional results with Mapillary Vistas are shown in \Fig~\ref{fig:more_vistas} and in the supplement.)
\begin{figure*}[thbp]
	\centering
	\newcolumntype{P}[1]{>{\centering\arraybackslash}p{#1}}
	\renewcommand{\arraystretch}{0.5}
	\begin{tabular}{@{}P{7.5cm}@{}P{3cm}@{}P{7.5cm}@{}}
		\footnotesize Input (GTA) & & \footnotesize Ours (trained on Mapillary Vistas)\\
		\begin{minipage}{7.5cm}
			\begin{tikzpicture}
			\node[anchor=north west,inner sep=0] (image) at (0,0) {\includegraphics[width=7.5cm]{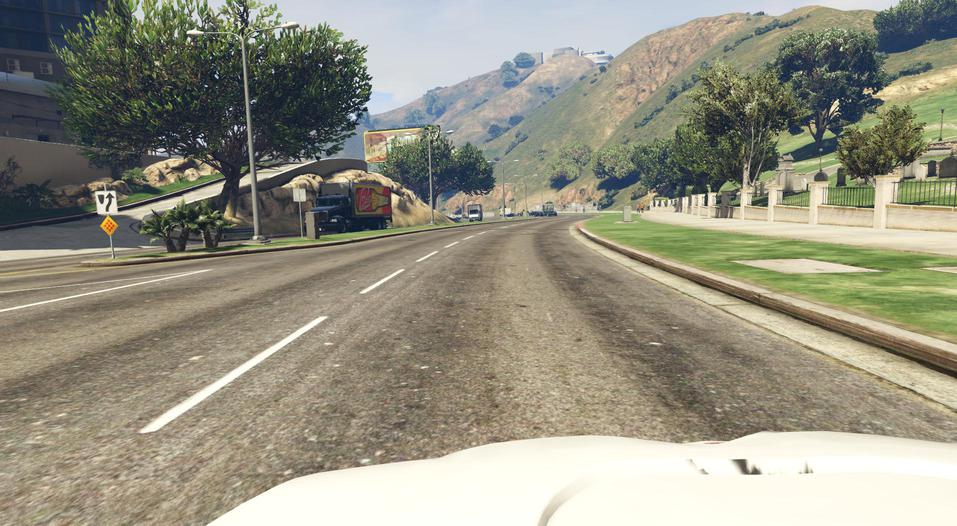}};
			\begin{scope}[x={(image.north east)},y={(image.south west)}]
				\draw[BurntOrange,thick] (0.59, 0.39) rectangle (0.795, 0.63);
		    \end{scope}
			\end{tikzpicture}
		\end{minipage}
		\vspace{0.1mm}
		&
		\begin{minipage}{3cm}
		\includegraphics[width=3cm,trim={20cm 7cm 7cm 6.91cm},clip,cfbox=BurntOrange 0.5mm -0.5mm]{figures/comparison/gta/01316.jpg}
		\includegraphics[width=3cm,trim={20cm 7cm 7cm 6.91cm},clip,cfbox=Cerulean 0.5mm -0.5mm]{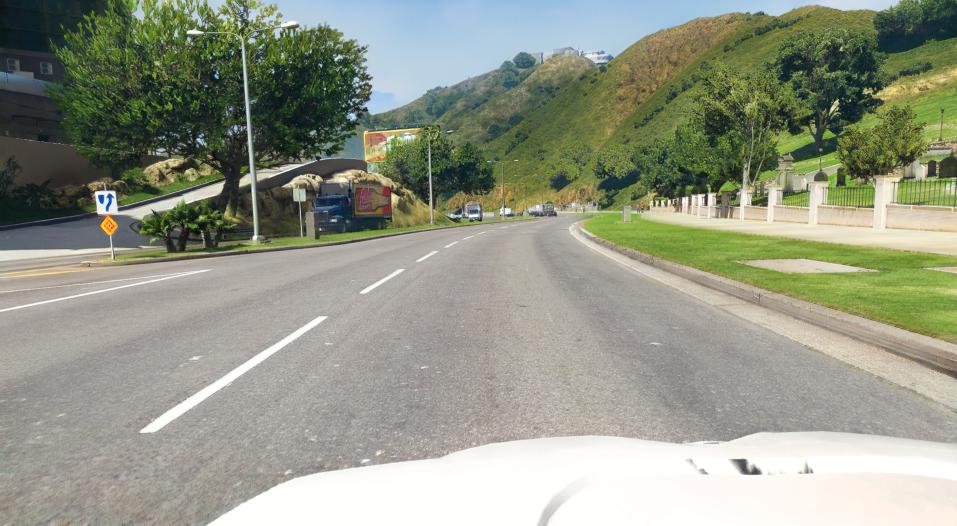}
		\end{minipage}
		\vspace{0.1mm}
		&
		\begin{minipage}{7.5cm}
			\begin{tikzpicture}
				\node[anchor=north west,inner sep=0] (image) at (0,0) {\includegraphics[width=7.5cm]{figures/comparison/ours_vistas/01316.jpg}};
				\begin{scope}[x={(image.north east)},y={(image.south west)}]
					\draw[Cerulean,thick] (0.59, 0.39) rectangle (0.795, 0.63);
			    \end{scope}
				\end{tikzpicture}
		\end{minipage}
		\vspace{0.1mm}\\
		\begin{minipage}{7.5cm}
			\begin{tikzpicture}
			\node[anchor=north west,inner sep=0] (image) at (0,0) {\includegraphics[width=7.5cm]{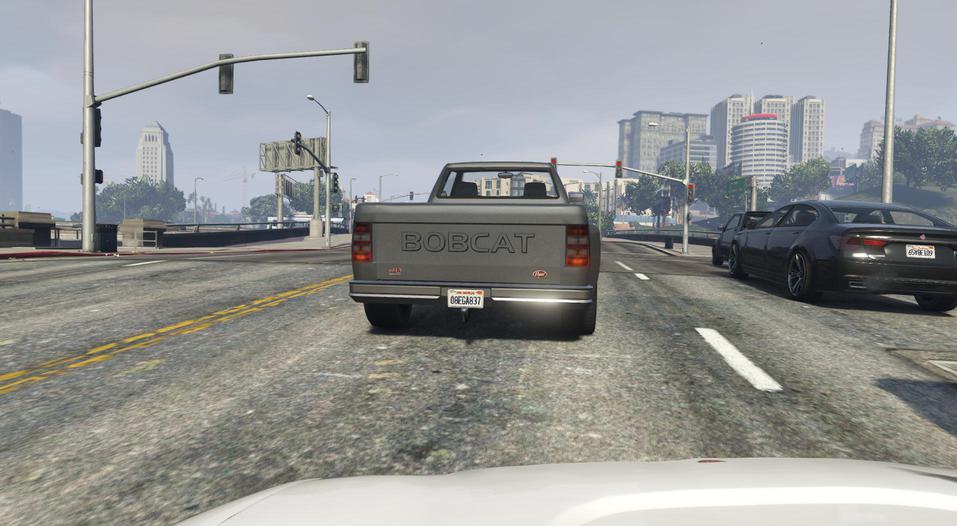}};
			\begin{scope}[x={(image.north east)},y={(image.south west)}]
				\draw[BurntOrange,thick] (0.531, 0.394) rectangle (0.736, 0.634);
		    \end{scope}
			\end{tikzpicture}
		\end{minipage}
		\vspace{0.1mm}
		&
		\begin{minipage}{3cm}
		\includegraphics[width=3cm,trim={18cm 7cm 9cm 6.91cm},clip,cfbox=BurntOrange 0.5mm -0.5mm]{figures/comparison/gta/07178.jpg}
		\includegraphics[width=3cm,trim={18cm 7cm 9cm 6.91cm},clip,cfbox=Cerulean 0.5mm -0.5mm]{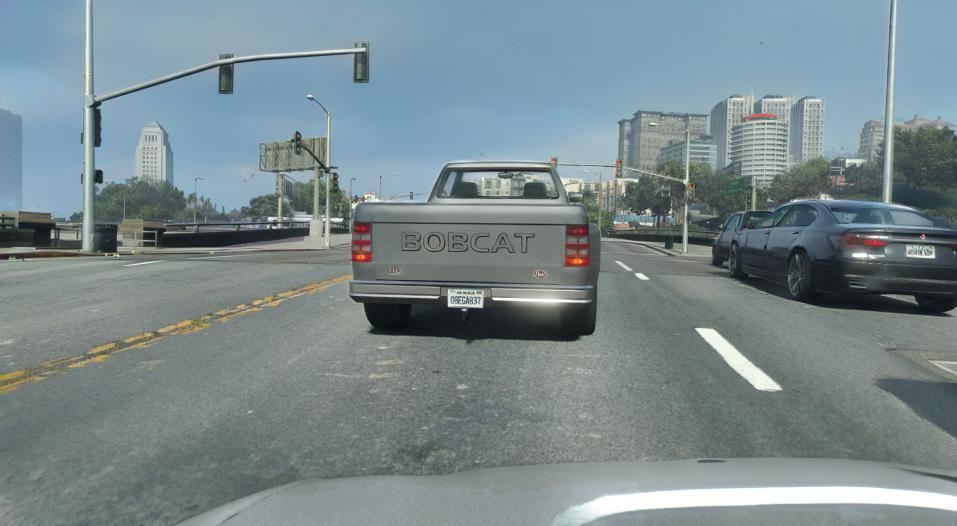}
		\end{minipage}
		\vspace{0.1mm}
		&
		\begin{minipage}{7.5cm}
			\begin{tikzpicture}
				\node[anchor=north west,inner sep=0] (image) at (0,0) {\includegraphics[width=7.5cm]{figures/comparison/ours_vistas/07178.jpg}};
				\begin{scope}[x={(image.north east)},y={(image.south west)}]
					\draw[Cerulean,thick] (0.531, 0.394) rectangle (0.736, 0.634);
			    \end{scope}
				\end{tikzpicture}
		\end{minipage}
		\vspace{0.1mm}\\
		\begin{minipage}{7.5cm}
			\begin{tikzpicture}
			\node[anchor=north west,inner sep=0] (image) at (0,0) {\includegraphics[width=7.5cm]{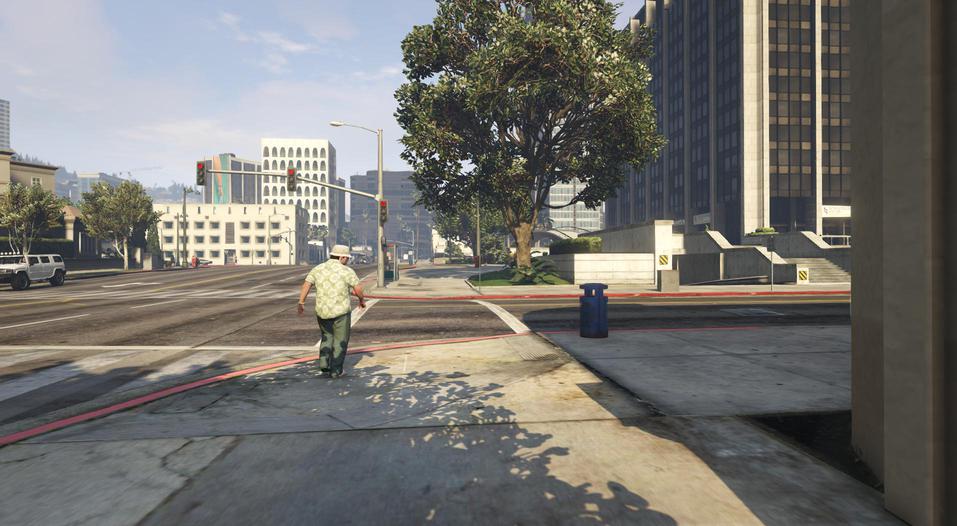}};
			\begin{scope}[x={(image.north east)},y={(image.south west)}]
				\draw[BurntOrange,thick] (0.295, 0.052) rectangle (0.5, 0.292);
		    \end{scope}
			\end{tikzpicture}
		\end{minipage}
		\vspace{0.1mm}
		&
		\begin{minipage}{3cm}
		\includegraphics[width=3cm,trim={10cm 13cm 17cm 0.91cm},clip,cfbox=BurntOrange 0.5mm -0.5mm]{figures/comparison/gta/15317.jpg}
		\includegraphics[width=3cm,trim={10cm 13cm 17cm 0.91cm},clip,cfbox=Cerulean 0.5mm -0.5mm]{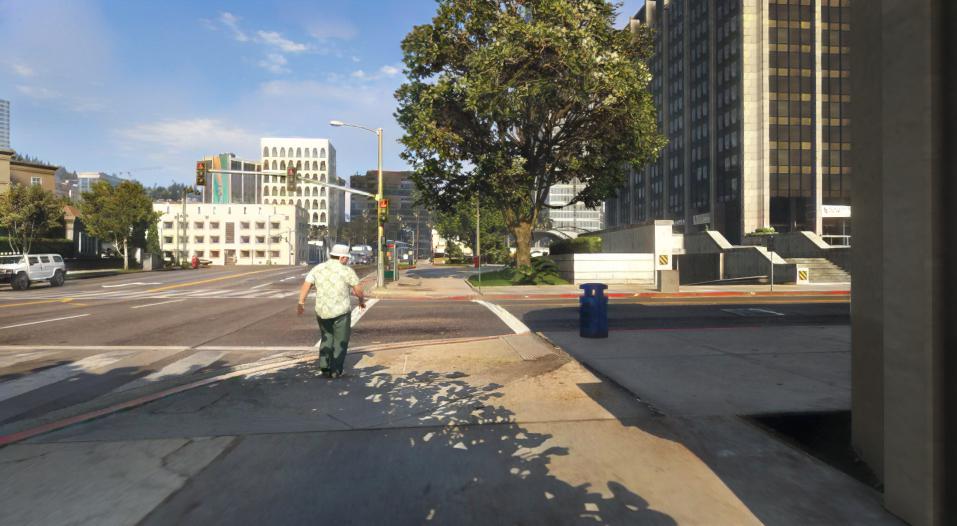}
		\end{minipage}
		\vspace{0.1mm}
		&
		\begin{minipage}{7.5cm}
			\begin{tikzpicture}
				\node[anchor=north west,inner sep=0] (image) at (0,0) {\includegraphics[width=7.5cm]{figures/comparison/ours_vistas/15317.jpg}};
				\begin{scope}[x={(image.north east)},y={(image.south west)}]
					\draw[Cerulean,thick] (0.295, 0.052) rectangle (0.5, 0.292);
			    \end{scope}
				\end{tikzpicture}
		\end{minipage}\vspace{0.1mm}\\
		\begin{minipage}{7.5cm}
			\begin{tikzpicture}
			\node[anchor=north west,inner sep=0] (image) at (0,0) {\includegraphics[width=7.5cm]{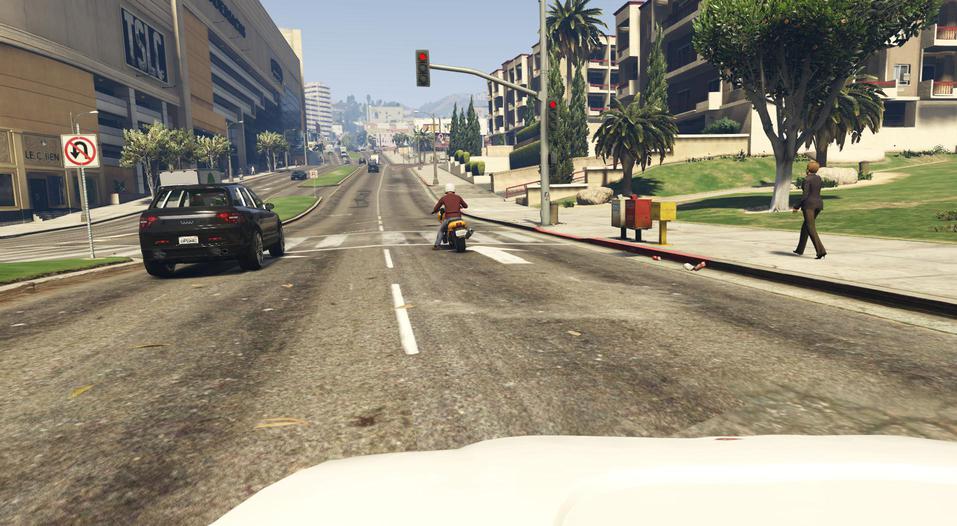}};
			\begin{scope}[x={(image.north east)},y={(image.south west)}]
				\draw[BurntOrange,thick] (0.3245, 0.1659) rectangle (0.5295, 0.4059);
		    \end{scope}
			\end{tikzpicture}
		\end{minipage}
		\vspace{0.1mm}
		&
		\begin{minipage}{3cm}
		\includegraphics[width=3cm,trim={11cm 11cm 16cm 2.91cm},clip,cfbox=BurntOrange 0.5mm -0.5mm]{figures/comparison/gta/00373.jpg}
		\includegraphics[width=3cm,trim={11cm 11cm 16cm 2.91cm},clip,cfbox=Cerulean 0.5mm -0.5mm]{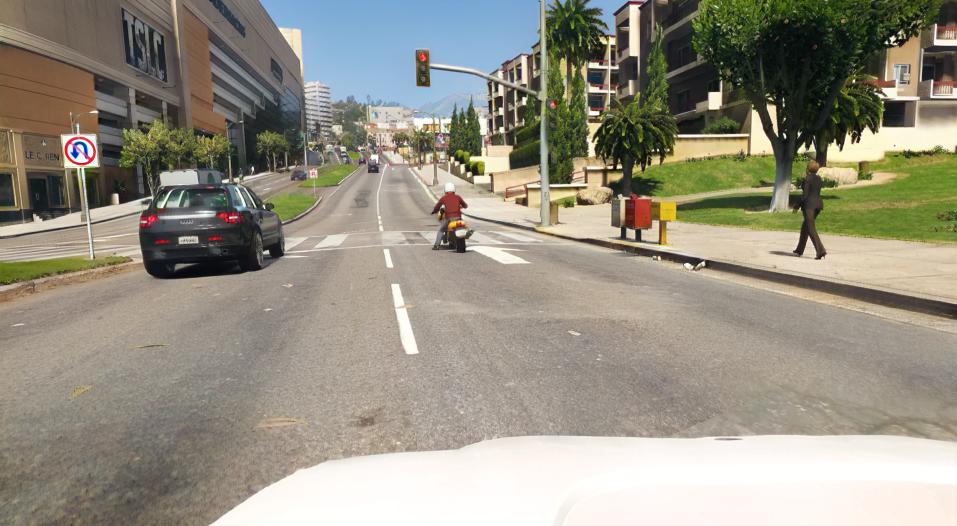}
		\end{minipage}
		\vspace{0.1mm}
		&
		\begin{minipage}{7.5cm}
			\begin{tikzpicture}
				\node[anchor=north west,inner sep=0] (image) at (0,0) {\includegraphics[width=7.5cm]{figures/comparison/ours_vistas/00373.jpg}};
				\begin{scope}[x={(image.north east)},y={(image.south west)}]
					\draw[Cerulean,thick] (0.3245, 0.1659) rectangle (0.5295, 0.4059);
			    \end{scope}
				\end{tikzpicture}
		\end{minipage}
		\vspace{0.1mm}\\
		\begin{minipage}{7.5cm}
			\begin{tikzpicture}
			\node[anchor=north west,inner sep=0] (image) at (0,0) {\includegraphics[width=7.5cm]{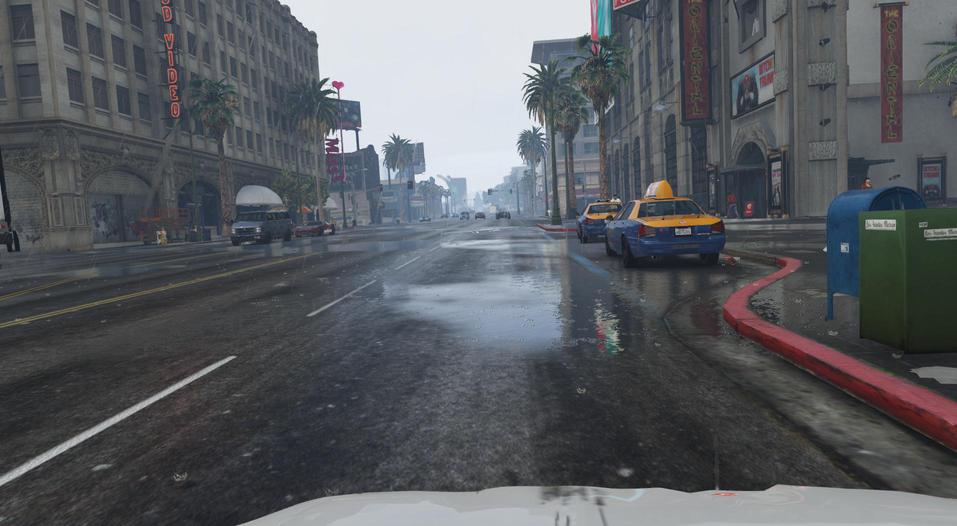}};
			\begin{scope}[x={(image.north east)},y={(image.south west)}]
				\draw[BurntOrange,thick] (0.413, 0.451) rectangle (0.618, 0.691);
		    \end{scope}
			\end{tikzpicture}
		\end{minipage}
		&
		\begin{minipage}{3cm}
		\includegraphics[width=3cm,trim={14cm 6cm 13cm 7.91cm},clip,cfbox=BurntOrange 0.5mm -0.5mm]{figures/comparison/gta/06359.jpg}
		\includegraphics[width=3cm,trim={14cm 6cm 13cm 7.91cm},clip,cfbox=Cerulean 0.5mm -0.5mm]{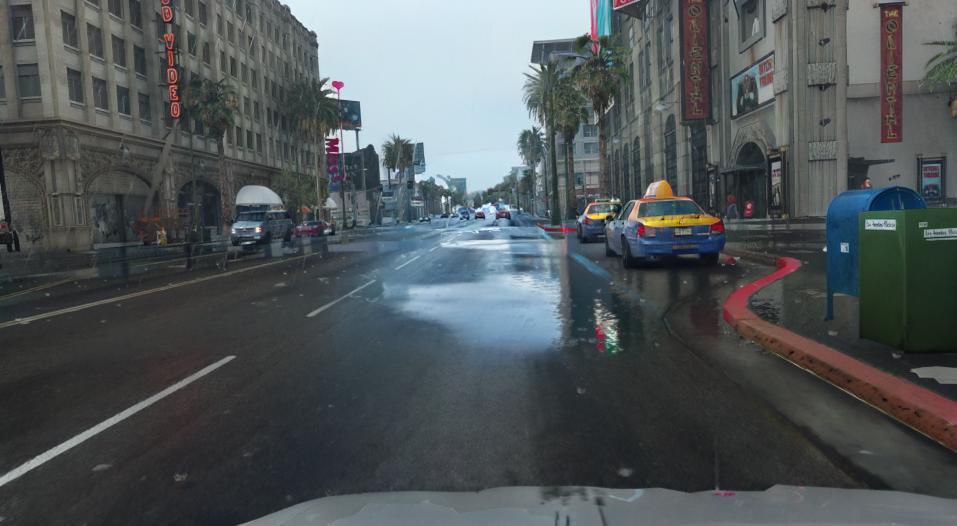}
		\end{minipage}
		&
		\begin{minipage}{7.5cm}
			\begin{tikzpicture}
				\node[anchor=north west,inner sep=0] (image) at (0,0) {\includegraphics[width=7.5cm]{figures/comparison/ours_vistas/06359.jpg}};
				\begin{scope}[x={(image.north east)},y={(image.south west)}]
					\draw[Cerulean,thick] (0.413, 0.451) rectangle (0.618, 0.691);
			    \end{scope}
				\end{tikzpicture}
		\end{minipage}\\
	\end{tabular}
	\caption{More results of enhancing GTA images with Mapillary Vistas as the target dataset. Our method rebuilds roads and makes car paint more glossy (rows 1 \& 2). It further increases saturation to match the vibrant colors of Vistas (row 3), reduces haze (row 4), and adjusts the intensity of Fresnel reflections (row 5). Insets magnify marked regions.}
	\label{fig:more_vistas}
\end{figure*}

Our method integrates learning-based approaches with conventional real-time rendering pipelines.
We expect our method to continue to benefit future graphics pipelines and to be compatible with real-time ray tracing.
Inference with our approach in its current unoptimized implementation takes half a second on a Geforce RTX 3090 GPU.
Since G-buffers that are used as input are produced natively on the GPU, our method could be integrated more deeply into game engines, increasing efficiency and possibly further advancing the level of realism.

Additional results are shown in the supplementary video.
Images produced by our method are structurally consistent with the input scenes, which can facilitate the use of ground-truth annotations that may be available for synthetic data~\cite{Richter2016, Richter2017}.
To support future research, we will release enhanced images for the GTA V and VIPER datasets.

\ifCLASSOPTIONcaptionsoff
  \newpage
\fi
\bibliographystyle{IEEEtran}
\bibliography{paper}
\end{document}